\documentclass[manuscript, screen]{acmart}
\usepackage{tabularx} % in the preamble
\usepackage{amsmath}
\usepackage{booktabs}
\usepackage{multirow}
\usepackage{float}
\usepackage{makecell}
\usepackage{siunitx}
\usepackage{graphicx}
\usepackage{comment}
\usepackage{url}
\usepackage{amsmath}
\usepackage[most]{tcolorbox}\tcbuselibrary{skins,breakable}
\usepackage{lipsum}
\usepackage{enumitem}
\usepackage{subcaption}
\usepackage{makecell}
\usepackage{comment}
\usepackage{xcolor}
\usepackage{graphicx}
\usepackage{makecell}
\usepackage{booktabs}
\usepackage{standalone}
\usepackage{tikz}
\usepackage{libertine}
\usepackage{graphicx}
\usepackage{placeins}   
\usetikzlibrary{
  positioning,
  arrows.meta,
  calc,
  shapes.geometric,
  shapes.symbols,
  fit,
  backgrounds,
  decorations.pathmorphing,
  decorations.pathreplacing
}
\usepackage{hyperref}
\hypersetup{
  colorlinks=true,
  urlcolor=blue        % oder z.B. teal, MidnightBlue
}
\AtBeginDocument{%
  }

\setcopyright{acmlicensed}
\copyrightyear{2018}
\acmYear{2018}
\acmDOI{XXXXXXX.XXXXXXX}

\acmJournal{POMACS}
\acmVolume{37}
\acmNumber{4}
\acmArticle{111}
\acmMonth{8}
\newcommand{\ctitle}[1]{{\bfseries\footnotesize #1}}
\newcommand{\cdesc}[1]{{\footnotesize #1}}

\begin{document}

\title{Nudging Sustainable Choices through LLM-Generated Recommendation Explanations}

%Nudging with Sustainability-Aware Explanations: LLM-Generated Nudges in Recommender Systems
%Sustainability-Aware Explanations for Nudging Greener Choices in Recommender Systems

\author{Haya Halimeh}
\email{haya.halimeh@upb.de}
\orcid{X}
\affiliation{%
  \institution{Paderborn University}
  \city{Paderborn}
  \country{Germany}
}

\author{Dietmar Jannach}
\affiliation{%
  \institution{University of Klagenfurt}
  \city{Klagenfurt}
  \country{Austria}}
\email{dietmar.jannach@aau.at}

\author{Oliver Müller}
\affiliation{%
  \institution{Paderborn University}
  \city{Paderborn}
  \country{Germany}}
\email{oliver.mueller@upb.de}

%%
%% By default, the full list of authors will be used in the 

%%%\renewcommand{\shortauthors}{Trovato et al.}

\begin{abstract}
Recommender systems mediate everyday consumption, offering a promising channel for encouraging sustainable choices. Prior research shows that explanations influence users' perceptions of recommendations and can support more informed decisions. We argue that explanations can also serve as behavioral nudges by foregrounding sustainability information at the moment of choice. This study investigates how different behavioral framings of sustainability information in recommendation explanations affect user choices and perceptions. Using generative AI, we generate sustainability-aware explanations by drawing on nudge theory and validate them through human evaluation and LLM-as-a-judge audits. Building on this foundation, we conduct two randomized studies ($N = 529$) in a low involvement domain (instant coffee) and a high involvement domain (hotel bookings), in which participants choose among preference matched recommendations accompanied by these explanations. Our results show that, across both domains, merely disclosing sustainability information in explanations does not change choices, whereas framing that information or invoking a descriptive social norm significantly increases sustainable selections and eases decision-making. Notably, perception and behavior diverge, as plain disclosure improves explanation evaluations without translating into more sustainable selection behavior. Our work demonstrates how LLMs can generate theory-grounded explanations at scale, pointing toward practical explanation-based interventions for social good. We conclude by discussing implications for adaptive explanation design with generative AI.
\end{abstract}




\maketitle

%%%%%%%%%%%%%%%%%%%%%%%%%%%START

\section{Introduction}
\label{sec:intro} 

% --- Para 1: opener ---
Meeting the United Nations Sustainable Development Goals (SDGs)---which call for protecting the planet and ensuring prosperity through social, economic, and environmental sustainability \cite{felfernig2023recommender}---will require substantial shifts in everyday consumption. A growing share of that consumption now takes place online. As of 2025, e-commerce accounts for more than 23\% of retail sales worldwide and is projected to approach a quarter of the global market by 2030 \cite{statista2025ecommerceshare}. On e-commerce platforms, recommender systems (RS) directly shape which options consumers see and consider, and thereby influence their decisions on a widespread scale \cite{jannach2010recommender}. 
This gives RS considerable leverage, as the systems that routinely drive consumption could equally shift it in a more sustainable direction.
Accordingly, researchers have started discussing the role of RS as a means toward meeting the SDGs
\cite{felfernig2023recommender,jannach2024recommender,said2024recommender}. 
One concrete way to leverage RS for more sustainable consumption is to nudge users toward greener options \cite{rostami2025recommender}, typically by adding cues to the user interface, such as traffic-light labels for nutritional values \cite{el2022nudging} or green transport prompts in point-of-interest recommenders \cite{mauro2024point}. Such signals, however, sit alongside the recommendation rather than within the reasoning offered for it.

% --- Para 2: explanations as intervention ---
Explanations offer another promising avenue of intervention. Since explanations justify why an item is recommended, they can bring environmental information to the forefront at the moment of decision. In this way, explanations can serve as a nudge themselves while preserving the available options and the user's freedom to choose among them \cite{thaler2009nudge}. Following \cite{tran2024less}, we refer to such approaches as \emph{sustainability-aware explanations}, as they weave sustainability considerations directly into a recommendation's rationale. To this end, explanations can be tailored to the user's decision context, for instance by comparing the health benefits of recipes in food recommenders \cite{starke2024tell} or by highlighting the environmental impact of mobility options in travel recommenders \cite{tran2024less}. Existing work in this area indicates that such explanations can raise users' sustainability awareness \cite{halimeh2025towards}, increase self-reported willingness to choose greener alternatives \cite{tran2024less}, and encourage people toward healthier recipe choices \cite{el2025nudging,starke2024tell}.

To date, however, sustainability-aware explanations are primarily static and template-based, whose fixed wording limits how closely an explanation fits a given user and item. Large language models (LLMs) offer an opportunity to lift this constraint by generating fluent, contextually rich explanations at scale \cite{o2016impact}. Using LLMs to generate explanations on-the-fly, nevertheless, raises challenges of its own. Beyond fluency, a generated explanation must remain anchored in the item's verified information, as LLMs
can otherwise produce plausible-sounding but unsubstantiated claims. While LLM-based explanations have received growing attention in RS \cite{said2025explaining}, their use within sustainability contexts remains comparatively underexplored, leaving this challenge largely untested \cite{el2025nudging}.

The empirical evidence for the effectiveness of sustainability-aware explanations is similarly thin. Although first design approaches have been proposed, some focus on intentions \cite{tran2024less}, while others observe choice only under restricted conditions, such as non-personalized settings \cite{halimeh2025towards} or a single domain like food \cite{starke2024tell}. These limitations become even more pronounced when personalization is considered. Digital nudges have largely been tested in one-size-fits-all settings \cite{starke2024psychologically,kaptein2015personalizing}, and their effect may weaken when recommendations closely match users' preferences, thus leaving little room to sway their choices \cite{starke2020little}.
Whether the known efficacy of nudging carries over when the nudge is delivered 
through a sustainability-aware explanation in personalized RS \cite{el2022nudging}, 
and whether any such effect holds across domains that differ in user involvement, 
are therefore open questions that research has only begun to address.

% --- Para 4: overture / transition to RQs ---
Addressing these gaps calls for empirical studies examining whether augmenting recommendations with LLM-generated sustainability-aware explanations can shift behavior in personalized RS, as well as approaches that can generate such explanations reliably. Since LLM outputs are sensitive to prompting strategies, with different strategies optimizing for different qualities \cite{oshchepkov2026prompting}, prompt design represents an important mechanism for controlling and evaluating explanation generation. In this work, we address these gaps by introducing a theory-grounded explanation-generation approach that operationalizes established behavioral nudging 
mechanisms---framing \cite{tversky1981framing} and descriptive social norms 
\cite{cialdini2004social,goldstein2008room}---as prompt-level templates. These templates guide an LLM in producing personalized sustainability-aware explanations that respect the item--user relationship while remaining grounded in the item's sustainability related attributes. We subsequently validate the generated explanations with respect to factual fidelity and adherence to the intended presentation mechanisms. Finally, building on this foundation, we then turn to investigate the behavioral effects through the following research questions.

% --- RQ1 ---
\noindent\emph{\textbf{RQ1}: Can LLM-generated sustainability-aware explanations (vs.\
preference-only ones) help increase the likelihood that users choose more sustainable recommended options?}

Explanations can serve different goals \cite{tintarev2007explanations,tintarev2015explaining,zanker2010knowledgeable,tintarev2022beyond,kouki2017user,peake2018explanation,zhang2020explainable}.
We identify persuasiveness---the goal of convincing users to consider an item \cite{gkika2014persuasive} and nudging them toward it \cite{nunes2017systematic,symeonidis2008providing}---as the one central to
our work, as we aim to promote greener selection behavior through our sustainability-aware explanations. While nudging toward sustainable selections has been studied in RS
\cite{el2022nudging,mauro2024point,starke2021promoting}, it has seldom been pursued through the content of an explanation \cite{el2025nudging}. Pursuing it this way for production requires a distinct, user--item explanation for every recommendation, which is what LLMs make feasible. RQ1 therefore tests whether such LLM-generated explanations are persuasive enough to shift choices, when recommendations are already tailored to the users' preferences.

% --- RQ2 ---
\noindent\emph{\textbf{RQ2}: Do different presentation strategies for LLM-generated sustainability-aware explanations differentially influence users' choices and their subjective evaluation?}

If explanations can act as nudges, how their content is presented may matter as much as the content itself. Behavioral science shows that the same information can be conveyed in different ways \cite{gena2019personalization}, and that these differences affect what users attend to and weigh at the moment of choice \cite{munscher2016review,jesse2021digital}. We therefore use an LLM to express the same sustainability content through two well-established strategies from the social sciences \cite{zhang2022green,farshbafiyan2025framing,allcott2011social,demarque2015nudging} that recur in the RS literature (see \cite{jesse2021digital})---\emph{framing} \cite{tversky1981framing} and \emph{descriptive social norms} \cite{cialdini1990focus}---each encoded as a prompt-level template. We compare these against two control conditions, a mechanism-free condition that discloses the same sustainability content without any presentation strategy, and a purely preference-only baseline that omits sustainability content altogether. This nested design separates two effects, that is, comparing the mechanism-free condition with the
baseline isolates the effect of the mere presence of sustainability content, while comparing the framing and descriptive social norms conditions with the mechanism-free condition isolates the effect of how that content is presented. Alongside choice, we also examine how these strategies affect users' subjective evaluation, drawing on standard measures from the RS literature
\cite{el2025nudging,knijnenburg2015evaluating,tintarev2012evaluating,tran2024less,willemsen2016understanding}.

% --- RQ3 ---
\noindent\emph{\textbf{RQ3}: Do the effects of LLM-generated sustainability-aware explanations on users' choices and subjective evaluation generalize across involvement domains (high vs.\ low)?}

Purchase decisions differ in how much involvement they elicit
\cite{zaichkowsky1985measuring,kapferer1985consumer}, and involvement governs how carefully users weigh the information they are given. The same explanation might therefore move users in one domain yet leave them unmoved in another, a pattern prior work already hints at, reporting that users perceive sustainability explanations differently across domains \cite{tran2024less}. We therefore run our full design in both a high- and a low-involvement domain, testing whether the effects on choice and on subjective evaluation hold in each.

% --- Study setup ---
We instantiate this design in two user studies set in domains chosen for their contrasting levels of involvement---a high-involvement domain, instantiated by hotel bookings \cite{lisha2025service}, and a low-involvement domain, instantiated by instant coffee \cite{zaichkowsky1985measuring}, with a total of $N=529$ participants. 
In each study, a preference-based recommender matches items to each participant's explicitly stated preferences and pairs every recommendation with an LLM-generated explanation that either omits sustainability content, discloses it plainly without additional presentation strategy, or presents it through framing or descriptive social norms nudging mechanisms. Eliciting preferences directly ensures every recommended item already matches the user's stated needs---an approach that has regained momentum with conversational recommenders, where users state their needs explicitly rather than 
relying on interaction history \cite{gao2021advances,jannach2021survey}.

Our results show that, First, on average, providing sustainability-aware explanations led to a significantly higher share of sustainable selections relative to a preference-only baseline, and this shift held after adjusting for demographics, domain expertise, and consumption frequency, and above and beyond users' own biospheric values \cite{de2007value}. By design, the explanations achieve this by altering only what is said about each option, not which options are offered, so that user autonomy is preserved as behavior shifts in a greener direction \cite{karlsen2019recommendations}. Second, this effect is not confined to a single setting but holds across both involvement domains. 
Third, and most importantly, behavioral change is driven by \emph{how} the sustainability content is presented, not merely \emph{that} it is present. Relative to the baseline, disclosing the content plainly left choices essentially unchanged, whereas presenting the same content through framing or descriptive social norms raised sustainable selections to 69\% and 67\%, respectively. 
Users' subjective evaluations refine this picture and reveal a notable divergence between users' perceptions and their behavioral responses. Merely making sustainability salient was enough to improve the subjective experience. Yet, this improved perception did not by itself translate into greener selection behavior. Concretely, making sustainability present improved how the system was perceived, whereas how it was presented eased the decision and changed how users act upon it, without any cost to choice satisfaction. Overall, our results suggest that LLM-generated sustainability-aware explanations are a scalable tool for nudging sustainable behavior on a large scale. 
We make all materials, code, and the item dataset publicly available at\footnote{
\href{https://github.com/HayaHalimeh/Sustainability-Aware-Explanations-in-Recommender-Systems}{https://github.com/HayaHalimeh/Sustainability-Aware-Explanations-in-Recommender-Systems}}.

% --- Roadmap ---
The remainder of this paper is organized as follows. Section~\ref{sec:background} reviews related work, positioning our contribution within research on nudging sustainable choices in recommender systems and on explanations design and the nudging mechanisms our intervention brings together. Section~\ref{sec:experimental-design} details the experimental design shared across both studies, including conditions, measures, analysis methods, and study execution. Section~\ref{sec:architecture} describes the technical approach  to retrieving the recommendations, instantiating the design in the two involvement domains, generating the explanations and auditing them. Section~\ref{sec:results} presents the results, while Section~\ref{sec:discussion} discusses the findings, limitations, and directions for future work, before Section~\ref{sec:conclusion} offers concluding remarks.

%%%%%%%%%%%%%%%%%%%%%%%%%%%%%%%%%%%%%%%%%%%%%%%%%%%%%%%%%%%%%%%%%%%%%%%%%%%%%%%%%%%%%%%%%%%%%%%%%

\section{Background}
\label{sec:background}

Our work is based on research on \emph{recommender systems for social good} \cite{jannach2024recommender,said2024recommender,felfernig2023recommender}, and more specifically on the use of RS to encourage environmentally sustainable consumption. Within this context, we study an intervention that operates through the explanation accompanying a recommendation.
The intervention draws together two bodies of work that have so far developed largely in parallel. The first, explanations in RS, concerns how an explanation is presented, how it is generated, and what content it conveys, thus supplying the vehicle of our intervention.
The second, digital nudging theory, accounts for how the arrangement and framing of decision-relevant information can steer behavior \cite{thaler2009nudge,munscher2016review}, thus providing the behavioral mechanisms we encode within that rationale.
Bringing these perspectives together reframes explanations as more than a means of communicating item relevance. Instead, they can serve as deliberate behavioral interventions designed to encourage users toward more sustainable choices. To the best of our knowledge, this role of explanations remains largely underexplored in personalized RS settings.

We develop our approach across the sections that follow. Section~\ref{nudging-sustainable} establishes the application context by situating this work within recommender systems for social good, introducing nudging and its relationship to persuasion, and casting recommenders as a form of choice architecture in which the explanation itself remains an underused channel for behavioral influence. Section~\ref{intervention} then turns to our intervention, treating its two constituent parts in turn, namely the design of the explanation that carries it (Section~\ref{explanations}) and the decision information mechanisms through which it is intended to act (Section~\ref{informational}).

\subsection{Nudging Sustainable Choices in Recommender Systems}
\label{nudging-sustainable}
 
Beyond optimizing for accuracy or engagement, a growing body of work asks how RS can serve broader societal goals, an agenda often framed as \emph{recommender systems for social good} \cite{jannach2024recommender,said2024recommender} and reflected in a series of dedicated academic forums \cite{boratto2024first,rostami2025recommender}. Its concerns reach across the RS lifecycle, from how models are designed and trained \cite{spillo2025training} to the explanations ultimately presented to end users \cite{tran2024less}. Environmental sustainability has become a particularly prominent concern within this agenda, because the everyday consumption choices that RS routinely influence carry ecological consequences at scale \cite{felfernig2023recommender}.

This makes recommender interfaces a natural place to steer users toward greener options, most directly through nudging \cite{rostami2025recommender}. At its core, a nudge is a modification to the choice architecture---the way options and information are presented to a decision maker---to predictably guide behavior toward a desired outcome the choice 
architect (i.e., the intervention designer) intends, while neither restricting options nor altering economic incentives \cite{thaler2009nudge}, and thus without compromising freedom of choice \cite{weinmann2016digital}.
As these behavioral interventions moved from physical to digital settings, they gave rise to digital nudging, the practice of influencing users within digital choice environments by adjusting interface elements such as defaults or the visual salience of options \cite{weinmann2016digital}. RS are well suited to digital nudging, since they already govern which options users see and how these are presented, making them a form of choice architecture \cite{jesse2021digital}.  For instance, by helping users integrate information so that desirable options become easier to recognize and adopt \cite{karlsen2019recommendations}. 
 
Persuasion is a concept closely related to nudging, commonly defined as an attempt to change or reinforce attitudes, beliefs, or behavior through communication or other forms of influence \cite{erdeniz2023employing}, without resorting to coercion or deception \cite{fogg2002persuasive}. Its primary goal is to convince someone to adopt a viewpoint and ultimately take a particular course of action. Classic work includes Cialdini's principles of influence, namely reciprocity, scarcity, authority, commitment, liking, and social proof, which describe strategies for increasing the effectiveness of persuasive arguments \cite{cialdini1993influence}. 
In practice, the distinction between nudging and persuasion is often treated non-dogmatically. Erdeniz et al.\ regard nudges as a form of persuasion insofar as they rely on subtle cues and framing rather than direct instruction \cite{erdeniz2023employing}. Similarly, Jesse and Jannach treat the two as overlapping families of influence strategies rather than mutually exclusive categories \cite{jesse2021digital}. 
For our purposes, both ultimately aim to influence behavior by altering how a choice, and/or the information surrounding it, are presented, while leaving decision itself to the user.
By extension, we adopt this view and note that a mechanism described here as a ``nudge'' might be termed a ``persuasive cue'' elsewhere (e.g., \cite{yoo2012persuasive}). 
 
Within RS, nudges have so far been applied mainly through cues placed alongside the recommendation. In their analysis of digital nudging in RS, Jesse and Jannach map a broad space of mechanisms but find only a few studied within recommenders to date, pointing to substantial untapped potential \cite{jesse2021digital}. The most common realization augments the interface with cues adjacent to the recommendation, such as traffic-light nutrition labels in recipe recommenders \cite{el2022nudging}, greener transport prompts in point-of-interest recommenders \cite{mauro2024point}, framing messages in news recommenders \cite{gena2019personalization}, or social cues encouraging energy-efficient behavior \cite{starke2021promoting}. 
Effective as these are \cite{hummel2019effective,caraban201923,karlsen2019recommendations,jesse2021digital}, such cues remain external to the rationale offered for the recommendation and can reduce users' perceived quality of the system and their intention to continue using it when done with due care \cite{alves2024digitally}.
An explanation, by contrast, is the part that tells the user \textit{why} an item is suggested, thereby shaping how a recommendation is interpreted and evaluated \cite{tintarev2015explaining}. Yet, explanations have attracted little attention as a deliberate behavioral intervention, leaving it an untapped channel for nudging. We posit that an explanation's rationale can be augmented with nudging principles and leveraged to move users toward more sustainable behavior.

\subsection{Intervention Design and Behavioral Mechanisms}
\label{intervention}

Our intervention combines the two literature strands introduced above into a single object. The explanation provides the form---a rationale for why an item is recommended---while a nudging mechanism determines how the sustainability information within that rationale is worded. An explanation designed in this way is therefore not only informative but also the place where the nudge takes effect.
In the following, we review the literature behind each part in turn. Section~\ref{explanations} treats the explanation as the design vehicle, situating our approach within the body of work on explanations in RS. Section~\ref{informational} then turns to the nudging mechanisms that we encode within the explanation and to the behavioral reasoning that motivates them.

\subsubsection{Explanations in Recommender Systems}
\label{explanations}

Explanations in RS are intended to communicate the reason behind a recommendation and can support multiple objectives beyond justifying the system's output. In their widely cited framework, Tintarev and Masthoff \cite{tintarev2015explaining} identify seven interrelated goals of explanations, namely transparency (communicating the rationale behind a recommendation), persuasiveness (convincing users to consider specific items), trust (increasing users' confidence in the system), scrutability (enabling users to adapt their profile), effectiveness (facilitating accurate, high-quality choices), efficiency (lowering the time cost of a choice), and satisfaction (enhancing usability or enjoyment of the system). Empirical work documents effects of explanations in practice. Explanations help users make more informed decisions \cite{lu2023user}, reduce the opacity of system reasoning \cite{zhang2020explainable}, increase perceived usefulness \cite{zanker2010knowledgeable}, improve satisfaction \cite{tran2021users}, and persuade users to accept recommendations \cite{gkika2014persuasive}. Some findings even indicate that users experience a sense of transparency and trust without consulting the explanations, as long as they are aware that such explanations are available \cite{hirschmeier2020approach}.

Among these goals, persuasiveness, the aim of convincing users to consider an item \cite{symeonidis2008providing} and nudging them toward a particular choice \cite{nunes2017systematic}, is arguably the one most directly engaged when an explanation seeks to affect choice.
The persuasive role of explanations has been recognized since early work on explaining collaborative-filtering recommendations, where certain explanations raised the acceptance of specific recommended items more than others \cite{herlocker2000explaining}. 
This observation motivated numerous studies comparing explanation styles and presentation formats by how persuasively they present
the same recommendation \cite{gkika2014persuasive,kouki2019personalized}. Explanations grounded in established persuasion principles \cite{cialdini1993influence} measurably shift selection behavior \cite{alslaity2021users}, and do so most strongly when the item they advocate already aligns with the user's preferences \cite{gkika2014investigating}. In light of these findings, persuasion has emerged as an explicit design goal for RS \cite{yoo2012persuasive}. 
Persuasiveness is also the goal our study engages most directly. Whereas prior work typically employs persuasive explanations to increase the acceptance of recommended items, we build on this objective by integrating sustainability aspects into the explanation to encourage users to choose greener alternatives.

In addition to their goals, explanations are commonly characterized along three axes: their presentation, their generation method, and the content they convey \cite{nunes2017systematic}.
The presentation axis concerns the format in which an explanation reaches the user, which Nunes and Jannach \cite{nunes2017systematic} divide into two main types, natural-language and multimedia. Natural-language formats are by far the most prevalent and include plain text \cite{chang2016crowd} as well as template-based output, where pre-defined templates are instantiated with item features before being shown. Multimedia formats comprise audio \cite{terano1989cses} and visualizations such as diagrams \cite{herlocker2000explaining}, visual annotations \cite{hou2019explainable}, and the highlighting of aspects of an alternative \cite{roitman2010increasing}. We focus on the natural-language format, since free text affords the fine-grained control over wording that our mechanisms require.

The generation axis is the one most consequential for our research. Explanation methods are commonly classified as local (providing insight into a specific recommendation) or global (enhancing understanding of how the model functions overall), and as model-specific (tailored to particular algorithm classes) or model-agnostic (applicable across different machine learning models) \cite{said2025explaining}. A related distinction separates black-box from white-box approaches \cite{friedrich2011taxonomy}. In the former, explanations are produced post hoc and remain independent of the underlying algorithm, whereas the latter derive explanations directly from the model's internal reasoning. 
Musto et al.\ note that post hoc explanations are often considered reliable and effective, and are typically preferred by end users \cite{musto2021generating}. At the same time, post hoc approaches have drawn some criticism, most prominently that a rationale produced after the fact need not faithfully reflect the process that actually generated the recommendation \cite{rudin2019stop}. We return to this caveat in our design, where we generate explanations by anchoring them directly in the recommender's underlying matching logic and the item's verified attributes (Section~\ref{generating}) and then validate them against these objectives (Section~\ref{sec:stimulus-validation}).

Within this axis, LLMs are increasingly employed, because they offer a flexible way to produce natural-language explanations that adapt to individual users and recommendation settings \cite{said2025explaining}.
For instance, Lubos et al.\ \cite{lubos2024llm} find that users prefer LLM-generated explanations for their creativity and depth, attributing this to the high text quality and broad background knowledge LLMs provide. This preference appears to carry behavioral weight as well, as such explanations can raise engagement by being human-centered and contextually adaptable \cite{silva2024leveraging}, and adjusting their tone can alter the recommender's perceived impact \cite{okoso2025impact}. They also differ in emphasis from earlier approaches by moving toward user-centric rationales for why an item fits a user \cite{guo2023towards}. We accordingly generate our explanations with an LLM, in a post hoc, model-agnostic fashion. Because LLM output is sensitive to how the model is prompted, with different prompting strategies optimizing for different qualities \cite{oshchepkov2026prompting}, the prompt is a central design lever of our intervention.
 
The content axis concerns what information an explanation actually conveys. The literature proposes a range of variants, often determined by the underlying recommender paradigm and the signals it provides, that is, the inference process. These include, among others, feature-based explanations (which emphasize the similarity between user preferences and item features) \cite{tran2021users}, collaborative-filtering-based explanations (which draw on similarities between users or items) \cite{herlocker2000explaining}, social-relation-based explanations (which highlight connections in a user's social network) \cite{zhang2020explainable}, and knowledge-based explanations (which foreground explicitly stated user requirements or desired item features) \cite{lubos2024llm}.
Because content is tied to the underlying inference process, it also governs which goal an explanation serves. An explanation such as ``This item is recommended based on your purchase history'' improves transparency and trust \cite{okoso2025impact}, whereas a justification-based one such as ``You might like this movie because it won a famous documentary award'' \cite{musto2019justifying} surfaces salient item features that enhance persuasiveness and satisfaction. The terms explanation and justification are mostly used interchangeably in RS research \cite{starke2024tell}, though Biran and Cotton \cite{biran2017explanation} distinguish them by defining explanations as describing how a recommendation is generated and justifications as describing why an item may appeal to the user.
In terms of content, our sustainability-aware explanations combine two of these variants. The first is a knowledge-based rationale that articulates how each recommended item satisfies the requirements the user has explicitly stated. Content of this kind is known to support persuasiveness \cite{tintarev2007explanations}, the goal our intervention targets, and aligns well with our recommender's logic, where users state their needs explicitly rather than relying on historical interaction data, an approach that has regained momentum with conversational recommender systems \cite{gao2021advances,jannach2021survey}. The second is a justification that surfaces the item's sustainability attributes by drawing on nudge theory to present them in a way that encourages greener choices.

When explanations are put to work for sustainability and social good, they aim to express why an item is a greener or healthier option and encourage users to choose it. A growing line of work has begun to pursue this by drawing on behavioral science to fold environmental and societal considerations into the recommendation's rationale.
Considered jointly, these studies indicate that sustainability-aware explanations can affect selection behavior positively, yet the evidence remains fragmentary, as each study is bounded along a different aspect. Tran et al.\ \cite{tran2024less} proposed explanations based on Cialdini's persuasive principles \cite{cialdini1993influence}, but measured mainly self-reported willingness to choose eco-friendly items rather than observed choice. Starke et al.\ \cite{starke2024tell} examined effects on choices, using natural-language justifications that emphasize nutritional trade-offs to steer users toward healthier recipes over popular ones, yet remained confined to a single food domain and targeted health rather than sustainability in general. In our own prior work \cite{halimeh2025towards}, we drew on nudging theory to design explanations and observed greener choices, but that study was limited by its non-personalized vignette setting with author-generated text. Most recently, and closest to our setting, El Majjodi et al.\ \cite{el2025nudging} use an LLM to generate explanations anchored in the nutritional information of recipes and show they nudge choices as effectively as established nudge forms such as traffic-light labels, yet still within the food domain.
In sum, prior work tests sustainability-aware explanations only under conditions that are synthetic, single-domain, or reliant on static text, while no single study brings the needed strands together. Personalization is the most notable of these omissions, as the potential impact of a nudge diminishes when recommendations already closely align with user preferences \cite{starke2020little,starke2024psychologically}. Two further questions remain open. First, whether any behavioral effect stems from the mere presence of sustainability content or from the way that content is presented. Second, whether such effects persist across domains that differ in how much deliberation a purchase invites.
Our intervention is designed to close this gap. The prompt serves as the design lever, encoding theory-grounded nudging strategies into a template that shapes how the sustainability justification is communicated.
These strategies draw on two established decision information mechanisms, which we use to guide the generated explanations and evaluate across two domains of differing involvement. We turn now to these mechanisms and the behavioral reasoning behind them.

\subsubsection{Decision Information Mechanisms}
\label{informational}
The mechanisms we use fall under \emph{decision information}, one of the mechanism classes in the taxonomy of M\"unscher et al.\ \cite{munscher2016review}. Its premise is that decision makers typically rely on the information available at the moment of choice, so that altering how this information is presented becomes a central lever of choice architecture \cite{thaler2009nudge}. The decision information class builds on this premise by specifying how decision-relevant information should be communicated to support better choices \cite{munscher2016review}, encouraging users to interpret it in light of their own values and goals and thereby facilitating more considered decision making \cite{mertens2022effectiveness}. From this class we draw two mechanisms, each belonging to a different subtype of 
the taxonomy and both well established for promoting sustainability across decision domains, namely  \emph{framing}, which realizes the \emph{translate Information} subtype, and \emph{descriptive social norms}, which realizes the \emph{provide a social-
reference point} subtype \cite{munscher2016review}. 
 
\textit{Translate information through framing.}  Translating information re-presents existing, decision-relevant information by changing its format or presentation. Framing realizes this subtype by changing the format of existing information to redirect attention and interpretation, thereby emphasizing different aspects of the same underlying facts \cite{munscher2016review}. It aims to alter the decision maker's perspective in ways that change their subjective evaluation of otherwise identical options \cite{weber2013doing}, for instance through the positive or negative valence attached to an attribute, which can affect how an entire statement is cognitively represented \cite{levin1998all}. A familiar illustration is that consumers prefer beef labeled ``75\% lean'' over the same beef labeled ``25\% fat,'' despite the informational equivalence \cite{levin1998all}. In sustainability settings, framing has influenced both sustainable purchase decisions \cite{zhang2022green} and sustainable hotel booking behavior \cite{farshbafiyan2025framing}.
 
\textit{Provide a social reference point through descriptive social norms.} People decide and act not in isolation but within social contexts in which others' behavior is embedded \cite{munscher2016review}, and that behavior can serve as a reference point, either through what the broader group does or through the actions of specific, highly valued individuals \cite{cialdini2003crafting}. Descriptive social norms operate through the former, indicating what most people do in a given situation and thereby signaling the typical, often implicitly sanctioned, course of action \cite{cialdini1990focus}. Making such a norm salient gives the decision maker a concrete standard against which to calibrate their own conduct, particularly under ambiguity or uncertainty \cite{cialdini2004social}. Informing hotel guests that most previous occupants reused their towels, for example, raises reuse rates relative to a conventional environmental appeal \cite{goldstein2008room}. It has also shifted real behavior across domains, from household energy use in field experiments \cite{allcott2011social} to pro-environmental product choices in online shopping \cite{demarque2015nudging}. In persuasion terms, this mechanism is the choice-architecture counterpart of Cialdini's social proof, as introduced in Section~\ref{nudging-sustainable}, thus reinforcing the earlier point that nudging and persuasion overlap.

The reason presentation should matter at all rests on dual-process accounts of cognition, which distinguish fast, automatic processing from slower, deliberate reasoning \cite{kahnemanMapsBoundedRationality2003}. Because framing and descriptive social norms plausibly engage these processes in different ways \cite{munscher2016review,thaler2021nudge}, they may produce different
behavioral effects even when the underlying facts are held constant. 
How strongly each takes hold further depends on the user's involvement. Under low involvement, people tend to decide quickly via a peripheral route that leans on heuristic cues, whereas higher involvement engages a central route of more effortful, deliberate evaluation \cite{petty1983central,petty1997elaboration}. 
The effect of a sustainability-aware explanation may therefore depend not only on whether sustainability content is present, but on the mechanism used to convey it, and on the level of involvement the domain elicits. By embedding these mechanisms into the explanation rather than presenting them as separate cues, we cast our intervention as a cognitively oriented nudge that invites users to reflect more deliberately on the choice before them.

%%%%%%%%%%%%%%%%%%%%%%%%%%%%%%%%%%%%%%%%%%%%%%%%%%%%%%%%%%%%%%%%%%%%%%%%%%%%%%%%%%%%%%%%%%%%%%%%%
\section{Experimental Design}
\label{sec:experimental-design}

Our research questions ask (RQ1) whether LLM-generated sustainability-aware explanations can be leveraged to promote more sustainable choices in a personalized recommender, (RQ2) whether applying different mechanisms to present the sustainability content matters (mechanism-free vs. framing vs. descriptive social norms) for both choices and subjective evaluations, and (RQ3) whether these effects generalize across domains that vary in their involvement level. In this section, we describe the experimental design through which we answer them. The design is shared by two randomized user studies, that differ only in their item domain.
We discuss the selection of the two item domains (Section~\ref{sec:item-domains}), and detail the procedure (Section~\ref{sec:procedure}), the experimental conditions (Section~\ref{sec:conditions}), the outcome variables and controls (Section~\ref{sec:measures}), study execution, participant recruitment and ethical safeguards (Section~\ref{sec:execution}), as well as the analysis methods (Section~\ref{sec:analysis}).
Prior to the main studies, a small-scale pilot study ($N=40$) preceded using the same experimental setup. The pilot study informed three design decisions, namely, the size of the recommendation set (Section~\ref{sec:procedure}), the pairing of textual explanations with a visual sustainability badge (Section~\ref{sec:procedure}), and the length constraints imposed on the generated explanations (Section~\ref{generating}).

\subsection{Item Domains: Low and High Involvement}
\label{sec:item-domains}

Purchase behavior can be classified by the level of involvement during the decision process \cite{zaichkowsky1985measuring}. Purchase decisions lie on an involvement continuum determined by factors such as self-image, perceived risk, purchase cost, and social influence. Routine, low-risk purchases occupy the low-involvement end, whereas high-stakes, information-intensive ones sit at the high-involvement end \cite{kapferer1985consumer,zaichkowsky1985measuring,zhang2021consumer}. Involvement is directly relevant to our intervention, as it governs how carefully users process the information an explanation provides (Section~\ref{informational}), and prior work reports that users perceive sustainability explanations differently across domains \cite{tran2024less}. Testing a nudging intervention in a single domain therefore risks overstating its generality \cite{starke2024psychologically}, a limitation of prior work we set out to overcome (Section~\ref{explanations}). We accordingly conduct two user studies in two domains that differ along this continuum, instantiated as instant coffee and hotel bookings. Instant coffee is typically treated as belonging to a low-involvement domain, given its habitual purchase nature and the low financial and social risk it carries \cite{kapferer1985consumer}. Hotel bookings represent the high-involvement domain, since such decisions are less frequent, entail greater financial commitment, and substantially affect the overall quality of a trip \cite{lisha2025service}.

\subsection{Procedure}
\label{sec:procedure}

\begin{figure}[t]
\centering
\resizebox{\linewidth}{!}{\begin{tikzpicture}[
  font=\small,
  stage/.style={draw, thick, rounded corners=3pt, align=center,
                minimum width=3.25cm, minimum height=3.45cm,
                inner sep=4pt, anchor=north, fill=gray!8},
  title/.style={font=\bfseries\small, align=center},
  body/.style={draw, rounded corners=3pt, align=left, font=\footnotesize,
               inner sep=5pt, minimum height=1.9cm, text width=2.55cm,
               anchor=north, fill=white},
  arr/.style={-{Stealth[length=3mm]}, thick},
]

\def\pitch{3.8}   % stage width + gap

% ---------- five stages ----------
\foreach \i/\ttl/\bd in {
  0/{Study Introduction \\ \& Consent}/{$\bullet$ Read study \\ description\\ $\bullet$ Provide informed \\consent},
  1/{Preference \\ Elicitation Task}/{$\bullet$  Select preferred \\ feature values \\ $\bullet$ Rank feature \\ importance},
  2/{Choice \\ Task}/{$\bullet$ View recommended items \& explanations\\ $\bullet$ Select most \\ appealing option},
  3/{Subjective \\ Evaluation}/{$\bullet$ Choice measures\\ $\bullet$ Explanation measures \\ $\bullet$ Manipulation check},
  4/{Control \\ Measures}/{$\bullet$ Background \& \\ demographics\\ $\bullet$ Biospheric value \\ orientation}
}{
  \coordinate (top\i) at ({\i*\pitch},0);
  \node[stage] (s\i) at (top\i) {};
  \node[title, anchor=north] at ($(top\i)+(0,-0.20)$) {\ttl};
  \node[body, anchor=north] at ($(top\i)+(0,-1.30)$) {\bd};
}

% ---------- connector arrows (stage to stage) ----------
\foreach \i in {0,1,2,3}{
  \pgfmathtruncatemacro{\j}{\i+1}
  \draw[arr] (s\i.east) -- (s\j.west);
}

\end{tikzpicture}}
\caption{The procedure of the experiment, shared across both studies.}
\Description{Flowchart of the experimental procedure: participants give informed 
consent, elicit their preferences by selecting and ranking item features, receive 
four recommended items (two higher- and two lower-sustainability), each shown with 
an explanation, select the item they would choose, and then complete the 
subjective-evaluation questionnaires, a manipulation check, and biospheric-value and 
demographic measures.}
\label{fig:procedure}
\end{figure}
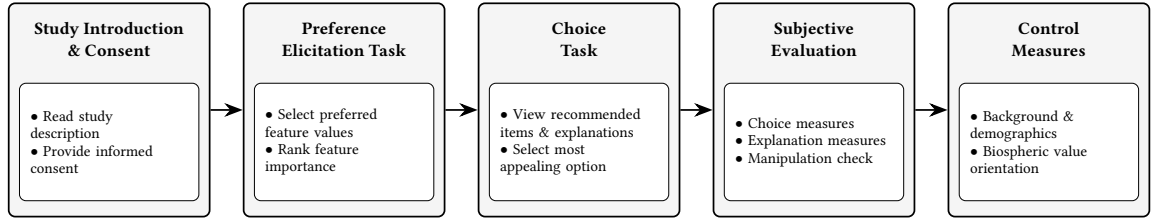

Figure~\ref{fig:procedure} illustrates the experimental procedure of both user studies. The elicited preference features, the item datasets, the sustainability operationalization, and the concrete stimuli are domain specific and follow in Section~\ref{instantiation}. 

At the beginning of each study, participants received a brief description and provided informed consent. They were told that the study concerned decision making in recommendation tasks, and sustainability was deliberately not mentioned at this point, so that participants' attention would not be primed toward it before the choice. Participants then completed a preference elicitation task by (1) selecting their preferred values from predefined, domain-tailored feature lists and (2) ordering the selected features by how important each was for their decision. In the instant-coffee study, for example, a participant might select a dark roast level and caffeinated coffee, and then rank roast level as more important than caffeine. In the hotels study, a participant might choose facilities such as a pool, bar, and restaurant, and then rank having a restaurant as more important than the other two facilities. 
Based on those stated preferences, a preference-based recommender (Section~\ref{recommender}) retrieved the items that best satisfied the participant's requirements \cite{tran2021users}.

The set of recommended items consisted of four items, two with a verified sustainability attribute from the item's metadata
and two without such an attribute. We refer to these as the \emph{higher-} and \emph{lower-sustainability} tiers, 
respectively, and use this terminology throughout the paper. We settled on this set size after pilot participants reported feeling overwhelmed by an initial set of six items \cite{el2025nudging}, and consistent with research on choice architecture showing that larger choice sets induce cognitive overload and decision paralysis \cite{iyengar2000choice} and with recommender systems research showing that fewer alternatives reduce choice difficulty and increase satisfaction \cite{bollen2010understanding}. Four items also remain within the capacity of human working memory \cite{cowan2004constant} while still permitting a balanced two-by-two split across sustainability tiers and sufficient variety for meaningful choice.
To keep informational complexity comparable across items and avoid overly long explanations \cite{hendrawan2024explanations},
sustainability was operationalized through a single verifiable, domain-specific attribute taken from the item's metadata. In particular, a packaging-related sustainability claim  such as recyclable packaging for instant coffee or a third-party certification such as Green Key for hotels.

Every item was accompanied by an LLM-generated explanation produced with the help of a condition-specific prompt (Section~\ref{generating}), so that no item was left unexplained and the mere presence or absence of an explanation could not act as a confound. In the three sustainability-aware conditions (mechanism-free, framing, and descriptive social norms), the two higher-sustainability items additionally carried a visual sustainability badge, which made the items' tier visually identifiable. 
This mirrors how sustainability is signaled in practice \cite{potter2021effects} and is is in line with evidence that sustainability cues combining an icon and text are perceived as more effective than text alone \cite{lee2025ecolabels,jesse2021digital}.
Based on pilot testing, we adopted this design while holding the badge constant across the three sustainability-aware conditions. Thus, the primary difference between these conditions lies in how the explanation text presents sustainability information, which directly reflects the contrast examined in RQ2.

\begin{figure}[ht]
\centering
\begin{subfigure}{0.49\textwidth}
  \centering
  \includegraphics[width=\textwidth]{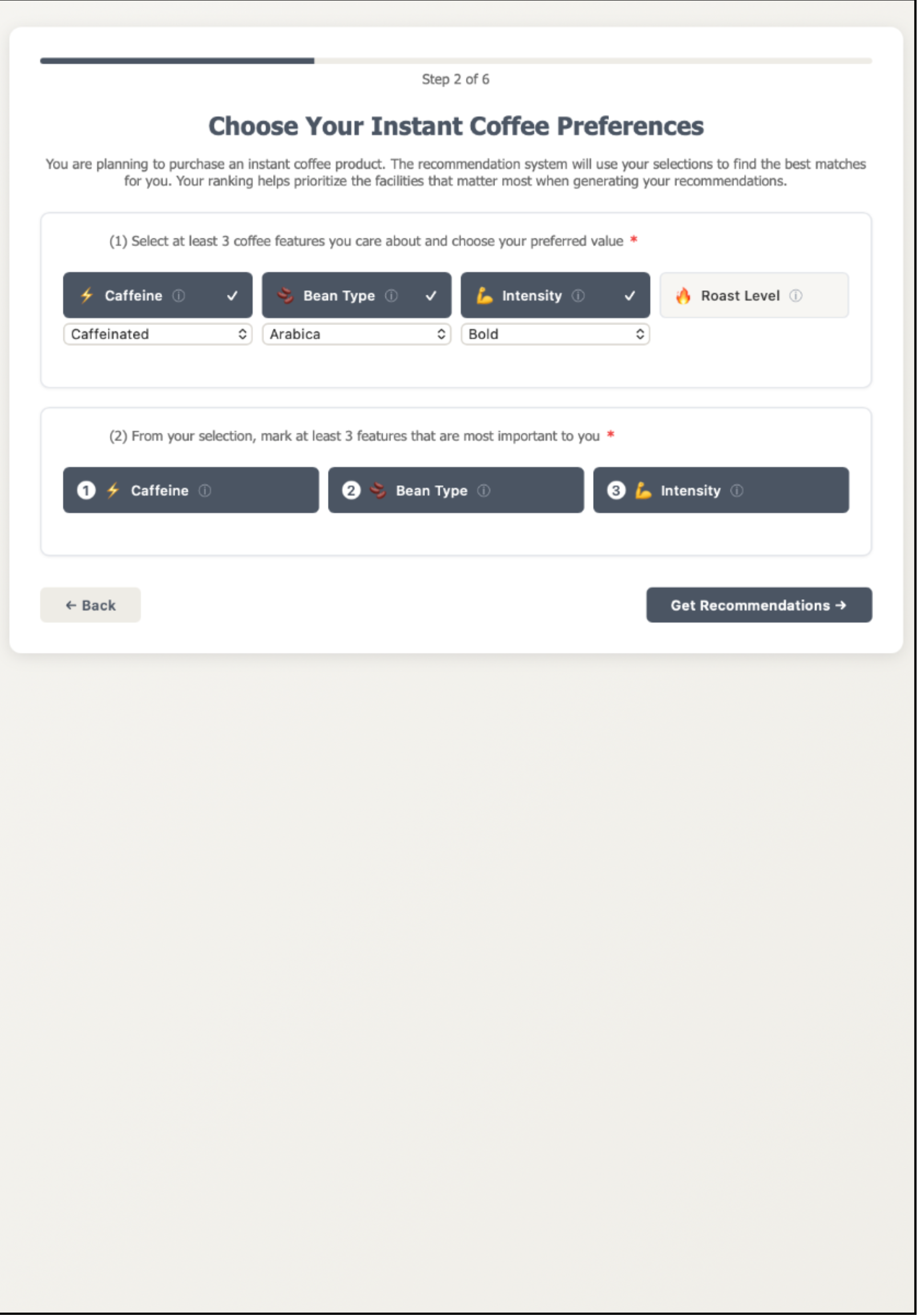}
  \caption{Preference elicitation task}
  \label{fig:example_low_pref}
\end{subfigure}
\hfill
\begin{subfigure}{0.49\textwidth}
  \centering
  \includegraphics[width=\textwidth]{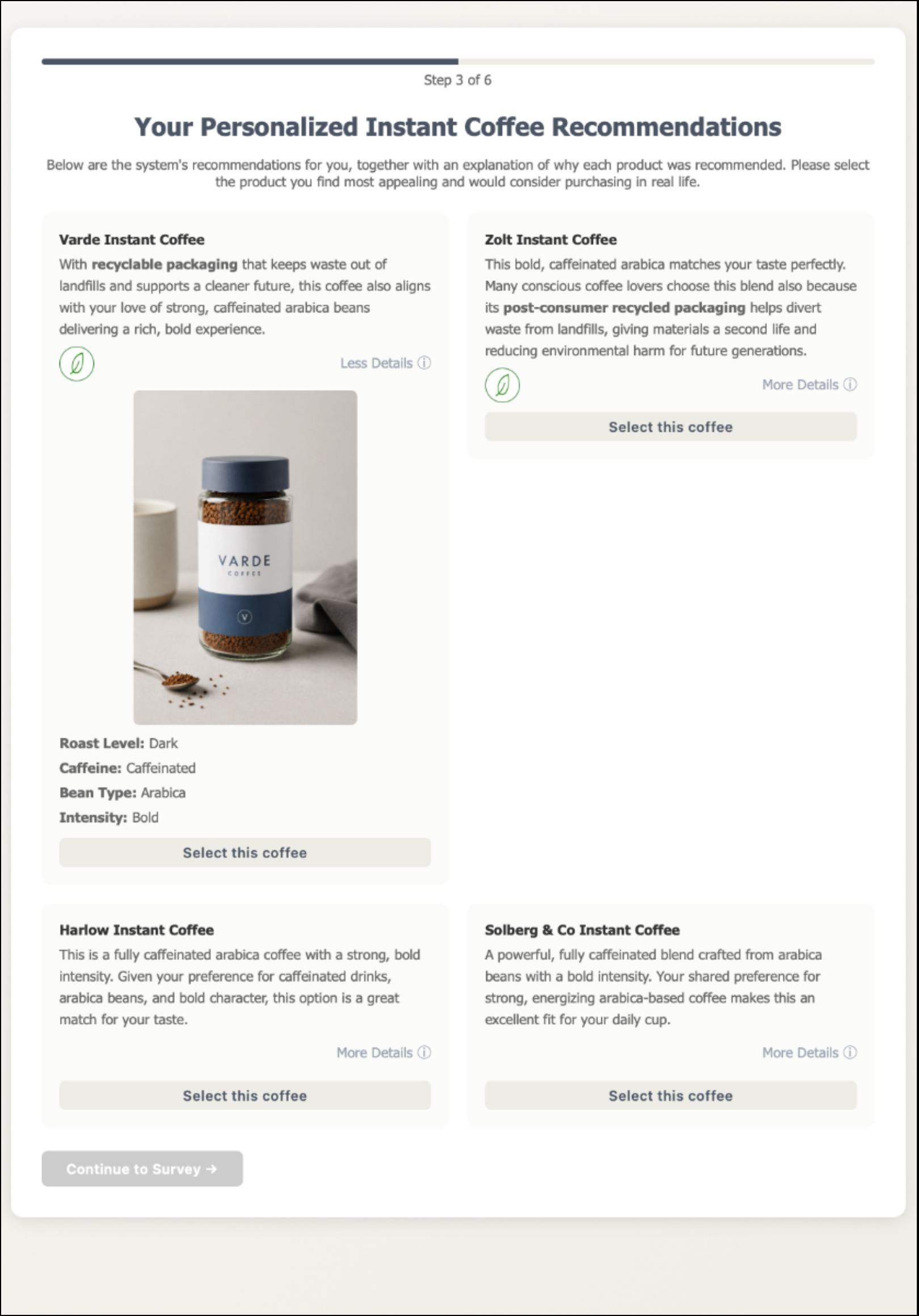}
  \caption{Choice task}
  \label{fig:example_low_choice}
\end{subfigure}
\caption{An example from the low-involvement user study. Participants first 
elicited their preferences by selecting and ranking coffee features (left), then 
chose among four recommended items, each shown with an explanation (right). Items 
initially displayed only a title and explanation. The image and further attribute 
details appeared after clicking ``More Details'' (shown expanded for one item).}
\Description{Two interface screenshots from the instant-coffee study. Left: the 
preference elicitation screen, where participants select and rank coffee features 
such as caffeine, bean type, and intensity. Right: the choice screen showing four 
recommended coffee products as cards, each with a title and explanation, one card 
expanded to reveal a product image and attribute details.}
\label{fig:example_low}
\end{figure}

\begin{figure}[h]
\centering
\begin{subfigure}{0.49\textwidth}
  \centering
  \includegraphics[width=\textwidth]{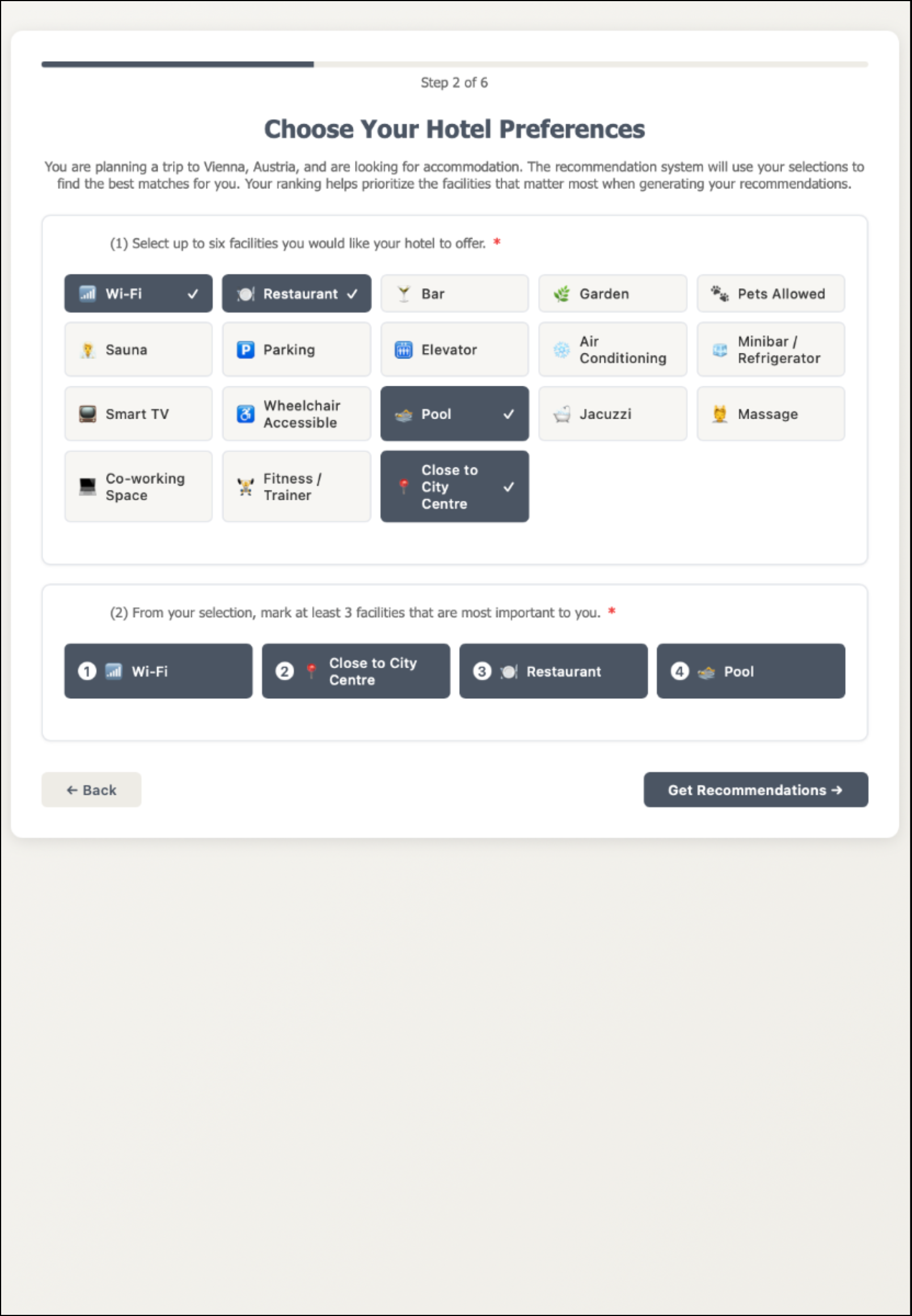}
  \caption{Preference elicitation task}
  \label{fig:example_high_pref}
\end{subfigure}
\hfill
\begin{subfigure}{0.49\textwidth}
  \centering
  \includegraphics[width=\textwidth]{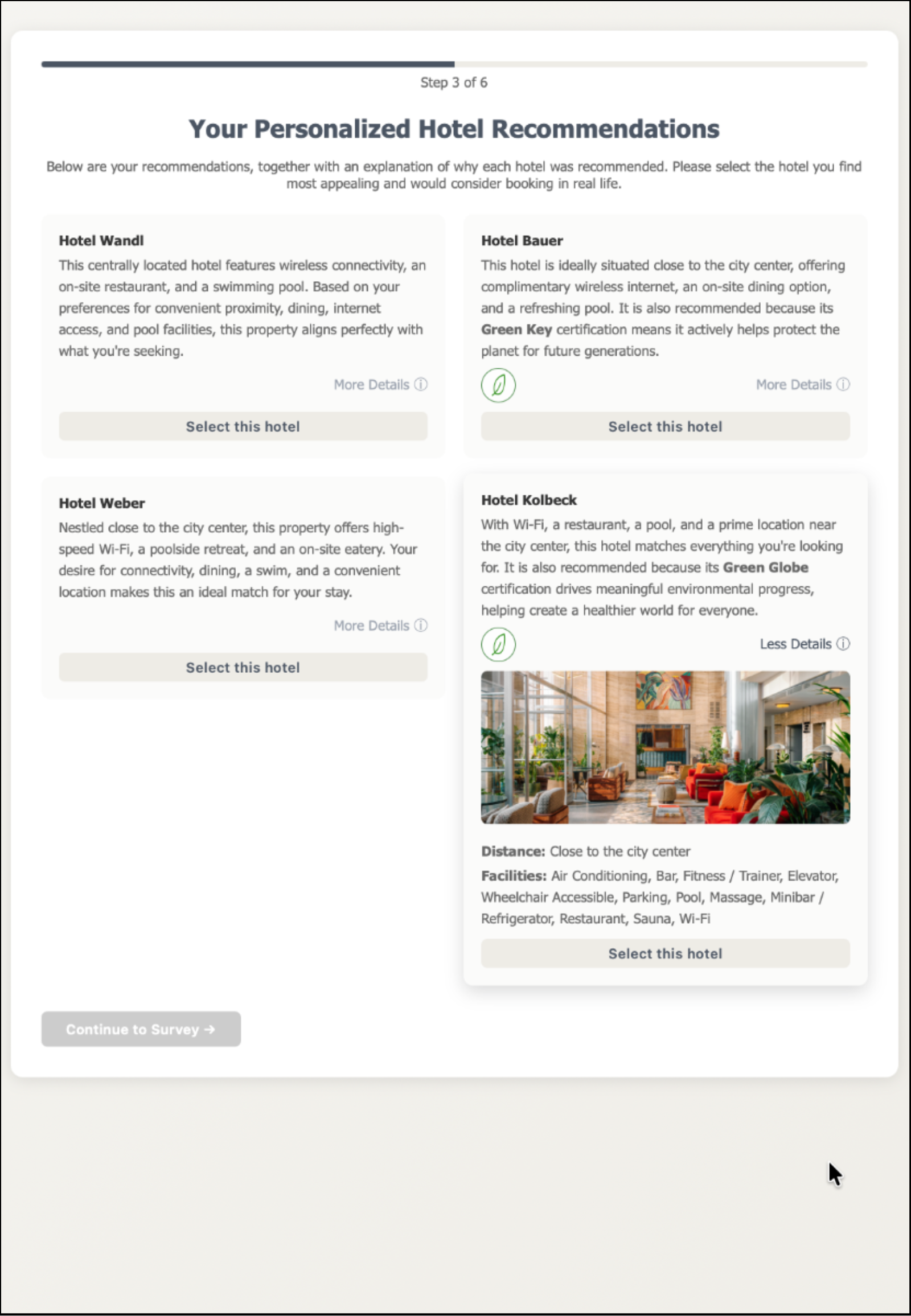}
  \caption{Choice task}
  \label{fig:example_high_choice}
\end{subfigure}
\caption{An example from the high-involvement user study. Participants first 
elicited their preferences by selecting and ranking hotel facilities (left), then 
chose among four recommended hotels, each shown with an explanation (right). As in 
the low-involvement study, items initially displayed only a title and explanation. 
The image and further facility details appeared after clicking ``More Details'' 
(shown expanded for one item).}
\Description{Two interface screenshots from the hotel-booking study. Left: the 
preference elicitation screen, where participants select and rank hotel facilities. 
Right: the choice screen showing four recommended hotels as cards, each with a title 
and explanation, one card expanded to reveal a hotel image and additional facility 
details.}
\label{fig:example_high}
\end{figure}

Item order was randomized for each participant to control for positional effects. 
When the recommendations were displayed, each item initially showed only its 
title and explanation. Participants could click any item to reveal its image and full attribute details. The explanations were consequently the information presented up front, while images and details were available on demand rather than shown by default.
Participants viewed all four items and were instructed to select the option they found most appealing and would consider purchasing or booking in real life. 
After the choice task, they completed questionnaires measuring their subjective perception additional to a manipulation-check item. (Section~\ref{sec:measures}). Finally, participants completed a measure of biospheric value orientation---the degree to which protecting nature and the environment is a guiding principle in one's life \cite{de2007value}---followed by their domain expertise and demographic characteristics. We included the biospheric-value measure as a control variable (Section~\ref{sec:measures}), since participants with strong pre-existing environmental values might choose sustainable items regardless of our manipulation \cite{o2016impact}. Figures~\ref{fig:example_low} and~\ref{fig:example_high} show the preference elicitation and choice tasks as participants experienced them in each domain.

\subsection{Experimental Conditions}
\label{sec:conditions}

Our manipulation concerned the content of the explanations. Each study employed a between-subjects design with four conditions. The first condition was the \emph{preference-only baseline}, in which each of the four recommended items received an explanation describing how well it matched the participant's stated preferences and contained no sustainability content, following the feature- and preference-based explanation style common in the literature \cite{lubos2024llm,tran2021users}. 
The second condition was the \emph{mechanism-free} condition, in which the explanations stated the sustainability attribute of the two higher-sustainability items factually and without applying any additional behavioral presentation mechanism. In the two \emph{mechanism-based} conditions, the same sustainability content was instead expressed through one of two decision information mechanisms, \emph{framing} or \emph{descriptive social norms}, each implemented with the help of theory-grounded condition-specific prompts detailed in Section~\ref{generating}. Across the three sustainability-aware conditions (mechanism-free, framing, and descriptive social norms), explanations for the two lower-sustainability items remained preference-based. 

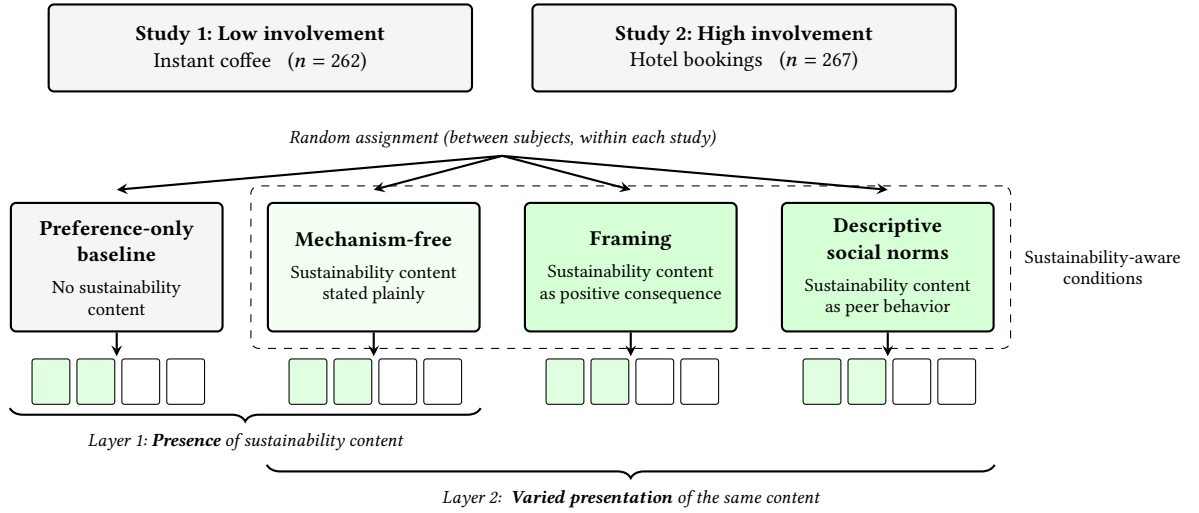
\begin{figure}[t]
    \centering
    \begin{tikzpicture}[
  font=\small,
  study/.style={draw, rounded corners=2pt, align=center, minimum width=5.6cm,
                minimum height=1.15cm, fill=gray!8, thick},
  cond/.style={draw, rounded corners=2pt, align=center, minimum width=2.8cm,
               minimum height=1.7cm, thick, fill=white},
  ctrl/.style={cond, fill=gray!8},
  mfree/.style={cond, fill=green!5},   % light green: sustainability content, no mechanism
  mechd/.style={cond, fill=green!16},   % darker green: mechanism-based (framing, descriptive)
  item/.style={draw, rounded corners=1pt, minimum width=0.5cm,
               minimum height=0.6cm, inner sep=0pt},
  hi/.style={item, fill=green!12},
  lo/.style={item, fill=white},
  lab/.style={font=\footnotesize\itshape, align=center},
  arr/.style={-stealth, thick},
  brace/.style={decorate, decoration={brace, amplitude=5pt}, thick}
]

% symmetric grid: conditions centered on x=0 at pitch 3.4cm
% ---------- studies ----------
\node[study] (s1) at (-3.2, 2.2) {\textbf{Study 1: Low involvement}\\ Instant coffee \; ($n = 262$)};
\node[study] (s2) at ( 3.2, 2.2) {\textbf{Study 2: High involvement}\\ Hotel bookings \; ($n = 267$)};

% ---------- random assignment ----------
\node[lab] (ra) at (0, 1.0)
  {Random assignment (between subjects, within each study)};

% ---------- four conditions ----------
\node[ctrl] (c1) at (-5.1, -0.7)
  {\textbf{Preference-only}\\ \textbf{baseline}\\[2pt] \footnotesize No sustainability\\[-2pt] \footnotesize content};
\node[mfree] (c2) at (-1.7, -0.7)   % was [ctrl]
  {\textbf{Mechanism-free}\\[2pt] \footnotesize Sustainability content\\[-2pt] \footnotesize stated plainly};
\node[mechd] (c3) at ( 1.7, -0.7)   % was [mech]
  {\textbf{Framing}\\[2pt] \footnotesize Sustainability content\\[-2pt] \footnotesize as positive consequence};
\node[mechd] (c4) at ( 5.1, -0.7)   % was [mech]
  {\textbf{Descriptive}\\ \textbf{social norms}\\[2pt] \footnotesize Sustainability content\\[-2pt] \footnotesize as peer behavior};

% sustainability-aware group frame
\node[draw, dashed, rounded corners=3pt, inner sep=6pt,
      fit=(c2)(c3)(c4)] (grp) {};
\node[font=\footnotesize, align=center, right=2pt of grp]
  {Sustainability-aware\\ conditions};
  
% arrows to conditions
\foreach \c in {c1,c2,c3,c4}{ \draw[arr] (ra.south) -- ($(\c.north)+(0,0.16)$); }

% ---------- choice set per condition: 4 items, 2 tiers ----------
\foreach \c in {c1,c2,c3,c4}{
  \node[hi] (\c-i1) at ($(\c.south)+(-0.87,-0.65)$) {};   
  \node[hi] (\c-i2) at ($(\c-i1.east)+(0.08,0)$) [anchor=west] {};
  \node[lo] (\c-i3) at ($(\c-i2.east)+(0.08,0)$) [anchor=west] {};
  \node[lo] (\c-i4) at ($(\c-i3.east)+(0.08,0)$) [anchor=west] {};
  \draw[arr, shorten >=1pt] (\c.south) -- ($(\c.south)+(0,-0.35)$);
}
% badge dots on higher-tier items in sustainability-aware conditions
%\foreach \c in {c2,c3,c4}{ \foreach \i in {i1,i2}{
%  \node[circle, fill=green!55!black, inner sep=1.0pt]
%    at ($(\c-\i.north east)+(-0.08,-0.08)$) {}; }}

% tier legend, kept short and clear of the braces
%\node[lab, anchor=west] at (-6.7,-2.35)
%  {2 higher-tier {\tikz\node[hi, minimum width=0.34cm, minimum height=0.4cm]{};}
%   (with badge {\textcolor{green!55!black}{$\bullet$}} when sustainability-aware)
%   \; + \; 2 lower-tier {\tikz\node[lo, minimum width=0.34cm, minimum height=0.4cm]{};}
%   \; per participant; one choice --- only the explanation text varies across conditions};

% ---------- nested contrasts (braces below) ----------
\draw[brace, decoration={mirror}]
  ($(c1.south west)+(0,-1)$) -- ($(c2.south east)+(0,-1)$)
  node[midway, below=6pt, lab] {Layer 1: \textbf{Presence} of sustainability content};
\draw[brace, decoration={mirror}]
  ($(c2.south west)+(0,-1.75)$) -- ($(c4.south east)+(0,-1.75)$)
  node[midway, below=6pt, lab] {Layer 2: \textbf{ Varied presentation} of the same content};

\end{tikzpicture}
    \caption{Overview of the experimental design. Both studies share an identical between-subjects design with four conditions. Every participant chooses among four preference-matched, randomly positioned recommended items---two from the higher- and two from the lower-sustainability tier. The two braces mark the nested contrasts: the \emph{presence} of sustainability content (mechanism-free vs.\ baseline) and its \emph{presentation} (framing and descriptive social norms vs.\ mechanism-free).}
    \Description{Diagram of the four-condition between-subjects design. Participants are randomly assigned to one of four conditions---preference-only baseline, mechanism-free, framing, or descriptive social norms---and each chooses among four preference-matched items, two from the higher- and two from the lower-sustainability tier. Two braces group the conditions into nested contrasts: presence of sustainability content (mechanism-free versus baseline) and its presentation (framing and descriptive social norms versus mechanism-free).}
    \label{fig:conditions}
\end{figure}

We refer to this design as nested because its conditions form two layered contrasts. The first layer contrasts the mechanism-free condition with the preference-only baseline, isolating the effect of simply including sustainability content. The second layer contrasts the framing and descriptive social norms conditions with the mechanism-free condition, isolating potential effects of how that same content is presented. Figure~\ref{fig:conditions} provides an overview of the experimental design and conditions.

\subsection{Outcome Variables and Controls}
\label{sec:measures}

The outcome variables include choice and  subjective evaluations, mapping directly onto the research questions. The choice outcome captures the behavioral effect addressed by all three research questions. The subjective evaluation measures capture participants' experiential response to the explanations, informing RQ2.
Control variables account for individual differences that could otherwise confound the estimated effects, either by masking true effects or creating spurious ones. For example, participants who are more environmentally conscious in nature may make greener choices regardless of any explanations.

\textit{Choice outcome.} As established in Section~\ref{explanations}, persuasiveness, here understood as moving users toward the sustainable option, is the explanation goal our study aims at. The persuasiveness of explanations can be assessed in several ways. It can be operationalized as the change between an initial rating and a subsequent re-rating of the same item after an explanation is shown \cite{gkika2014persuasive,zhu2012switch}, or measured behaviorally, by comparing how often users select or purchase recommended items under different explanation interfaces \cite{cosley2003seeing,starke2024tell}. Our analysis followed the latter paradigm. The unit of analysis was the individual participant, each contributing a single choice among the four recommended items. This choice was our primary outcome, coded $y=1$ if the selected item belonged to the higher-sustainability tier and $y=0$ otherwise.

\textit{Subjective evaluations.} \textit{Subjective evaluations.} Choice shows \emph{whether} the intervention 
works, but not how users \emph{experience} it. 
We therefore complemented choice with subjective measures for two reasons. First, they let us locate the route of the effect by assessing wether the explanations act by changing how users perceive the recommendation or by influencing 
the decision process itself. Second, they reveal whether any persuasive gains come at the cost of other explanation goals such as clarity or satisfaction \cite{tintarev2015explaining}. We measured several subjective evaluation constructs that reflect participants' perceptions, each built from items adapted from prior work. The full item wordings are reported alongside their measurement statistics in Appendix~\ref{app:measurement-validation}. Two constructs concerned the decision.  \emph{Choice satisfaction} captured how satisfied participants were with their decision, with items adapted from \cite{knijnenburg2015evaluating,starke2021promoting}. \emph{Choice difficulty} measured how effortful the decision felt, adapted from \cite{willemsen2016understanding,el2022nudging}. The remaining constructs concerned the explanations. \emph{Explanation clarity} captured how comprehensible the explanation was, that is, the extent to which participants found it easy to understand and clear about why an item was recommended, combining items on understandability \cite{lubos2024llm,peake2018explanation} and transparency \cite{tintarev2022beyond,tintarev2007explanations}. \emph{Explanation usability}\footnote{Note that this item concerns the explanations, not the usability of the system as a whole.} \cite{tintarev2012evaluating}, measured the extent to which the explanations helped participants choose among the recommended options, with a single item adapted from \cite{el2025nudging}. Additionally, because persuasiveness can conflict with other explanation goals such as effectiveness \cite{chen2014sentiment}, we grouped three items into a construct we term \emph{sustainability impact},
that assessed whether the sustainability information in the explanation was itself perceived as impactful. This construct combined three items on how convincing the sustainability information was \cite{tran2024less}, how far it raised awareness of sustainability considerations \cite{tran2024less}, and how well it helped participants understand the items' sustainability aspects \cite{el2025nudging}.

\textit{Control variables.} As individuals' own interest in sustainability can influence how they respond to sustainability information \cite{o2016impact}, we controlled for participants' biospheric value orientation, measured through four items rating preventing pollution, respecting the earth, unity with nature, and protecting the environment as guiding principles in their lives \cite{de2007value}. We further controlled for background characteristics, namely age, gender, ethnicity, domain expertise, and consumption frequency.

Beyond these outcome and control measures, we included a single \textit{manipulation-check} 
item---``The explanations sometimes used sustainability considerations to justify 
the recommendations''---to verify that 
participants in the sustainability-aware conditions noticed the sustainability 
content.
All questionnaire items were assessed on a 7-point Likert scale (1 = strongly disagree, 7 = strongly agree). The reliability and validity of the resulting scales are examined in Appendix~\ref{app:measurement-validation}.

\subsection{Study Execution, Participants, and Ethics}
\label{sec:execution}

\textit{Sample size determination.} We determined the target sample size through an a priori power analysis in G*Power \cite{faul2007g} for a logistic regression with a binary predictor (one-tailed, $\alpha = .05$, power $= .80$). In plain terms, this analysis estimates how many participants are needed so that an effect of an assumed size, if it exists, is detected with 80\% probability. We assumed an odds ratio of $2.5$, which corresponds to the explanations raising a baseline sustainable-choice probability of $.50$ to roughly $.71$, an effect magnitude in line with nudging effects reported in comparable choice experiments \cite{el2025nudging}. Under these assumptions, the analysis indicated a required sample of $129$ per contrast, that is, approximately $65$ participants per condition, or roughly $260$ participants per study. Across the two involvement domains, the target was therefore approximately $520$ participants in total.

\textit{Recruitment and compensation.} We recruited participants from the crowdsourcing platform Prolific \cite{palan2018prolific}, with eligibility restricted to fluent English speakers, and aimed for a gender-balanced sample. The study was presented as a 6-minute task, and participants were compensated at \pounds10 per hour, above Prolific's recommended minimum, to ensure fair payment and support data quality \cite{palan2018prolific}.

\textit{Exclusions.} Prior to analysis, participants were excluded if they failed one or more attention checks embedded in the questionnaire \cite{oppenheimer2009instructional}, or if their completion time fell outside two standard deviations of the sample mean, a common criterion for identifying inattentive responding in online studies \cite{aguinis2013best}.

\textit{Sample.} The final sample comprised $N = 529$ participants across the two studies (low: $n = 262$; high: $n = 267$). Age was recorded in bands, with most participants in the 25--34 range ($41.8\%$), followed by 18--24 ($21.7\%$) and 35--44 ($20.2\%$). In total, 265 participants ($50.1\%$) identified as female and 264 ($49.9\%$) as male.

\textit{Ethical statement.} Before taking part, participants were informed about the study's purpose and procedure and consented to the use of their demographic data and responses in anonymized form for research purposes. Participation was voluntary, and participants could withdraw at any time without penalty. All data were used exclusively for research, stored in anonymized form, and not shared with third parties. The study protocol was approved by our institutional ethics committee and reviewed by our institution's data protection officer in accordance with the General Data Protection Regulation (GDPR).

\subsection{Analysis Methods}
\label{sec:analysis}

Having defined the outcomes, we next describe the statistical analyses used to address each research question. all models include age, gender, ethnicity, domain expertise, consumption frequency, and item domain as covariates. Our analysis under different explanation variants mirrors prior explanation-based studies \cite{el2025nudging,starke2024tell}. 

RQ1 examines whether the known efficacy of nudging carries over when the nudge is delivered through a sustainability-aware explanation in personalized RS. For this purpose, we fit a binary logistic regression \cite{hosmer2013applied} on the pooled sample (the combined data of both studies), to predict whether the selected item belonged to the higher-sustainability tier. The focal predictor contrasts the three sustainability-aware conditions (mechanism-free, framing, and descriptive social norms), taken together, against the preference-only baseline that omitted any sustainability content. We report the odds ratio (OR), its 95\% confidence interval, and the corresponding $p$-value. An OR significantly above 1 would indicate that sustainability-aware explanations raise the odds of a sustainable choice, supporting RQ1, while an OR near 1 would indicate that the explanations leave choices significantly unaffected. 

RQ2 concerns whether presenting the sustainability content through different nudging strategies differ in their effect, on both choice and participants' subjective evaluations. For choice, we fit a binary logistic regression on the pooled sample with condition as a four-level categorical predictor (preference-only, mechanism-free, framing, and descriptive social norms) and estimate two sets of contrasts from it, corresponding to the two layers of the nested design (Section~\ref{sec:conditions}). The first set takes the preference-only baseline as the reference and estimates the effect of each sustainability-aware condition relative to the purely preference-only baseline without any sustainability content. The second set takes the mechanism-free sustainability condition as the reference and estimates the effect of the two mechanism-based sustainability conditions relative to it, thereby isolating the effect of how the content is presented. 
For the subjective evaluations (explanation clarity, explanation usability, sustainability impact, choice satisfaction, and choice difficulty), we fit instead a linear regression model for each construct with the same predictor variables and evaluate the same two sets of contrasts. We report coefficients ($b$), representing mean differences on the 7-point scale, alongside 95\% confidence intervals and $p$-values to assess whether participants perceived the explanations differently in each condition.

RQ3 asks whether the effects examined in RQ2 hold across the two involvement levels, that is, whether results from the low-involvement domain and the high-involvement domain agree. For this purpose, for each outcome, we extend the corresponding model with a condition-by-domain interaction that tests whether the condition effects differ between the two domains. For choice, we assess the overall interaction with a likelihood-ratio test comparing the interaction model against the main-effects model \cite{hosmer2013applied}. For each subjective evaluation, we use the analogous nested-model $F$-test, which compares the fit of the two linear models. A non-significant interaction would indicate that the effect of condition does not differ by domain. Such an outcome would support the generalization of our intervention across involvement levels, whereas a significant interaction would indicate domain-dependent effects, which we would probe through predicted values by condition and domain.

%%%%%%%%%%%%%%%%%%%%%%%%%%%%%%%%%%%%%%%%%%%%%%%%%%%%%%%%%%%%%%%%%%%%%%%%%%%%%%%%%%%%%%%%%%%%%%%%%

\section{Technical Approach}
\label{sec:architecture}

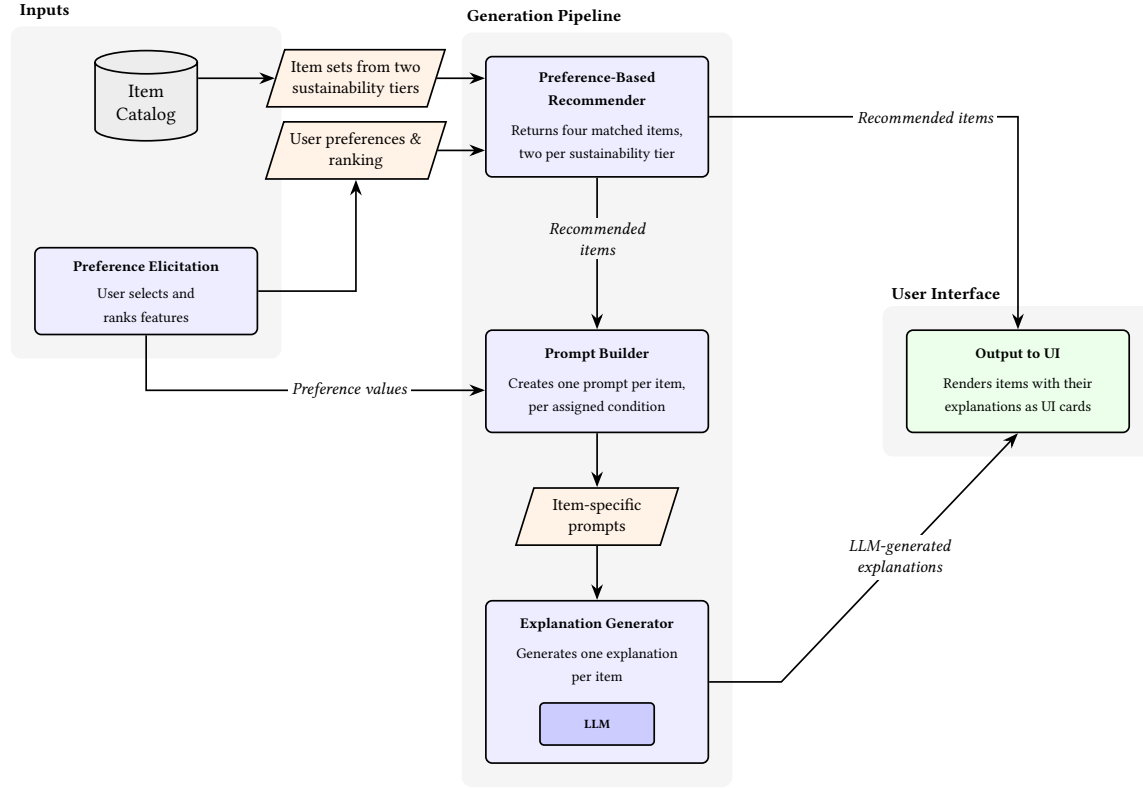
\begin{figure}[t]
\centering
\resizebox{\textwidth}{!}{\begin{tikzpicture}[
  font=\normalsize,
  comp/.style={draw, thick, rounded corners=3pt, align=center, fill=blue!7,
               minimum width=3.7cm, minimum height=1.7cm, inner sep=6pt},
  data/.style={draw, thick, align=center, fill=orange!10, trapezium,
               trapezium left angle=70, trapezium right angle=110,
               trapezium stretches body, minimum width=2.7cm,
               minimum height=0.95cm, inner sep=3pt, font=\small},
  store/.style={draw, thick, cylinder, shape aspect=0.32, aspect=0.32, shape border rotate=90, fill=gray!14, align=center, minimum width=1.7cm, minimum height=1.5cm, inner sep=3pt},
  flow/.style={-{Stealth[length=2.6mm]}, thick},
  lbl/.style ={font=\small\itshape, align=center, fill=white,   inner sep=1.8pt}, % on white background (outside panels)
lblin/.style={font=\small\itshape, align=center, fill=gray!8, inner   sep=1.8pt}, % on grey background (inside a panel)
  layer/.style={rounded corners=5pt, inner sep=11pt},
]

% ===================== components =====================
% --- inputs (left column) ---
\node[store]                        (db)     at (0, 1.1)     
{Item\\Catalog};

\node[comp, minimum height=1.3cm]   (elicit) at (0,-2)   
{\ctitle{Preference Elicitation}\\[2pt]
\cdesc{User selects and}\\
\cdesc{ranks features}};

% --- generation spine (x=7.5) ---
\node[comp, minimum height=2.0cm]   (rec)    at (7.5, 0.9)  
{\ctitle{Preference-Based}\\
\ctitle{Recommender}\\[3pt]
\cdesc{Returns four matched items,}\\
\cdesc{two per sustainability tier}};

\node[comp]                         (prompt) at (7.5,-3.5)   
{\ctitle{Prompt Builder}\\[3pt]
\cdesc{Creates one prompt per item,}\\
\cdesc{per assigned condition}};

% explanation generator
% --- explanation generator: LLM is the engine, wrapped by a titled container ---

\node[comp, minimum height=2.7cm]   (gen)    at (7.5,-8.5)   {};

\node[align=center, anchor=north] at ($(gen.north)+(0,-0.18)$)
{
\ctitle{Explanation Generator}\\[3pt]
\cdesc{Generates one explanation}\\
\cdesc{per item}
};

\node[draw, thick, rounded corners=2pt, fill=blue!20,
      font=\scriptsize\bfseries,
      minimum width=1.9cm, minimum height=0.7cm,
      anchor=south]
      at ($(gen.south)+(0,0.30)$) (llm) 
      {LLM};

% --- output (right) ---
\node[comp, fill=green!8] (ui) at (14.5,-3.5)
{
\ctitle{Output to UI}\\[3pt]
\cdesc{Renders items with their}\\
\cdesc{explanations as UI cards}
};

% ===================== data artifacts =====================
\node[data] (pools)  
at ($(db.east)!0.55!(rec.west)+(0,0.55)$)  
{Item sets from two \\ sustainability tiers};

\node[data] (prefs)  
at ($(db.east)!0.55!(rec.west)+(0,-0.65)$) 
{User preferences \& \\ ranking};

\node[data] (prm)    
at ($(prompt.south)!0.5!(gen.north)$)
{Item-specific\\prompts};

% ===================== flows =====================
% catalog -> pools -> recommender 

\draw[flow] (pools.east) -- ([yshift=0pt]rec.west |- pools.east);

% elicitation -> preferences -> recommender
\draw[flow] (elicit.east) -| ($(elicit.east)+(1.2,0)$) -| (prefs.south);

% inputs -> recommender
\draw[flow] (db.east |- pools.west) -- (pools.west);

\draw[flow] (prefs.east) -- (prefs.east -| rec.west);

% recommender -> prompt (main vertical branch down)
\draw[flow] (rec.south) -- (prompt.north) node[lblin, pos=0.4] {Recommended\\items};

% preferences also feed the prompt builder (parallel path)
\draw[flow] (elicit.south) |- ($(prompt.west)+(0,-0.15)$)
      node[lbl, pos=0.8] {Preference values};

% prompt -> generator (through the prompts artifact)
\draw[flow] (prompt.south) -- (prm.north);
\draw[flow] (prm.south)    -- (gen.north);

% return branch to UI
\draw[flow] (gen.east) -| ($(gen.east)+(1.2,0)$) -- (ui.south) node[lbl, , pos=0.5] {LLM-generated\\explanations};
\draw[flow] (rec.east) -| (ui.north) node[lbl, pos=0.35] {Recommended items};

% ===================== layer backdrops =====================
\begin{scope}[on background layer]
  \node[layer, fill=gray!8,  fit=(db)(elicit),
        label={[font=\bfseries\small]above, xshift=-48pt:Inputs}] {};
  \node[layer, fill=gray!8,  fit=(rec)(prompt)(gen)(llm)(prm),
        label={[font=\bfseries\small]above, xshift=-25pt:Generation Pipeline}] {};
  \node[layer, fill=gray!8, fit=(ui),
        label={[font=\bfseries\small]above, xshift=-34pt:User Interface}] {};
\end{scope}

\end{tikzpicture}}
\caption{System architecture, grouped into three layers (Inputs, Generation Pipeline, User Interface). Purple rounded boxes represent the active components (Preference Elicitation, Preference-Based
Recommender, Prompt Builder, Explanation Generator, Output to UI).
Orange trapezoids are the named data artifacts exchanged between components. Arrows show the direction of data flow. For each participant, the recommender returns a balanced item set, which flows through prompt construction and LLM-based explanation generation. The items and their explanations are then returned to the user interface.}
\Description{Architecture diagram organized into three layers: Inputs, Generation Pipeline, and User Interface. In Inputs, a Preference Elicitation component and an item catalog feed user preferences and two tiered item sets into the Generation Pipeline. There, a Preference-Based Recommender returns four matched items to a Prompt Builder, which creates one prompt per item and condition; these prompts pass to an Explanation Generator powered by an LLM. The generated explanations and the recommended items then flow to the Output to UI component in the User Interface layer, which renders them to the participant. Rounded boxes denote active components, trapezoids denote data artifacts, and arrows denote the direction of data flow.}
\label{fig:system_architecture}
\end{figure}

Figure~\ref{fig:system_architecture} shows the technical pipeline for delivering the recommendations and their explanations in both user studies, implemented as a self-hosted web application, which participants accessed through their browser during the study. The backend comprises a preference-based recommender that selects the four recommended items for each participant (Section~\ref{recommender}), and an LLM-based generator that produces the accompanying explanations under the condition-specific prompts (Section~\ref{generating}). The frontend then renders the recommended items and their explanations to the participant.
Since the recommender and generator are domain-agnostic, we first describe these components before instantiating the pipeline in the two involvement domains. For each domain, we specify the dataset, features, and sustainability signal, then illustrate the resulting explanations with concrete examples (Section~\ref{instantiation}). Finally, we present a validation stage that evaluates the generated explanations across multiple dimensions (Section~\ref{sec:stimulus-validation}).

\subsection{Recommendation Process}
\label{recommender}

The recommender follows a preference-based approach in the tradition of knowledge-based and constraint-based RS \cite{jannach2010recommender,felfernig2008constraint}, in which items are scored directly against users' explicitly stated preferences. Such approaches were prominent in earlier days of the field, for instance in critiquing-based recommenders \cite{chen2012critiquing}, and have regained relevance with LLM-based conversational RS, which again elicit users' stated needs rather than inferring them from historical data \cite{gao2021advances,jannach2021survey}. Our choice of this paradigm is motivated by two considerations. First, it enables explanations that directly reference users' stated requirements, thereby supporting explanation goals such as persuasiveness \cite{tintarev2007explanations,tintarev2007survey} and user satisfaction \cite{bilgic2005explaining}. Second, alternative approaches such as collaborative filtering typically depend on historical interactions or implicitly inferred preferences \cite{lubos2024llm}, which in our setting would require exposing participants to items before the choice task and risk introducing priming or learning effects.

The central requirement of our design is that all four recommended items fit the participant's stated preferences comparably well. If, say, the higher-sustainability hotels matched a participant's wishes better than the lower-sustainability ones, preference fit rather than the explanations could drive choices, thus confounding the results. The recommender meets this requirement in two steps, which we illustrate with a participant in the high-involvement study who selected pool, Wi-Fi, and restaurant as desired facilities and ranked them in that order of importance.

\emph{Step 1: Scoring within each tier.} Working separately on the higher- and the lower-sustainability pool (Section~\ref{instantiation}), the recommender scores every item by how many of the participant's preferred feature values it matches, weighting each match by the feature's importance rank:
\begin{equation}
\mathrm{score}(i \mid u)
=
\sum_{f \in F} w_f \cdot \mathbb{I}\!\left[x_{i,f} = u_f\right],
\end{equation}
where $F$ is the set of features for which the user provided a preference, $u_f$ is the user's preferred value for feature $f$, $x_{i,f}$ is item $i$'s value for that feature, $\mathbb{I}$ indicates a match, and $w_f$ is a rank-based weight that decreases linearly with the importance ranking \cite{stillwell1981comparison,barron1996decision}. For our example participant, a hotel offering a pool and Wi-Fi but no restaurant thus outscores one offering only Wi-Fi and a restaurant, because the pool was ranked most important.\footnote{The $k$-th ranked feature receives weight $\max(0.2,\,1.0-0.2(k-1))$, and selected but unranked features receive a weight of $0.1$. The linear decrement follows the rank-based weighting schemes of multi-attribute decision analysis \cite{stillwell1981comparison}. The step size of $0.2$ spreads up to five ranked features evenly across the weight range, and since participants may select up to six features but rank as few as three, the floor of $0.2$ and the default of $0.1$ keep every stated preference influential. Items matching no ranked feature (score below $\tau=1$) are discarded, and the top $K=30$ candidates per pool are retained to provide a large enough buffer for the comparability search in Step 2.}

\emph{Step 2: Enforcing comparability across tiers.} High within-pool scores alone do not guarantee an unconfounded comparison, because the best-matching item in each pool may earn its score through different features. Suppose the best hotel from the higher-sustainability tier offers the pool and Wi-Fi, while the best hotel from the lower-sustainability tier offers only Wi-Fi and a restaurant, because no hotel with a pool exists among the top candidates in that tier. Each is the strongest match within its own tier, yet the participant, having ranked the pool as most important, may simply pick the hotel from the higher-sustainability tier for its pool. 
To rule out such asymmetries, we require all four recommended items to match the participant's preferred values on the same set of their most important ranked features. This also protects the explanation manipulation. Because all four items match the participant on the same features, every explanation within a set cites an equally strong preference match, so no explanation is more persuasive merely through the fit it can claim.

\subsection{Generating Explanations}
\label{generating}

As argued in Sections~\ref{sec:intro} and~\ref{explanations}, LLMs make our intervention feasible at scale because nudging through explanation content requires a distinct, item-user-specific text for every recommendation. Our generation approach is model-agnostic and requires only an instruction-following LLM. For the main studies, we generated the explanations with Claude Sonnet~4.6, a state-of-the-art large language model available at the time of the study.\footnote{\url{https://www.anthropic.com/claude/sonnet}} 
We selected this model after pilot testing because it consistently adhered to the structural constraints described below while producing fluent explanations across the large number of generated stimuli.
The choice of generator model is an implementation decision rather than a precondition for the intervention and the prompting strategy transfers to other LLMs, whether accessed through commercial APIs or run locally as open-source models. To facilitate replication, the accompanying repository includes scripts for generating the explanations with multiple LLMs.
Concretely, the three sustainability-aware conditions share the prompt below and differ only in their second sentence (Table~\ref{tab:sentence2}).

\begin{tcolorbox}[
    enhanced, breakable, colback=white, colframe=black, boxrule=0.5pt, arc=3mm,
    left=4mm,right=4mm,top=4mm,bottom=4mm,
    title=Explanation-generation prompt (sustainability-aware conditions),
    attach boxed title to top left={xshift=4mm, yshift*=-\tcboxedtitleheight/2},
    boxed title style={colback=white, size=minimal, frame hidden},
    fonttitle=\bfseries\color{black},
]
Your task is to explain \textcolor{blue}{\{item domain\}} recommendations based on overlap between user preferences and item attributes as well as sustainability aspects.\\
The explanation must consist of exactly two sentences, using no more than 50 words total.\\

Sentence 1: Explain why this item is suitable for the user using only the overlapping user preferences and item attributes of the features: \textcolor{blue}{\{overlapping features\}}.\\
Sentence 2: \textit{[condition-specific; see Table~\ref{tab:sentence2}]}\\

- Vary the sentence order naturally across the explanation for fluency. Place ``also'' in whichever sentence comes second.\\
- Keep the explanation fluent and in natural language.\\
- Convert all structured attributes into natural phrasing.\\
- Avoid repeating the same nouns, adjectives, or attribute values across both sentences.\\

User preferences: \textcolor{blue}{\{user preferences\}}.\\
Item attributes: \textcolor{blue}{\{item attributes\}}.
\end{tcolorbox}

The preference-only baseline uses the same scaffold but omits the sustainability 
aspect: its first sentence instead \textbf{describes} the item (\textit{``Describe this 
item using only the features: \textcolor{blue}{\{overlapping features\}}''}), and 
its second sentence states the item's suitability.

\begin{table}[ht]
  \centering\small
  \caption{Second sentence of the shared prompt for each condition, reproduced 
  verbatim. \textcolor{blue}{Blue} marks runtime-filled slots.}
  \label{tab:sentence2}
  \begin{tabular}{p{0.24\textwidth} p{0.66\textwidth}}
    \toprule
    \textbf{Condition} & \textbf{Sentence 2} \\
    \midrule\midrule
    Preference-only & State that this item is suitable for the user based on overlapping user preferences with item attributes of the mentioned features. \\
    \midrule
    Mechanism-free & State that this item is recommended to the user due to the sustainability aspect of the item \textcolor{blue}{\{sustainability attribute\}} and name it plainly, without evaluative language and without explaining it. \\
    \midrule
    Framing \& Descriptive social norms & State that this item is recommended to the user due to the sustainability aspect of the item \textcolor{blue}{\{sustainability attribute\}} and express it according to the \textcolor{blue}{\{mechanism name\}} mechanism which \textcolor{blue}{\{mechanism description\}}.  \\
    \bottomrule
  \end{tabular}
\end{table}

We realized framing through a positive, gain-oriented frame that foregrounds the beneficial consequence of the sustainability attribute. While the academic discussion on the relative efficacy of gain versus loss framing \cite{tversky1981framing} in environmental communication is mixed \cite{homar2021effects}, some evidence indicates that gain frames elicit more favorable responses than loss frames for consumer products \cite{segev2015effects}.
For descriptive social norms, a central design decision concerns the peer group, that is, the reference group of other people whose behavior the norm statement describes. Following the conceptualization of descriptive norms as information about the behavior of similar others \cite{cialdini2004social}, we defined the peer group by the decision context (i.e., purchasing, booking) rather than by environmental values (e.g., green customers), to ensure a situationally relevant reference group. This is consistent with evidence that descriptive norms are most effective when they refer to behavior in a comparable decision context, rather than value-based groupings that may introduce identity-related or injunctive connotations \cite{goldstein2008room}.

The concrete instantiation of both mechanisms per domain follows in Section~\ref{study:mechanisms}, but as a brief preview, consider a coffee item whose verified sustainability attribute is recyclable packaging. The slot \textcolor{blue}{\{sustainability 
attribute\}} is then filled with ``recyclable packaging,'' and \textcolor{blue}{\{mechanism description\}} with the description of the assigned mechanism from Table~\ref{tab:mechanisms}. For the framing condition, the prompt's second sentence thus reads: \emph{``State that this item is also recommended for its recyclable packaging, and express it according to the framing mechanism, which means rephrasing the sustainability attribute to foreground its 
positive consequence or benefit~[\ldots]''}.

Because our conditions differ only in what the explanation says about sustainability, any systematic difference in tone, length, or linguistic complexity across conditions would confound the manipulation. Our generation procedure is therefore designed to hold the linguistic form of the explanations constant while varying only the manipulated content, so that differences across conditions can be attributed to the explanation content rather than to linguistic variation.
To this end, we constrained the model's outputs with a fixed structure and style constraints, applied identically across both involvement domains. Every explanation consisted of exactly two sentences and at most 50 words, a length settled on in pilot testing as sufficient to convey the rationale while remaining short enough to avoid overly long explanations \cite{hendrawan2024explanations}. To further prevent templated phrasing across the many stimuli, we included the explanations previously generated for the same recommendation set in the model's 
context, with the instruction not to reuse their phrasing. Without this, explanations for different items tend to collapse onto a fixed template (for 
instance, every explanation opening \emph{``This item suits you well because it~\ldots''}), while including the prior explanations pushes the model to vary sentence structure and word choice across the items.

\subsection{Domain Instantiation}
\label{instantiation}

This section describes the shared design of Section~\ref{sec:experimental-design} in the two involvement domains. As introduced in Section~\ref{sec:item-domains}, we selected instant coffee as a prototypical low-involvement product, given its habitual, low-cost, and low-risk purchase character \cite{kapferer1985consumer}, 
and hotel bookings as prototypical for the high-involvement domain, since it entails infrequent, financially consequential decisions \cite{lisha2025service}.
All datasets described below are part of our publicly shared material.
We note that the resulting datasets for each domain after pre-processing differ in their size, with $194$ items for hotels and $92$ for instant coffee. We used the largest sample available in each domain to maximize statistical power rather than fixing a common size, which would have required discarding usable, well-matched items in the hotel study \citep{cohen2013statistical}.

\subsubsection{Dataset Low-Involvement Domain:}\label{study1:dataset}\mbox{}\\[-.9\baselineskip]
\\
\textit{Item catalog.} We built the item catalog from product metadata in the Amazon Reviews corpus \cite{hou2026bridging}, restricted to the Grocery and Gourmet Food category. We retained items that mentioned instant coffee in the title or description, removed duplicates (by product identifier and primary image), and for each item kept its textual metadata (title, description, features, and details).

\textit{Feature pre-processing.} The recommender requires every item to be described by the same fixed set of features, each with a predefined set of possible values, so that stated preferences can be matched against item attributes. Since the Amazon metadata is unstructured in nature, we defined a schema of categorical features and mapped each item's text onto it using GPT-5.2\footnote{\url{https://openai.com/gpt-5}} in an LLM-based annotation step, an approach shown to match or exceed crowd annotation for structured text-labeling tasks \cite{gilardi2023chatgpt}. For every feature, the model returned a value drawn from a predefined set of possible feature values along with a flag indicating whether the value was stated explicitly or inferred. Values directly supported by the text were accompanied by an exact substring (e.g., the value \emph{dark} for roast level backed by the phrase ``dark roast'' in the description), and values inferred from indirect cues (e.g., \emph{dark} inferred from ``French roast'') were marked as such. To ensure the reliability of this step, all outputs were manually reviewed for correctness. To keep the preference elicitation task and the explanations concise, we restricted preferences to the four features most consistently available in the item metadata and most characteristic of instant coffee products, namely roast level, bean type, caffeine, and intensity. During preference elicitation, participants selected their preferred value for at least three of the four features and then ranked them by importance.

\textit{Sustainability tiers.} Motivated by SDG~12 (Responsible Consumption and Production), we operationalized environmental sustainability through explicit packaging-related claims \cite{norton2022exploring}. An item was labeled higher-sustainability when its metadata contained a packaging cue such as recyclable, recycled, post-consumer recycled, compostable, biodegradable, plastic-free, or otherwise sustainable or eco-friendly packaging \cite{bocoli2025perspectives}. Such attributes are recognized characteristics of sustainable packaging and are read by consumers as signaling a product's environmental friendliness \cite{norton2022exploring}, while remaining present in the metadata, verifiable, and readily interpretable by users. Like any single indicator, however, a packaging claim serves as a proxy for environmental sustainability rather than a comprehensive assessment.
We then constructed the lower-sustainability pool as a matched comparison set. The goal of this matching was to obtain two pools that mirror each other on the recommendation features, so that the tiers differ systematically only in the sustainability attribute. The matching proceeded in two stages. First, for each higher-sustainability item we retrieved its five nearest neighbors from the candidate pool by cosine similarity over dense sentence embeddings of the concatenated textual metadata, using a Sentence-Transformers model \cite{reimers2019sentence}, and mapped these candidates onto the same feature schema. Second, from each item's neighbors we selected exactly one match that maximized the number of agreeing specified attributes, breaking ties by semantic similarity and then by neighbor rank. Matching was performed so that no candidate item served as the match for more than one higher-sustainability item, thus keeping the recommended item sets distinct and non-overlapping. This produced two equally sized, individually paired pools of 46 items per sustainability tier.

\textit{Attribute balance.} A key requirement of our design is that the recommendations differ only in the explanations we manipulate, and not in product features such as roast level. If items differed on such features, participants' choices could reflect those feature differences  rather than the explanations, confounding the results. To confirm that the  matching procedure produced comparable item pools, we therefore tested the balance of the item features across both pools. For each feature we used a  chi-square test of independence, together with Cram\'er's $V$ as a measure of effect size. The chi-square test is appropriate here because all minimum expected cell counts exceeded five, the conventional threshold for the validity of the test's large-sample approximation \citep{cochran1954}. All associations were small ($V \leq 0.151$) and none was statistically significant ($p \geq 0.350$), indicating that the pools were highly comparable and that any differences in participants' choices can not be attributed to feature imbalance or to differences in how well the recommended items met user preferences.
The concrete attributes and test are shown in Appendix~\ref{app:balance_low}.

\subsubsection{Dataset High-Involvement Domain:}
\label{study2:dataset}\mbox{}\\[-.9\baselineskip]
\\
\textit{Item catalog.} We built the hotel catalog by scraping the official Vienna tourism portal,\footnote{\url{https://www.wien.info/en/travel-info/hotels-accomodations}} which lists accommodations in Vienna alogside structured facility information and, where applicable, official sustainability certifications. For 
each hotel we collected the attributes the portal exposes, including its facilities, its distance to the city center, a representative image, and any sustainability certification shown on its page. The attribute set is therefore determined by what the source natively provides for every listing. %, and not by a discretionary selection on our part.

\textit{Feature pre-processing.} Each hotel was described by a set of binary facility attributes (for example, Wi-Fi, restaurant, bar, sauna, parking, and wheelchair accessibility) along with a coarse indicator of distance to the city center. These attributes were taken directly from the structured catalog, so no language-model extraction was required. During preference elicitation, participants selected up to six facilities they would like having in hotels and ranked at least three of them by importance. The recommender scored hotels on this elicited subset.

\textit{Sustainability tiers.} Motivated by SDG~13 (Climate Action), we operationalized environmental sustainability through third-party sustainability certifications, such as the EU Ecolabel, the Austrian Ecolabel, Green Key, and Green Globe, which serve as independently verified signals that consumers interpret as evidence of a hotel's environmental commitment \cite{martinez2018customer}. A hotel was labeled higher-sustainability when its page listed at least one such certification. %As in the low-involvement domain, a certification serves as a proxy for environmental sustainability rather than a comprehensive assessment of a hotel's footprint.
We constructed the lower-sustainability pool as a matched comparison set so that the two tiers resemble each other on the facility attributes. Each certified hotel was paired with exactly one uncertified hotel that maximized the number of agreeing attributes, breaking ties in favor of the closest match. Matching was again performed without replacement, so that no uncertified hotel served as the match for more than one certified hotel. Because this dataset offered 
structured attributes rather than free text, matching was performed directly on the facility attributes without a separate semantic-similarity step. This resulted in two equally sized, individually paired pools of 97 hotels per sustainability tier.

\textit{Attribute balance.} As described earlier, a key requirement is that the two item pools do not differ in any facility attributes, because otherwise participants' choices could reflect those  attribute differences instead of differences in manipulation. To confirm that the matching  procedure produced comparable pools, we tested the balance of the facility attributes (e.g. Wi-Fi) across the two item pools using chi-square tests of independence along with Cram\'er's $V$ as a measure of effect size. For one rarely offered facility (elevator), the minimum expected cell count fell below five, so the chi-square approximation was no longer valid. For this attribute we instead used Fisher's exact test \citep{fisher1922interpretation}. All associations were small ($V \leq 0.085$) and none was statistically significant ($p \geq 0.238$), indicating that the pools were highly comparable and that any differences in participants' choices do not reflect attribute imbalance or differences in how well the recommended items met user preferences.
The concrete attributes and test are shown in Appendix~\ref{app:balance_high}.

\subsubsection{Mechanism Wording and Example Explanations:}
\label{study:mechanisms}\mbox{}\\[-.9\baselineskip]
\\
We now make the intervention concrete for both domains. Table~\ref{tab:mechanisms} shows how the two decision information mechanisms were instantiated as prompt instructions in each domain, with the sustainability attribute referring to the item's packaging claim in the low-involvement domain (instant coffee) and to the hotel's certification in the high-involvement domain. Tables~\ref{tab:example_low} and~\ref{tab:example_high} show the explanations generated for one higher-sustainability item in each domain under each of the four conditions and for the same preferences. Figures~\ref{fig:example_low} and~\ref{fig:example_high} illustrate the preference elicitation and choice task as participants experienced it in the two domains.

To prevent confounds from real brand identities, we did not display the items' original brand names, packaging photographs, prices, or ratings, any of which 
could carry their own quality or familiarity signals that are known to bias user-centric recommender evaluations \cite{jannach2015item}. Instead, each recommended item was shown under a neutral, fictional brand name and a generic image drawn from a small fixed set, assigned at random and independently for each 
participant. The only systematic information distinguishing items was therefore their elicited feature values and the accompanying explanation, ensuring that differences in choice across conditions stemmed from the explanation manipulation rather than from incidental branding cues.

\begin{table}[ht]
  \centering
  \caption{Prompt instantiation of the two decision information mechanisms 
  \cite{munscher2016review,jesse2021digital} across both domains. Each  \textit{mechanism description} is inserted into the mechanism-based prompt  (Section~\ref{generating}) to govern how the verified sustainability attribute is expressed. The \textit{Translate information (Framing)}  instruction is domain-independent, while the \textit{Provide a social reference point (Descriptive 
  social norms) } instruction is adapted to the item type of each domain.}
  \label{tab:mechanisms}
  \begin{tabular}{p{0.12\textwidth} p{0.12\textwidth} p{0.12\textwidth} p{0.44\textwidth}}
    \toprule
    \textbf{Subtype} & \textbf{Mechanism} & \textbf{Domain} & \textbf{Mechanism description (prompt instruction)} \\
    \midrule
    \midrule
    \makecell[tl]{Translate\\information} & Framing & \multirow{2}{*}{\makecell[tl]{Both\\domains}} &
    \textit{means rephrasing the sustainability attribute to foreground its 
    positive consequence or benefit. Do not simply say the item ``has'' the 
    attribute. Instead, restate what that attribute means for the world, the 
    user, or the future.} \\
    \midrule
    \multirow{2}{*}{\makecell[tl]{Provide a social\\reference point}} &
    \multirow{2}{*}{\makecell[tl]{Descriptive\\social norms}} &
    \makecell[tl]{Low\\involvement\\(instant \\coffee)} &
    \textit{means giving a concrete behavioral observation about what similar 
    customers actually did regarding the sustainability attribute. Define the 
    peer group by context only (e.g., customers browsing or looking for instant 
    coffee), never by values (e.g., eco-conscious, sustainability-minded, or 
    green customers).} \\
    \cmidrule{3-4}
    & & \makecell[tl]{High\\involvement\\(hotel)} &
    \textit{means giving a concrete behavioral observation about what similar 
    guests actually did regarding the sustainability certification. Define the 
    peer group by context only (e.g., guests at this hotel, travelers booking in 
    this area), never by values (e.g., eco-conscious, sustainability-minded, or 
    green traveler).} \\
    \bottomrule
  \end{tabular}
\end{table}

\begin{table}[ht]
  \centering
  \caption{Example explanations for one higher-sustainability item in the low-involvement domain (instant coffee). The preference match content is held constant (caffeinated, arabica beans, dark roast level, bold intensity). The sentence carrying the sustainability content is set in bold.}
  \label{tab:example_low}
  \begin{tabular}{p{0.26\textwidth} p{0.66\textwidth}}
    \toprule
    \textbf{Condition} & \textbf{LLM-generated explanation} \\
    \midrule
    \midrule
    \makecell[tl]{Preference-based} &
    \emph{Crafted from arabica beans, this caffeinated coffee offers a dark roast with a boldly intense character. Given your taste for strong, dark arabica with caffeine, this item lines up perfectly with every preference you've shared.} \\
    \midrule
    \makecell[tl]{Sustainability-aware\\(Mechanism-free)} &
    \emph{A caffeinated, dark-roasted arabica with bold intensity, it aligns perfectly with everything you enjoy in a cup. \textbf{This item is also recommended because it uses post-consumer recycled packaging.}} \\
    \midrule
    \makecell[tl]{Sustainability-aware\\(Framing)} &
    \emph{With its caffeinated, dark-roasted arabica beans and bold intensity, this coffee aligns seamlessly with your taste profile. \textbf{By choosing this item, you also help reduce landfill waste, as its recyclable packaging gives materials a second life and supports a cleaner future.}} \\
    \midrule
    \makecell[tl]{Sustainability-aware\\(Descriptive\\social norms)} &
    \emph{This dark-roasted arabica delivers the bold, caffeinated kick you're after. \textbf{Customers browsing similar full-bodied blends have also widely chosen this product for its recyclable packaging, making it a popular pick in that shopping context.}} \\
    \bottomrule
  \end{tabular}
\end{table}

\begin{table}[ht]
  \centering
  \caption{Example explanations for one higher-sustainability item in the high-involvement domain (hotel bookings). The preference match content is held constant (Wi-Fi, close to city center, restaurant, pool). The sentence carrying the sustainability content is set in bold.}
  \label{tab:example_high}
  \begin{tabular}{p{0.26\textwidth} p{0.66\textwidth}}
    \toprule
    \textbf{Condition} & \textbf{LLM-generated explanation} \\
    \midrule
    \midrule
    \makecell[tl]{Preference-based} &
    \emph{Nestled close to the city center, this hotel boasts complimentary Wi-Fi, a refreshing pool, and an on-site restaurant. Based on your interest in connectivity, dining, swimming, and a central location, this property aligns perfectly with everything you're looking for.} \\
    \midrule
    \makecell[tl]{Sustainability-aware\\(Mechanism-free)} &
    \emph{With Wi-Fi, a restaurant, a pool, and a location close to the city center, this property aligns perfectly with your preferences. \textbf{It is also recommended for its Green Key certification.}} \\
    \midrule
    \makecell[tl]{Sustainability-aware\\(Framing)} &
    \emph{This hotel is ideally suited for you, offering Wi-Fi, an on-site restaurant, a pool, and a convenient location close to the city center. \textbf{It is also recommended because its Green Key certification means it actively reduces environmental harm, helping preserve the planet for generations to come.}} \\
    \midrule
    \makecell[tl]{Sustainability-aware\\(Descriptive\\social norms)} &
    \emph{\textbf{Many guests booking in this area have also chosen this hotel for its Green Key certification, reflecting a growing trend among travelers who prioritize certified sustainable stays.} It offers Wi-Fi, an on-site restaurant, a pool, and is conveniently located close to the city center.} \\
    \bottomrule
  \end{tabular}
\end{table}

\subsection{Stimulus Validation}
\label{sec:stimulus-validation}

Since the explanations are generated on-the-fly during the experiment, the exact stimulus set presented to participants is only determined after data collection is complete. Through prompt development and pilot testing 
(Sections~\ref{sec:experimental-design} and~\ref{generating}), we established that the generation procedure resulted in explanations aligned with the intended design. 
Because LLM generation is nonetheless stochastic, individual explanations may still deviate, and such deviations could easily go unnoticed if the outputs were taken at face value. We therefore audited the realized stimulus set---on which our 
behavioral results rely---along several dimensions before interpreting any outcomes.

The audit combined three complementary analyses. First, we examined statistical properties of the explanation texts (sentence count, length, lexical complexity, and textual separability) to establish that the conditions did not differ in incidental form and were textually distinguishable.
Second, we applied an LLM-as-a-judge approach to assess whether each explanation instantiated its intended mechanism and whether its sustainability claim was factually supported by the item's verified attribute. 
Third, to confirm these automated judgments, two researchers independently re-labeled a stratified subset of explanations by hand without access to condition labels or the LLM judge's verdicts. 
A screening of the generated texts prior to analysis identified six explanations containing residual generation artifacts (e.g., a leaked ``Here is a new explanation: \ldots'' preamble). Because these artifacts could have affected participants' exposure, the six affected participants were excluded from all analyses (final $N = 529$), and their explanations are omitted from the analysis reported below.

\subsubsection{Statistical Analysis of Text Properties:}
\label{val:stats}\mbox{}\\[-.9\baselineskip]
\\
\textit{Structural Form.} 
The structural constraints of Section~\ref{generating} held, as explanations consisted of almost exactly two sentences in every condition (means $1.97$--$2.00$; between-condition $\eta^2 = .021$), and average word length was near-identical across conditions, differing by at most $0.2$ characters per word ($M = 6.60$--$6.79$; $\eta^2 = .021$), indicating the conditions differ in length but not in lexical complexity. Word count did vary by condition ($\eta^2 = .35$): the two mechanism-based conditions were longest (framing $M = 37.2$, descriptive social norms $M = 36.7$ words), the mechanism-free condition shortest ($M = 29$), with the baseline in between ($M = 34.1$). This gap of roughly eight words between the longest and shortest condition is an inherent property of the manipulation, as expressing a benefit or a peer behavior takes more words than plainly naming an attribute. Adjusting for it in the main choice models would condition on a post-treatment variable and remove part of the manipulation itself \cite{montgomery2018conditioning,rosenbaum1984consequences}, thus treating a property of the manipulation improperly as a separable confound. Appendix~\ref{app:length} accordingly reports a robustness analysis confirming that explanation length itself does not predict choice.

\textit{Textual Distinctiveness.} 
Matched form guards against incidental confounds but to constitute distinct treatments the explanations must also differ in content---otherwise they 
would not represent separate experimental conditions.
Intuitively, explanations generated for the same condition (e.g. framing) should be more similar to one another than to explanations from other conditions (e.g. descriptive social norms).
To test this, we represent each explanation as a point in a high-dimensional sentence-embedding space using a Sentence-Transformers model \cite{reimers2019sentence}\footnote{Modelused for embeddings: \textit{all-mpnet-base-v2}}, where semantically similar explanations lie close together and dissimilar explanations lie farther apart. Under successful manipulation, explanations generated for the same condition should cluster together, while explanations from different conditions should occupy distinct regions of this space. We assess this in two complementary ways, that is, whether a classifier can assign explanations to their correct condition, and whether explanations are more similar within (same condition) than between conditions (different conditions).

\begin{figure}[t]
\centering
\includegraphics[width=\linewidth]{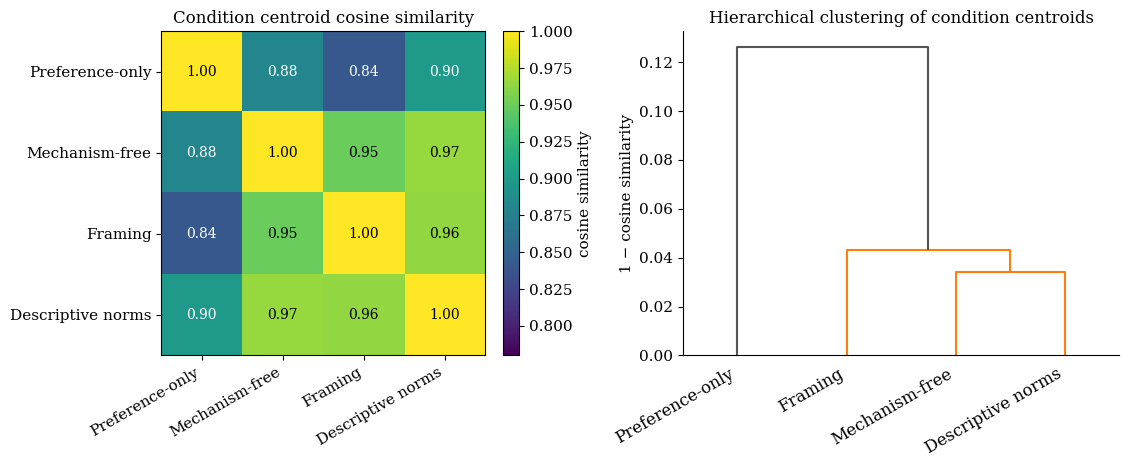}
\caption{Distinctiveness of the four explanation conditions in sentence-embedding space. Left: cosine similarity between condition centroids. Right: hierarchical clustering of the centroids. The preference-only baseline is the most distinct, lacking sustainability content, while the three sustainability-aware conditions share sustainability vocabulary yet remain reliably distinguishable.}
\Description{Two-panel figure showing the distinctiveness of the four explanation conditions in sentence-embedding space. The left panel is a heatmap of cosine similarity between the condition centroids, where the preference-only baseline shows the lowest similarity to the others and the three sustainability-aware conditions (mechanism-free, framing, descriptive social norms) are more similar to one another. The right panel is a dendrogram from hierarchical clustering in which the baseline branches off first and the three sustainability-aware conditions cluster together, yet remain separable.}
\label{fig:distinctiveness}
\end{figure}

A logistic-regression classifier over the embeddings, evaluated with five-fold cross-validation, recovered the condition of held-out explanations with near-perfect accuracy
(balanced accuracy $.99$, chance $= .25$; Cohen's $\kappa = .99$), with accuracy remaining near-ceiling within each domain ($\kappa = .99$ low-, $.97$ high-involvement).
Consistent with this result, the mean semantic similarity between explanations from the same condition exceeded that between explanations from different conditions beyond chance
(one-sided permutation $p < .001$).
Figure~\ref{fig:distinctiveness}  visualizes this semantic structure. Each condition is represented by its \emph{centroid}, which means the average embedding across all explanations in that condition and can be interpreted as its "typical" explanation. The left panel shows the cosine similarity between centroids, while the right panel applies hierarchical clustering, which successively merges the most similar centroids to reveal their relative proximity.
The preference-only baseline forms a clearly separate cluster as it lacks sustainability content altogether (centroid similarities $.84$--$.90$). In contrast, the three sustainability-aware conditions are more similar to one another because they share sustainability-related vocabulary, yet remain clearly distinguishable (centroid similarities $.95$--$.97$). We read this as evidence that the manipulation produced linguistically distinct conditions. Whether each treatment also instantiated its intended \emph{mechanism}, rather than merely differing in wording, is examined next.

\subsubsection{LLM-as-a-judge}
\label{val:judge}\mbox{}\\[-.9\baselineskip]
\\
A distinct text is not necessarily a correct one. Each explanation must also implement the mechanism prescribed by its condition and remain factually truthful. We therefore complement our lexical analysis with an LLM-based evaluation.
We opted for GPT-5.2 as a judge, a model from a different provider than the generator model, to avoid the documented tendency of LLM evaluators to favor outputs of their own model family \cite{panickssery2024llm}.  

($n = 790$),
\textit{Adherence to the intended presentation mechanism.}
We  fed all sustainability-aware explanations, blind to condition, to the LLM judge. The judge classified each explanation as containing no sustainability content, a mechanism-free explanation, a framing explanation, or a descriptive social norm explanation. Agreement with the intended mechanism was $99.4\%$ ($\kappa = .99$), with per-mechanism agreement of $1.00$ for framing, $.99$ for mechanism-free, and $1.00$ for descriptive social norms. Additionally, descriptive norm explanations defined the peer group by the intended decision context (e.g., ``customers browsing similar coffee'' or ``guests booking in this area'') as prescribed in Section~\ref{generating}. As a negative control, all preference-only baseline explanations were judged to contain no sustainability content, confirming that the judge did not over-attribute sustainability framing.

\textit{Factual fidelity.} A correctly framed explanation could still be factually wrong, and an invented environmental claim would turn the intervention into misinformation. We therefore audited whether every sustainability statement is entailed by the item's verified attribute (i.e., the certification for hotels and the manually reviewed packaging attribute for instant coffee). The LLM judge classified each of the sustainability-aware explanations as \emph{supported} (naming or restating the verified attribute), \emph{unsupported} (asserting a sustainability claim the attribute does not establish), or \emph{missing}. In total, $99.2\%$ of explanations were supported, and none contradicted the verified attribute. The six unsupported cases ($0.8\%$) all restated the verified attribute with more specific wording, such as turning ``recyclable packaging'' into ``will be recycled rather than ending up in landfill,''. We retained these cases because they did not contradict or otherwise violate the underlying factual claim.

\subsubsection{Human validation}
\label{val:human}\mbox{}\\[-.9\baselineskip]
\\
To ensure that the automated audit reflects human judgment, two independent researchers re-labeled a stratified subset of explanations, assigning each to a mechanism category and rating factual support, while remaining blind to condition and to the LLM judge's verdicts. For adherence to the intended mechanism, they re-labeled $79$ explanations. Inter-rater agreement was almost perfect (Human~A vs.\ Human~B $97.5\%$, $\kappa = .97$), as was agreement between each rater and the judge ($96.2\%$ and $98.7\%$, $\kappa = .95$ and $.98$). For factual fidelity, they re-labeled $59$ sustainability-aware explanations. Because the judgments were almost entirely one category (supported), Cohen's $\kappa$ is uninformative under such extreme skew, so we report raw agreement instead \cite{cicchetti1990high}. The two raters agreed on $98.3\%$ of explanations and agreed with the judge on $98.3\%$ and $100\%$. The automated full-set results above therefore rest on judgments that humans independently reproduce.

%%%%%%%%%%%%%%%%%%%%%%%%%%%%%%%%%%%%%%%%%%%%%%%%%%%%%%%%%%%%%%%%%%%%%%%%%%%%%%%%%%%%%%%%%%%%%%%%%

\section{Results}
\label{sec:results}
We organize the results around our three research questions. We first establish the 
validity of the experimental design by checking randomization balance and 
manipulation effectiveness (Section~\ref{sec:checks}), and then answer the three 
research questions in turn (Sections~\ref{sec:results-rq1}--\ref{sec:results-generalization}).

\subsection{Randomization and Manipulation Checks}
\label{sec:checks}

\textit{Randomization Check.} 
To verify that random assignment did not yield experimental groups differing systematically in background characteristics, we compared the four conditions on age, gender, ethnicity, consumption frequency, and domain expertise using chi-square tests of independence. No variable differed significantly across conditions, in the pooled sample (all $p \geq .22$) and within each involvement domain separately (low involvement all $p \geq .14$; high involvement all $p \geq .27$). Table~\ref{tab:demographics} reports the pooled demographic composition, and Appendix~\ref{app:randomization-per-domain} gives the per-domain balance tests.

\begin{table*}[h]
\centering
\caption{Participant characteristics by condition. Cell entries are counts with column percentages in parentheses. BASE = preference-only baseline, MF = mechanism-free, FR = framing, DSN = descriptive social norms.}
\label{tab:demographics}
\begin{tabular}{lccccc}
\hline
\textbf{Variable} & \textbf{Category} & \textbf{BASE ($n=134$)} & \textbf{MF ($n=131$)} & \textbf{FR ($n=134$)} & \textbf{DSN ($n=130$)} \\
\hline

\textbf{Age$^{a}$}
 & 18--24 & 29 (21.6\%) & 28 (21.4\%) & 32 (23.9\%) & 26 (20.0\%) \\
 & 25--34 & 52 (38.8\%) & 59 (45.0\%) & 53 (39.6\%) & 57 (43.8\%) \\
 & 35--44 & 30 (22.4\%) & 22 (16.8\%) & 24 (17.9\%) & 31 (23.8\%) \\
 & 45--54 & 15 (11.2\%) & 13 (9.9\%)  & 19 (14.2\%) & 11 (8.5\%) \\
 & 55+    & 8 (6.0\%)   & 9 (6.9\%)   & 6 (4.5\%)   & 5 (3.8\%) \\

\textbf{Gender}
 & Female & 67 (50.0\%) & 65 (49.6\%) & 68 (50.7\%) & 65 (50.0\%) \\
 & Male   & 67 (50.0\%) & 66 (50.4\%) & 66 (49.3\%) & 65 (50.0\%) \\

\textbf{Ethnicity$^{a}$}
 & Caucasian & 64 (47.8\%) & 49 (37.4\%) & 53 (39.6\%) & 59 (45.4\%) \\
 & Other     & 70 (52.2\%) & 82 (62.6\%) & 81 (60.4\%) & 71 (54.6\%) \\

\textbf{Consumption frequency$^{a}$}
 & Frequent   & 76 (56.7\%) & 84 (64.1\%) & 73 (54.5\%) & 84 (64.6\%) \\
 & Infrequent & 58 (43.3\%) & 47 (35.9\%) & 61 (45.5\%) & 46 (35.4\%) \\

\textbf{Domain expertise}
 & Novice        & 3 (2.2\%)   & 3 (2.3\%)   & 4 (3.0\%)   & 2 (1.5\%) \\
 & Knowledgeable & 96 (71.6\%) & 90 (68.7\%) & 96 (71.6\%) & 96 (73.8\%) \\
 & Advanced      & 28 (20.9\%) & 31 (23.7\%) & 28 (20.9\%) & 27 (20.8\%) \\
 & Expert        & 7 (5.2\%)   & 7 (5.3\%)   & 6 (4.5\%)   & 5 (3.8\%) \\

\hline
\multicolumn{6}{p{0.95\textwidth}}{\small
$^{a}$ Response categories were collapsed for analysis to ensure sufficient cell counts: age (55--64 and 65+ combined into 55+); ethnicity (Caucasian retained as the largest category, all remaining categories grouped as ``Other''); and consumption frequency (fine-grained response options collapsed into the broader categories shown).
}
\\
\hline
\end{tabular}
\end{table*}

\begin{figure*}[h]
\includegraphics[width=\textwidth]{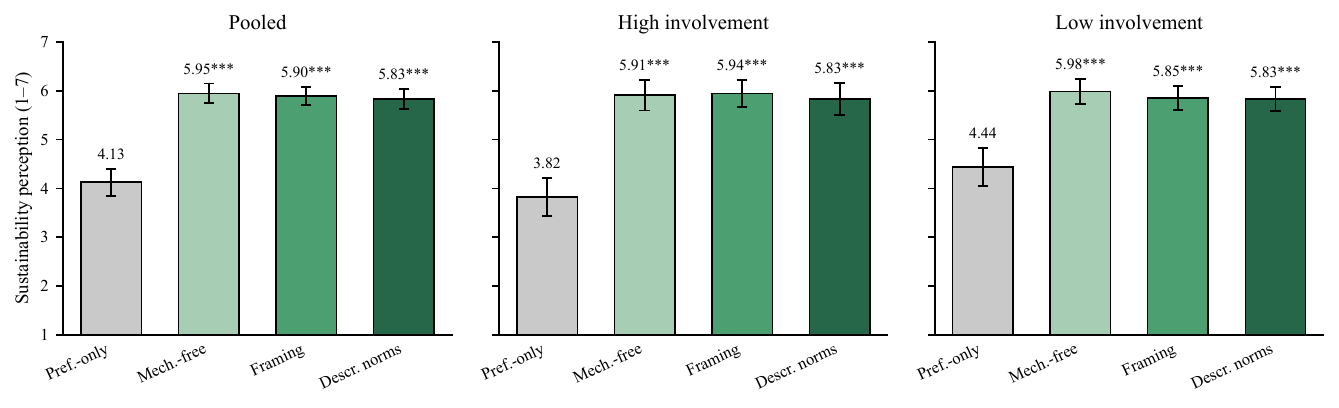}
\caption{Manipulation check: agreement with the manipulation check item by condition, pooled and per involvement domain. Error bars denote 95\% confidence intervals ($t$-based). Bars show raw condition means.}
\Description{Grouped bar chart of agreement with the manipulation-check item on a seven-point scale, by condition (preference-only baseline, mechanism-free, framing, descriptive social norms), shown pooled and separately for the low- and high-involvement domains. The three sustainability-aware conditions have substantially higher agreement than the preference-only baseline in every panel, with error bars showing 95\% confidence intervals.}
\label{fig:manipulation}
\end{figure*}

\textit{Manipulation Check.} 
A manipulation check verified that participants in the sustainability-aware conditions actually noticed the sustainability content, a precondition for it to influence their choices. Agreement with the manipulation-check item (\emph{``The explanations sometimes used sustainability considerations to justify the recommendations''}) was substantially higher in the three sustainability-aware conditions ($M = 5.83$--$5.95$) than in the preference-only baseline ($M = 4.13$). Each of the three sustainability-aware conditions individually exceeded the baseline (all two-tailed $p < .001$, all $d \geq 1.19$), and the pattern replicated within both domains. This suggests that the manipulation was successful and worked uniformly across the three sustainability-aware conditions. Figure~\ref{fig:manipulation} visualizes the results of the manipulation checks per condition and domain.

\FloatBarrier

\subsection{RQ1: Do Sustainability-Aware Explanations Shift Choice?}
\label{sec:results-rq1}

\begin{figure*}[h]\includegraphics[width=\textwidth]{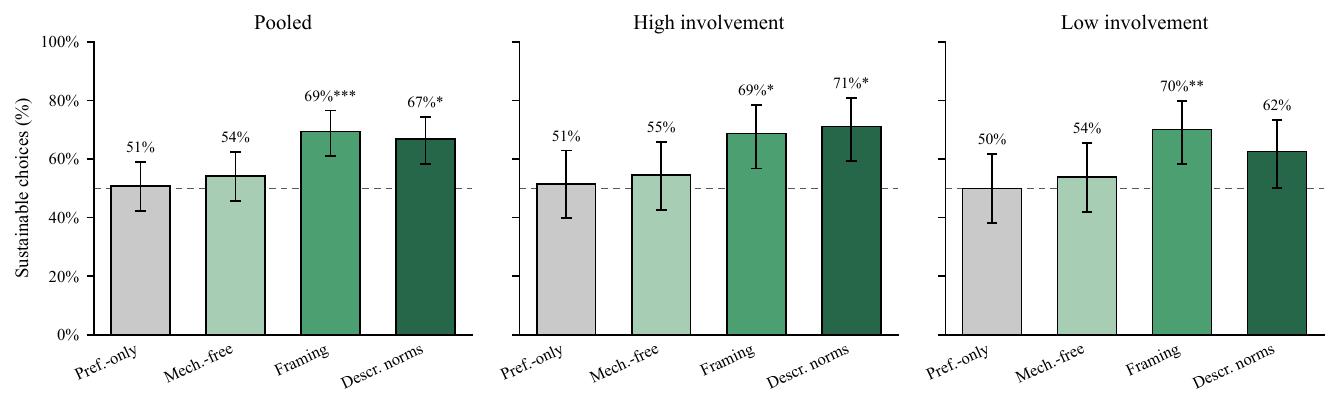}
\caption{Share of sustainable choices by condition, pooled and per involvement domain. Error bars denote 95\% Wilson confidence intervals; the dashed line marks the 50\% chance level implied by the balanced recommendation sets. Bars show raw condition means.}
\Description{Grouped bar chart showing the share of sustainable choices by condition (preference-only baseline, mechanism-free, framing, descriptive social norms), pooled and separately for the low- and high-involvement domains. The baseline and mechanism-free conditions sit near the 50 percent chance line, while framing and descriptive social norms rise well above it in every panel. Error bars show 95 percent Wilson confidence intervals and a dashed line marks the 50 percent chance level.}
\label{fig:choice}
\end{figure*}

Descriptively, Figure~\ref{fig:choice} shows the share of sustainable choices per condition. In the preference-only baseline, participants 
chose a higher-sustainability item at chance level in 51\% of cases, confirming that neither tier was intrinsically more attractive absent sustainability content. Plain disclosure in the mechanism-free condition increased this share marginally (54\%), whereas framing raised it to 69\% and descriptive social norms to 67\%.

To test whether these differences are statistically reliable and robust to individual differences, we model choice with binary logistic regression on the 
\emph{pooled} sample of both studies ($N = 529$), adjusting for participant covariates and item domain. We report estimates from the pooled model in Table~\ref{tab:choice-models} and formally test whether the condition effects differ by domain (RQ3) in Section~\ref{sec:results-generalization}. 
Collectively, the three sustainability-aware conditions 
raised the odds of a sustainable choice by 69\% over the preference-only baseline (Model~1: OR${}=1.69$, 95\% CI $[1.13, 2.52]$, $p=.011$). Biospheric value orientation independently predicted sustainable choice (OR${}=1.29$ per scale 
point, $p=.001$), yet the explanation effect held beyond it, indicating the intervention does not merely reach participants already predisposed to choose sustainably. \textbf{In answer to RQ1, sustainability-aware explanations, on average, shifted 
choices toward the sustainable option.}

\begin{table}[h]
\centering
\caption{Logistic regression of sustainable choice (pooled sample, $N = 529$). 
Entries are odds ratios with 95\% confidence intervals in brackets. Model~1 
addresses RQ1 (sustainability-aware conditions jointly vs.\ baseline); Models~2 and~3 address RQ2, using the preference-only baseline and the mechanism-free condition as reference, respectively, to isolate the two layers of the nested design. Sustainability-aware explanations raise the odds of a sustainable choice 
(Model~1), but only when the content is delivered through framing or descriptive social norms, while mechanism-free is statistically indistinguishable from the baseline (Models~2--3). All models adjust for the same covariates---age, gender, ethnicity, consumption frequency, domain expertise, item domain, and biospheric value orientation---but only biospheric value orientation is shown, as no other covariate approached significance in any model (all $p \geq .32$). $^{*}p<.05$, $^{**}p<.01$, $^{***}p<.001$.}
\label{tab:choice-models}
\begin{tabular*}{\textwidth}{@{\extracolsep{\fill}}llll@{}}
\toprule
 & \textbf{Model 1} & \textbf{Model 2} & \textbf{Model 3} \\
\textbf{Predictor} & \textbf{(RQ1)} & \textbf{(RQ2)} & \textbf{(RQ2)} \\
\hline
\textit{Focal condition effects} & & & \\
\quad Sustainability-aware (vs.\ preference-based)        & \textbf{1.69}$^{*}$ & --- & --- \\
                                                          & [1.13, 2.52] & & \\[3pt]
\quad Mechanism-free (vs.\ preference-only)              & --- & 1.08 & --- \\
                                                          & & [0.66, 1.77] & \\[3pt]
\quad Framing (vs.\ preference-only)                     & --- & \textbf{2.39}$^{***}$ & --- \\
                                                          & & [1.43, 3.99] & \\[3pt]
\quad Descriptive social norms (vs.\ preference-only)    & --- & \textbf{1.89}$^{*}$ & --- \\
                                                          & & [1.14, 3.15] & \\[3pt]
\quad Framing (vs.\ mechanism-free)                       & --- & --- & \textbf{2.21}$^{**}$ \\
                                                          & & & [1.31, 3.73] \\[3pt]
\quad Descriptive social norms (vs.\ mech.-free)          & --- & --- & \textbf{1.75}$^{*}$ \\
                                                          & & & [1.05, 2.93] \\[6pt]
\textit{Covariate} & & & \\
\quad Biospheric value orientation                        & 1.29$^{**}$ & 1.33$^{***}$ & 1.33$^{***}$ \\
                                                          & [1.11, 1.51] & [1.13, 1.56] & [1.13, 1.56] \\[3pt]
Other covariates adjusted (n.s.) & Yes & Yes & Yes \\
\hline
Observations & 529 & 529 & 529 \\
\bottomrule
\end{tabular*}
\end{table}

\subsection{RQ2: Does the Presentation of the Sustainability Content Matter?}
\label{sec:results-rq2}
\begin{figure}[t]
\centering
% Row 1: explanation-related
\begin{subfigure}{0.32\textwidth}\centering
  \includegraphics[width=\textwidth]{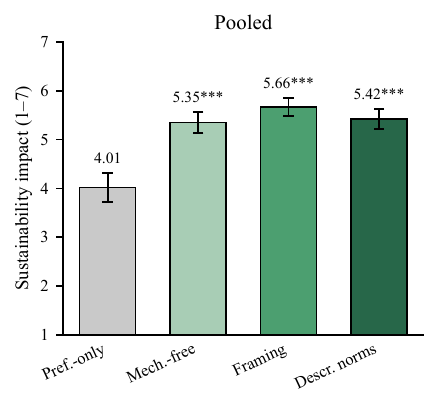}
  \caption{Sustainability impact}\end{subfigure}\hfill
\begin{subfigure}{0.32\textwidth}\centering
  \includegraphics[width=\textwidth]{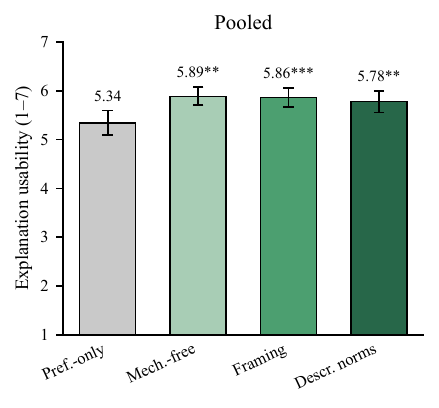}
  \caption{Explanation usability}\end{subfigure}\hfill
\begin{subfigure}{0.32\textwidth}\centering
  \includegraphics[width=\textwidth]{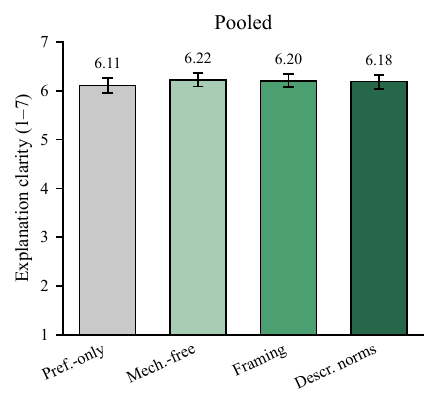}
  \caption{Explanation clarity}\end{subfigure}

\vspace{2mm}
% Row 2: decision-related
\begin{subfigure}{0.32\textwidth}\centering
  \includegraphics[width=\textwidth]{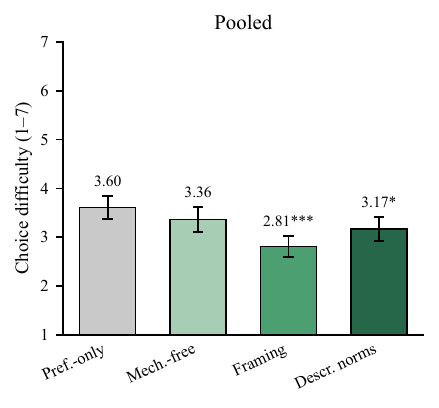}
  \caption{Choice difficulty}\end{subfigure}\hfill
\begin{subfigure}{0.32\textwidth}\centering
  \includegraphics[width=\textwidth]{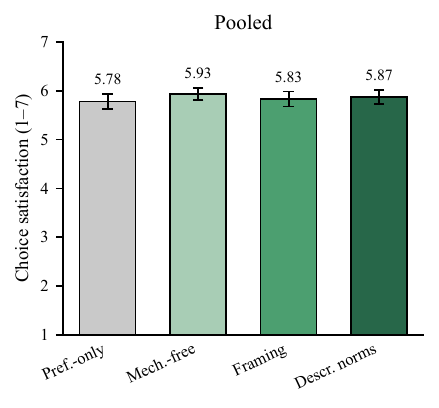}
  \caption{Choice satisfaction}\end{subfigure}\hfill
\begin{subfigure}{0.32\textwidth}\end{subfigure} % spacer to left-align row 2

\caption{Subjective evaluations by condition (pooled sample, $N=529$): explanation-
related measures (top) and decision-related measures (bottom). Bars show raw 
condition means; error bars denote 95\% confidence intervals ($t$-based). Asterisks 
mark covariate-adjusted contrasts against the preference-only baseline 
($^{*}p<.05$, $^{**}p<.01$, $^{***}p<.001$). Per-domain breakdowns appear in 
Appendix~\ref{app:per-domain-figures}.}
\Description{Five bar charts of subjective evaluations by condition (preference-only 
baseline, mechanism-free, framing, descriptive social norms) on the pooled sample. 
Top row: sustainability impact, explanation usability, and explanation clarity. 
Bottom row: choice difficulty and choice satisfaction. Sustainability impact and 
usability rise above baseline in the sustainability-aware conditions, choice 
difficulty falls under framing and descriptive norms, and clarity and satisfaction 
are similar across conditions.}
\label{fig:subjective}
\end{figure}

To isolate the effect of \emph{how} the content is presented, we compare the mechanism-based conditions against the baseline and against the mechanism-free condition separately, both on choice behavior (Models~2 and~3 of Table~\ref{tab:choice-models}) and on subjective evaluation (Table~\ref{tab:subjective-results}, Figure~\ref{fig:subjective}). 

On Choice, relative to the preference-only baseline (Model~2), plain disclosure in the mechanism-free condition did not significantly affect choice behavior (OR${}=1.08$, $p=.76$), whereas framing more than doubled the odds of choosing a sustainable item (OR${}=2.39$, $p=.001$) and descriptive social norms increased them by 89\% (OR${}=1.89$, $p=.014$). Using the mechanism-free condition as the reference 
(Model~3) shows that both mechanisms significantly outperformed plain disclosure of the same content, with framing showing the strongest effect (+121\%; OR${}=2.21$, $p=.003$) and descriptive social norms a smaller but significant increase (+75\%; OR${}=1.75$, $p=.032$). Choices thus 
changed only when the sustainability content was expressed through a targeted decision information mechanism, not from its mere presence.

On subjective evaluation, all three sustainability-aware conditions raised perceived sustainability impact well above the baseline ($M=4.01$; mechanism-free $b=+1.20$, framing $b=+1.67$, descriptive social norms $b=+1.33$; all $p<.001$) and modestly improved explanation usability ($M=5.34$; $b=+0.40$ to $+0.50$; all $p\le.008$). Explanation clarity was near-ceiling in every condition ($M=6.11$--$6.22$; all contrasts $p\ge.26$), so adding sustainability content---with or without a mechanism---did not make the explanations harder to follow.
Presentation also shaped the decision experience. Framing, and to a lesser degree descriptive social norms, reduced how difficult the choice felt, whereas plain disclosure in the mechanism-free condition did not. Choice difficulty fell relative to the baseline ($M=3.60$) under framing ($b=-0.84$, $p<.001$) and descriptive social norms ($b=-0.40$, $p=.017$), but not under mechanism-free disclosure ($b=-0.21$, $p=.21$). Choice satisfaction stayed uniformly high across conditions ($M=5.78$--$5.93$; all contrasts $p\ge.47$), so steering participants toward the sustainable option did not come at the cost of satisfaction with the choice.
Across constructs, participants higher in biospheric value orientation and in 
domain expertise gave more favorable evaluations throughout. The full covariate 
coefficients are reported in Appendix~\ref{app:subjective-models}.
\textbf{In answer to RQ2, only framing and descriptive social norms shifted choice and eased the decision, whereas plainly disclosing the same sustainability content did not, even though it still improved how the explanation was perceived.}

\begin{table}[ht]
  \centering
    \caption{Covariate-adjusted condition contrasts for the five subjective constructs 
(pooled sample, $N=529$). BASE = preference-only baseline, MF = mechanism-free, FR = framing, DSN = descriptive social norms. The BASE column gives the preference-only 
mean on the 7-point scale; each other column gives that condition's covariate-
adjusted difference from baseline ($b$), so a positive $b$ is above baseline and a 
negative $b$ below it (e.g., framing's sustainability impact is $4.01+1.67=5.68$). Making sustainability salient raised its perceived impact and the explanations' usability. Both framing and descriptive social 
norms additionally made the choice feel easier. Clarity and satisfaction contrasts are near zero and non-significant, indicating the added sustainability content did not come at the cost of comprehension or satisfaction. $^{*}p<.05$, $^{**}p<.01$, $^{***}p<.001$.}
  \label{tab:subjective-results}
  \begin{tabular}{l llll}
    \toprule
    \textbf{Construct} & \textbf{BASE} & \textbf{MF} & \textbf{FR} & \textbf{DSN} \\
    \midrule
    Sustainability impact & 4.01 & $+1.20^{***}$ & $+1.67^{***}$ & $+1.33^{***}$ \\
    Explanation usability & 5.34 & $+0.47^{**}$  & $+0.50^{***}$ & $+0.40^{**}$ \\
    Explanation clarity   & 6.11 & $+0.06$       & $+0.11$       & $+0.04$ \\
    Choice difficulty     & 3.60 & $-0.21$       & $-0.84^{***}$ & $-0.40^{*}$ \\
    Choice satisfaction   & 5.78 & $+0.07$       & $+0.07$       & $+0.04$ \\
    Other covariates adjusted & Yes & Yes & Yes & Yes \\
\hline
Observations & 529 & 529 & 529 & 529 \\
    \bottomrule
  \end{tabular}
\end{table}

\subsection{RQ3: Do the Effects Generalize Across Involvement Domains?}
\label{sec:results-generalization}

To test whether the effects hold in both domains, we extend each model reported above with a condition-by-domain interaction.
On choice, this interaction was far from significant, with and without covariates 
($\chi^2(3) = 0.21$, $p = .98$ and $\chi^2(3) = 0.88$, $p = .83$, respectively), so 
the condition effects on sustainable selection did not vary by domain.
As Figure~\ref{fig:choice} shows, the framing and descriptive-social-norms conditions 
raised the sustainable-selection rate into the 62--71\% range in each domain.
The subjective measures showed the same pattern, with no interaction approaching significance (sustainability impact $F(3,510) = 0.36$, $p = .78$; usability $F(3,510) = 1.48$, $p = .22$; clarity $F(3,510) = 0.10$, $p = .96$; choice satisfaction $F(3,510) = 2.05$, $p = .11$; choice difficulty $F(3,510) = 0.94$, $p = .42$).

We read these consistent nulls as evidence that the intervention generalizes across the two involvement contexts.
As such, the pooled models already capture the condition effects in both domains. We therefore report pooled estimates throughout and present the per-domain pattern descriptively in Appendix~\ref{app:per-domain-figures}.
\textbf{In answer to RQ3, the behavioral and experiential effects of sustainability-aware explanations are statistically indistinguishable across the routine, low-stakes purchase and the infrequent, high-stakes one alike.}

%%%%%%%%%%%%%%%%%%%%%%%%%%%%%%%%%%%%%%%%%%%%%%%%%%%%%%%%%%%%%%%%%%%%%%%%%%%%%%%%%%%%%%%%%%%%%%%%%

\section{Discussion, Limitations and Future Work}
\label{sec:discussion}

\subsection{Interpretation of Findings}
\label{sec:discussion-findings}

Our central finding is that sustainability-aware explanations changed choices among preference-matched recommendations (RQ1), and that this effect depended on how sustainability information was communicated rather than on its mere inclusion (RQ2). Simply disclosing a verified sustainability attribute did not increase sustainable choices beyond chance levels. In contrast, presenting the same attribute through established behavioral mechanisms approximately doubled the odds of selecting the sustainable option, increasing sustainable choices to 62--71\%. This result speaks against an information-deficit account 
\cite{schultz2002knowledge}, according to which providing environmental information alone should be sufficient to influence behavior. Instead, it aligns with insights from environmental psychology showing that knowledge alone is often insufficient to change environmentally relevant decisions \cite{kollmuss2002mind}.

Notably, while both mechanisms affected choice behavior, positive consequence framing consistently produced larger effects than social descriptive norms. Relative to the baseline, it also yielded the largest reduction in reported choice difficulty and the highest perceived sustainability impact.
This ordering is consistent with the underlying behavioral theories. Framing alters the subjective valuation of the reference point against which outcomes are evaluated \cite{tversky1981framing,levin1998all}, whereas social descriptive norms provide a heuristic based on others' behavior \cite{cialdini2004social,goldstein2008room}, thus requiring an additional inferential step. That is, the decision-maker must regard the referenced group as relevant and use its behavior as guidance.
Accordingly, norm effects depend on the salience and perceived similarity of the reference group \cite{goldstein2008room} and may weaken or even reverse when identification with that group is low \cite{schultz2007constructive}. 
Because our descriptive norm referred to users in the same decision context (e.g., 
``customers browsing similar products'') rather than to a personally meaningful 
in-group (e.g. peers of their own age, or their social ties), it likely represents a comparatively conservative implementation of the mechanism and may have exerted a 
weaker normative pull than an identity-based reference group would 
\cite{goldstein2008room}.
More generally, given meta-analytic evidence on substantial heterogeneity in nudge effectiveness across interventions and contexts \cite{mertens2022effectiveness}, we interpret this ordering as specific to our setting rather than as a general ranking of framing and normative appeals.

Choice difficulty traces one route by which the behavioral effect could arise. Both behavioral mechanism-based conditions reduced perceived choice difficulty, whereas the mechanism-free disclosure did not. Difficulty therefore tracked changes in choice behavior rather than perceived explanation quality, suggesting that the behavioral mechanisms did more than communicate additional information. They also supplied a heuristic that helped resolve otherwise balanced decisions between equally well-matched alternatives \cite{cialdini2004social,bollen2010understanding}. Instead if changing what participants believed, the explanations appear to have changed how participants reached a decision. At the same time, explanation clarity and choice satisfaction remained uniformly high across conditions, indicating that increased persuasive effectiveness did not come at the expense of other explanation objectives \cite{tintarev2015explaining}.

These findings suggest that subjective evaluations alone are insufficient as dependent variables for assessing persuasive explanations, because favorable judgments of explanations need not translate into corresponding behavioral effects.
Perceived sustainability impact and explanation usability rose in all three sustainability-aware conditions. Participants in the mechanism-free condition read that an option was more sustainable and judged the explanations as being more useful, yet this did not affect their choices, so a favorably rated explanation could be entirely inert in behavioral terms. This dissociation mirrors the attitude--behavior gap in sustainable consumption, in which favorable evaluations of green information fail to translate into green choices \cite{kollmuss2002mind,white2019shift}. Evaluations that rely on questionnaire measures alone therefore risk missing the outcome that matters.

The observed effects generalized across the two involvement domains (RQ3), which 
anchor opposite ends of the involvement continuum (high and low). Neither behavioral nor experiential outcomes differed between the routine, low-stakes coffee purchase and the infrequent, high-stakes hotel booking. Although consumer involvement influences the extent to which decision information is processed \cite{zaichkowsky1985measuring,kapferer1985consumer}, and sustainability explanations have previously been evaluated differently across domains \cite{tran2024less}, the present interventions produced comparable effects in both contexts. 

One possible explanation for this robustness is that the nudge was delivered through personalized explanations. Because each explanation was generated for a specific user--item pair, the behavioral mechanism was embedded directly within an option that already matched the user's preferences, rather than presented as a generic appeal. Prior work suggests that personalized interventions can be more effective than one-size-fits-all nudges \cite{mills2022personalized,matz2024potential}. Explanations naturally provide such personalization at the point of decision, where users evaluate individual alternatives \cite{munscher2016review,soler2010point}. If so, the effectiveness of the intervention may depend less on the general level of involvement associated with the product category than on its integration into the immediate decision context.
Biospheric value orientation independently predicted sustainable choices \cite{de2007value,o2016impact}. Nevertheless, the explanation effects remained significant after accounting for this predisposition, indicating that the interventions operated in addition to, rather than merely reflecting, pre-existing environmental values.

Finally, the study provides evidence that LLMs can operationalize behavioral interventions in recommender systems in real time, generating a distinct explanation for every user--item pair live during each session while implementing 
theoretically grounded strategies. The resulting behavioral effects indicate that framing and descriptive norms remain effective even when realized through dynamically generated language instead of standardized experimental stimuli, supporting their applicability in personalized RS \cite{matz2024potential}. 
This conclusion depends on the fidelity of the generated explanations. Automated and human validation confirmed that the four conditions remained textually distinct, that each explanation faithfully instantiated the theoretical mechanism, and that sustainability statements were almost exclusively grounded in verified item attributes. In other words, the language model served as an implementation tool for predefined behavioral strategies rather than as an unconstrained generator of persuasive text.
Such distinction is important, as LLM-generated explanations can only support valid behavioral conclusions if they preserve both theoretical validity and factual accuracy. A system that hallucinates or embellishes sustainability claims could also influence decisions, but any resulting effects would arise from misinformation rather than from the behavioral mechanism under investigation \cite{ji2023survey}, violating the faithfulness expected of post-hoc explanations \cite{rudin2019stop}. The behavioral effects observed in our study therefore reflect differences in how verified sustainability information was presented while holding the underlying factual content constant.

\subsection{Implications}
\label{sec:discussion-implications}

We summarize the implications of this work and its findings as follows.

\textit{Methodological implications.} 
The study demonstrates that LLMs can operationalize theory-grounded behavioral interventions under experimental control. Generating experimental stimuli with an LLM poses two methodological challenges: the generated text must consistently implement the intended behavioral mechanism while remaining factually accurate. Our approach addresses both through constrained prompts derived from behavioral theory, a blind cross-provider audit of treatment distinctiveness and mechanism adherence, and an attribute-level fidelity check with independent human labeling. Rather than allowing the language model to determine the intervention, behavioral theory specifies the manipulation, while validation establishes that the generated text faithfully instantiates it. Grounding every sustainability claim in a verified item attribute is particularly important because it establishes that the observed behavioral effects arise from the intended manipulation rather than unsupported or hallucinated claims \cite{ji2023survey}. Recent work has argued that LLMs can serve as instruments in social-scientific research provided that their outputs are systematically validated rather than assumed to be correct \cite{bail2024can,ziems2024can}. Our validation framework provides one concrete realization of this principle. Although developed for sustainability explanations, it is applicable wherever LLM-generated text is used as a controlled behavioral treatment.

\textit{Theoretical implications.} 
Our findings further position recommendation explanations as a component of digital choice architecture, extending existing work on RS for social good. Prior research shows that explanations can influence user perceptions and acceptance of recommendations \cite{herlocker2000explaining,gkika2014persuasive}, while digital nudging research has predominantly examined structural interface interventions such as defaults, rankings, and labels \cite{karlsen2019recommendations,jesse2021digital,caraban201923}. We connect these perspectives by showing how behavioral mechanisms can be embedded within the natural-language rationale accompanying a personalized recommendation. The nested design further separates two layers that are often conflated: the informational effect of disclosing an attribute and the behavioral effect of presenting that attribute through a specific mechanism. In our setting, disclosure alone did not change choices, whereas theory-grounded presentation did, indicating that information provision and behavioral framing represent distinct explanatory functions rather than different intensities of the same intervention. These findings suggest that recommendation explanations can serve not only to justify why an option was selected, but also as a channel through which socially relevant decision support can be delivered.

\textit{Practical implications.}
Integrating sustainability attributes into recommendation explanations offers a seamless intervention for e-commerce platforms. Because LLMs can generate these explanations on-the-fly for each user--item pair, the approach encourages sustainable choices simply by changing how verified attributes are communicated, and practitioners can incorporate these behavioral mechanisms into existing workflows without altering underlying algorithms, re-ranking items, or restricting user choices. To maintain user trust, platforms must validate all sustainability claims to prevent misleading environmental statements \cite{de2020concepts}. Generating each explanation in live systems also carries a performance cost, since it must run within interactive latency and budget limits that tighten at scale.

\textit{Ethical implications.} The efficacy of presentation-based interventions introduces risks, as the same mechanisms can covertly steer users toward objectives that do not serve their interests. Consequently, deploying explanation-embedded nudges necessitates transparency regarding their persuasive intent. Although such disclosure is traditionally assumed to neutralize behavioral interventions, prior evidence suggests that transparency and efficacy are not mutually exclusive, as nudges frequently retain their behavioral impact even when their presence and intent are made explicit \cite{bruns2018can,loewenstein2015warning}. This caution is reinforced by evidence that nudging in recommendation, though effective in the short run, can erode users' quality perceptions and long-term 
intention to use the system \cite{alves2024digitally}, underscoring the need to 
monitor such effects when deploying explanation-embedded nudges.

\subsection{Limitations}
\label{sec:discussion-limitations}

Our conclusions should be interpreted in light of several limitations.
First, participants made a hypothetical choice in an online study rather than a real purchase. While measuring choice is arguably a more demanding criterion than the self-reported ratings common in explanation research \cite{cosley2003seeing}, effect sizes in consequential settings may still differ.
Second, our operationalization of sustainability relied on a single verified attribute per domain, a packaging claim or a certification, which served as a proxy rather than a comprehensive assessment of environmental impact. 
Third, the textual manipulation in the three sustainability-aware conditions was accompanied by a visual badge, adopted after pilot testing indicated that text alone produced markedly weaker effects, which would have required a substantially larger sample to detect reliably. This does not threaten our main comparisons. Because the badge was held identical across the three sustainability-aware conditions, it cannot account for the differences among them, so the RQ2 comparisons still isolate textual presentation as intended.
Fourth, participants encountered the explanations in a single interaction, so whether the effects persist under repeated exposure remains open, as habituation, wear-out, or reactance to a familiar mechanism could attenuate them over time \cite{allcott2014short,sunstein2017nudges}. 
Finally, two factors bound the generalizability of our findings. Our design covers only two decision information mechanisms, framing and descriptive social norms, so the observed effects are specific to these mechanisms and should not be read as a comprehensive assessment of nudging strategies more broadly. Moreover, although our framework is model-agnostic, a systematic comparison of how different generator 
models affect the \emph{resulting behavioral effects} remains open.

\subsection{Future Work}
\label{sec:discussion-future}
First, field studies on live platforms with consequential transactions and longitudinal exposure would establish whether the effects persist outside the laboratory and over time, directly addressing the hypothetical, single-interaction nature of our design. This matters because nudge effects estimated in academic studies are often substantially larger than those in at-scale field deployments \cite{dellavigna2022rcts}, and can attenuate under repeated exposure \cite{allcott2014short}. Such settings would also permit a test our design cannot support, namely whether an explanation-embedded nudge retains its behavioral effect once its persuasive intent is disclosed to users, which laboratory evidence on transparent nudging suggests is plausible \cite{bruns2018can,sunstein2016ethics}.
Second, the independent effect of biospheric values invites \emph{personalized} nudging that selects or adapts the mechanism to psychological user characteristics, consistent with evidence that persuasive appeals are more effective when tailored to the recipient's traits \cite{hirsh2012personalized} and with recommender systems research showing that personality-aligned explanation framings can make recommendations appear more persuasive, clearer, and more accurate \cite{alves2026exploring}.
LLM-based generation removes the principal scalability barrier to doing so per user. Widening the mechanism repertoire beyond the two mechanisms studied here is also a natural companion step.
Third, the approach extends beyond static explanation slots. In conversational recommender systems and emerging LLM-based agents \cite{jannach2021survey,gao2021advances}, explanations are co-constructed in dialogue, raising the question of how 
sustainability-aware explanations fare when users can probe, challenge, or negotiate them---a setting where the persuasive intent becomes salient and reactance more likely \cite{sunstein2017nudges}. Such settings also dissolve a boundary our design held fixed. Rather than matching recommendations to preferences elicited in advance, conversational systems help construct and refine those preferences through the interaction itself \cite{bettman1998constructive}, so 
explanations could influence not only which option is chosen but the also the preferences against which options are evaluated. More broadly, explanation-as-nudge approaches could serve other recommender systems for social-good objectives \cite{boratto2024first,jannach2024recommender}, wherever a verified item attribute aligns with user and societal welfare.

\section{Conclusion}
\label{sec:conclusion}

This work investigated whether LLM-generated sustainability-aware explanations can support sustainable choice without altering the recommender that produces them. In two randomized between-subjects studies spanning a low- and a high-involvement domain ($N = 529$), we embedded a verified sustainability attribute into LLM-generated explanations and varied only how that attribute was expressed. Disclosing the attribute plainly left choices at chance. Expressing the same attribute through positive consequence framing or a descriptive social norm roughly doubled the odds of a sustainable choice, reduced perceived choice difficulty, and left choice satisfaction and explanation clarity unaffected. The effects held in both domains and above participants' pre-existing environmental values.

This work thereby establishes the explanation as an element of choice architecture, and it shows that LLMs can instantiate theory-grounded behavioral mechanisms at the scale personalization requires, provided the generated text is audited for adherence to the intended mechanism and for grounding in a verified attribute. The explanations that recommender systems already display therefore constitute a practical point of intervention for supporting sustainable consumption, one that requires no change to the underlying algorithm and no restriction of the user's options.

\bibliographystyle{ACM-Reference-Format}
\bibliography{sample-base}

@String{Computing = "Computing" }

@String{Computer = "{IEEE} Computer" }

@String{Springer = "Springer-Verlag" }

@book{jannach2010recommender,
  title={Recommender systems: an introduction},
  author={Jannach, Dietmar and Zanker, Markus and Felfernig, Alexander and Friedrich, Gerhard},
  year={2010},
  publisher={Cambridge university press}
}

@inproceedings{tintarev2007explanations,
  title={Explanations of recommendations},
  author={Tintarev, Nava},
  booktitle={Proceedings of the 2007 ACM Conference on Recommender Systems},
  pages={203--206},
  year={2007}
}

@inproceedings{tintarev2007survey,
  title={A survey of explanations in recommender systems},
  author={Tintarev, Nava and Masthoff, Judith},
  booktitle={2007 IEEE 23rd International Conference on Data Engineering Workshop},
  pages={801--810},
  year={2007},
  organization={IEEE}
}

@inproceedings{felfernig2008constraint,
  title={Constraint-based recommender systems: technologies and research issues},
  author={Felfernig, Alexander and Burke, Robin},
  booktitle={Proceedings of the 10th 
  International Conference on Electronic Commerce},
  pages={1--10},
  year={2008}
}

@inproceedings{tran2024less,
  title={Less is More: Towards Sustainability-Aware Persuasive Explanations in Recommender Systems},
  author={Tran, Thi Ngoc Trang and Polat Erdeniz, Seda and Felfernig, Alexander and Lubos, Sebastian and El Mansi, Merfat and Le, Viet-Man},
  booktitle={Proceedings of the 18th ACM Conference on Recommender Systems},
  pages={1108--1112},
  year={2024}
}

@article{jesse2021digital,
  title={Digital nudging with recommender systems: Survey and future directions},
  author={Jesse, Mathias and Jannach, Dietmar},
  journal={Computers in Human Behavior Reports},
  volume={3},
  pages={100052},
  year={2021},
  publisher={Elsevier}
}

@article{gena2019personalization,
  title={When personalization is not an option: An in-the-wild study on persuasive news recommendation},
  author={Gena, Cristina and Grillo, Pierluigi and Lieto, Antonio and Mattutino, Claudio and Vernero, Fabiana},
  journal={Information},
  volume={10},
  number={10},
  pages={300},
  year={2019},
  publisher={MDPI}
}

@incollection{tintarev2015explaining,
  title={Explaining recommendations: Design and evaluation},
  author={Tintarev, Nava and Masthoff, Judith},
  booktitle={Recommender Systems Handbook},
  pages={353--382},
  year={2015},
  publisher={Springer}
}

@article{tintarev2022beyond,
  title={Beyond Explaining Single Item Recommendations},
  author={Tintarev, Nava and Masthoff, Judith},
  journal={Recommender Systems Handbook},
  pages={711},
  year={2022},
  publisher={Springer}
}

@article{felfernig2023recommender,
  title={Recommender systems for sustainability: overview and research issues},
  author={Felfernig, Alexander and Wundara, Manfred and Tran, Thi Ngoc Trang and Polat-Erdeniz, Seda and Lubos, Sebastian and El Mansi, Merfat and Garber, Damian and Le, Viet-Man},
  journal={Frontiers in Big Data},
  volume={6},
  pages={1284511},
  year={2023},
  publisher={Frontiers Media SA}
}

@book{cialdini1993influence,
  title     = {{Influence: The Psychology of Persuasion}},
  author    = {Cialdini, Robert B.},
  year      = {1993},
  edition   = {Rev. ed.},
  publisher = {Morrow},
  address   = {New York}
}

@book{thaler2009nudge,
  title={Nudge: Improving decisions about health, wealth, and happiness},
  author={Thaler, Richard H and Sunstein, Cass R},
  year={2009},
  publisher={Penguin}
}

@article{karlsen2019recommendations,
  title={Recommendations with a nudge},
  author={Karlsen, Randi and Andersen, Anders},
  journal={Technologies},
  volume={7},
  number={2},
  pages={45},
  year={2019},
  publisher={MDPI}
}

@article{weinmann2016digital,
  title={Digital nudging},
  author={Weinmann, Markus and Schneider, Christoph and Brocke, Jan vom},
  journal={Business \& Information Systems Engineering},
  volume={58},
  pages={433--436},
  year={2016},
  publisher={Springer}
}

@article{starke2024tell,
  title={“Tell Me Why”: using natural language justifications in a recipe recommender system to support healthier food choices},
  author={Starke, Alain D and Musto, Cataldo and Rapp, Amon and Semeraro, Giovanni and Trattner, Christoph},
  journal={User Modeling and User-Adapted Interaction},
  volume={34},
  number={2},
  pages={407--440},
  year={2024},
  publisher={Springer}
}

@inproceedings{boratto2024first,
  title={First International Workshop on Recommender Systems for Sustainability and Social Good (RecSoGood 2024)},
  author={Boratto, Ludovico and De Filippo, Allegra and Lex, Elisabeth and Ricci, Francesco},
  booktitle={Proceedings of the 18th ACM Conference on Recommender Systems},
  pages={1239--1241},
  year={2024}
}

@article{zhang2021consumer,
  title={Consumer attitude towards sustainability of fast fashion products in the UK},
  author={Zhang, Bo and Zhang, Yaozhong and Zhou, Peng},
  journal={Sustainability},
  volume={13},
  number={4},
  pages={1646},
  year={2021},
  publisher={MDPI}
}

@article{kapferer1985consumer,
  title={Consumer involvement profiles: a new and practical approach to consumer involvement},
  author={Kapferer, Jean-Noel and Laurent, Gilles},
  journal={Journal of advertising research},
  volume={25},
  number={6},
  pages={48--56},
  year={1985},
  publisher={Taylor \& Francis}
}

@inproceedings{spillo2025training,
  title={Training Green and Sustainable Recommendation Models: Introducing Carbon Footprint Data into Early Stopping Criteria},
  author={Spillo, Giuseppe and De Filippo, Allegra and Fontana, Emanuele and Milano, Michela and Semeraro, Giovanni},
  booktitle={Proceedings of the 33rd ACM Conference on User Modeling, Adaptation and Personalization},
  pages={341--346},
  year={2025}
}

@inproceedings{mauro2024point,
  title={Point-of-Interest Recommender Systems: Nudging towards Sustainable Tourism},
  author={Mauro, Noemi and Scarpinati, Livio and Ferrero, Fabio and Geninatti Cossatin, Angelo and Mattutino, Claudio},
  booktitle={Adjunct Proceedings of the 32nd ACM Conference on User Modeling, Adaptation and Personalization},
  pages={491--495},
  year={2024}
}

@inproceedings{peake2018explanation,
  title={Explanation mining: Post hoc interpretability of latent factor models for recommendation systems},
  author={Peake, Georgina and Wang, Jun},
  booktitle={Proceedings of the 24th ACM SIGKDD nternational Conference on Knowledge Discovery \& Data Mining},
  pages={2060--2069},
  year={2018}
}

@article{jannach2024recommender,
  title={Recommender Systems for Good (RS4Good): Survey of Use Cases and a Call to Action for Research that Matters},
  author={Jannach, Dietmar and Said, Alan and Tkalcic, Marko and Zanker, Markus},
  journal={ACM Transactions on Recommender Systems},
  year={2024},
  publisher={ACM New York, NY}
}

@article{starke2021promoting,
  title={Promoting energy-efficient behavior by depicting social norms in a recommender interface},
  author={Starke, Alain and Willemsen, Martijn and Snijders, Chris},
  journal={ACM Transactions on Interactive Intelligent Systems (TiiS)},
  volume={11},
  number={3-4},
  pages={1--32},
  year={2021},
  publisher={ACM New York, NY}
}

@article{tintarev2012evaluating,
  title={Evaluating the effectiveness of explanations for recommender systems: Methodological issues and empirical studies on the impact of personalization},
  author={Tintarev, Nava and Masthoff, Judith},
  journal={User Modeling and User-Adapted Interaction},
  volume={22},
  pages={399--439},
  year={2012},
  publisher={Springer}
}

@inproceedings{bilgic2005explaining,
  title={Explaining recommendations: Satisfaction vs. promotion},
  author={Bilgic, Mustafa and Mooney, Raymond J},
  booktitle={Beyond Personalization Workshop, IUI},
  volume={5},
  pages={153},
  year={2005}
}

@article{fogg2002persuasive,
  title={Persuasive technology: using computers to change what we think and do},
  author={Fogg, Brian J},
  journal={Ubiquity},
  volume={2002},
  number={December},
  pages={2},
  year={2002},
  publisher={ACM New York, NY, USA}
}

@incollection{starke2024psychologically,
  title={Psychologically Informed Design of Energy Recommender Systems: Are Nudges Still Effective in Tailored Choice Environments?},
  author={Starke, Alain D and Willemsen, Martijn C},
  booktitle={A Human-Centered Perspective of Intelligent Personalized Environments and Systems},
  pages={221--259},
  year={2024},
  publisher={Springer}
}

@inproceedings{said2024recommender,
  title={Recommender Systems for Social Good: The Role of Accountability and Sustainability},
  author={Said, Alan},
  booktitle={International Workshop on Recommender Systems for Sustainability and Social Good},
  pages={1--4},
  year={2024},
  organization={Springer}
}

@article{friedrich2011taxonomy,
  title={A taxonomy for generating explanations in recommender systems},
  author={Friedrich, Gerhard and Zanker, Markus},
  journal={AI Magazine},
  volume={32},
  number={3},
  pages={90--98},
  year={2011}
}

@article{nunes2017systematic,
  title={A systematic review and taxonomy of explanations in decision support and recommender systems},
  author={Nunes, Ingrid and Jannach, Dietmar},
  journal={User Modeling and User-Adapted Interaction},
  volume={27},
  number={3},
  pages={393--444},
  year={2017},
  publisher={Springer}
}

@inproceedings{el2025nudging,
  title     = {{Nudging Healthy Choices: Leveraging LLM-Generated Hashtags and Explanations in Personalized Food Recommendations}},
  author    = {El Majjodi, Ayoub and Starke, Alain and Trattner, Christoph and Petruzzelli, Alessandro and Musto, Cataldo},
  booktitle = {{IntRS 2025: Interfaces and Human Decision Making for Recommender Systems}},
  year      = {2025},
}

@article{said2025explaining,
  title={On explaining recommendations with Large Language Models: a review},
  author={Said, Alan},
  journal={Frontiers in Big Data},
  volume={7},
  pages={1505284},
  year={2025},
  publisher={Frontiers Media SA}
}

@inproceedings{lubos2024llm,
  title={LLM-generated explanations for recommender systems},
  author={Lubos, Sebastian and Tran, Thi Ngoc Trang and Felfernig, Alexander and Polat Erdeniz, Seda and Le, Viet-Man},
  booktitle={Adjunct Proceedings of the 32nd ACM Conference on User Modeling, Adaptation and Personalization},
  pages={276--285},
  year={2024}
}

@inproceedings{silva2024leveraging,
  title={Leveraging chatgpt for automated human-centered explanations in recommender systems},
  author={Silva, {\'I}tallo and Marinho, Leandro and Said, Alan and Willemsen, Martijn C},
  booktitle={Proceedings of the 29th International Conference on Intelligent User Interfaces},
  pages={597--608},
  year={2024}
}

@article{munscher2016review,
  title={A review and taxonomy of choice architecture techniques},
  author={M{\"u}nscher, Robert and Vetter, Max and Scheuerle, Thomas},
  journal={Journal of Behavioral Decision Making},
  volume={29},
  number={5},
  pages={511--524},
  year={2016},
  publisher={Wiley Online Library}
}

@article{mertens2022effectiveness,
  title={The effectiveness of nudging: A meta-analysis of choice architecture interventions across behavioral domains},
  author={Mertens, Stephanie and Herberz, Mario and Hahnel, Ulf JJ and Brosch, Tobias},
  journal={Proceedings of the National Academy of Sciences},
  volume={119},
  number={1},
  pages={e2107346118},
  year={2022},
  publisher={National Academy of Sciences}
}

@incollection{knijnenburg2015evaluating,
  title={Evaluating recommender systems with user experiments},
  author={Knijnenburg, Bart P and Willemsen, Martijn C},
  booktitle={Recommender Systems Handbook},
  pages={309--352},
  year={2015},
  publisher={Springer}
}

@inproceedings{kouki2017user,
  title={User preferences for hybrid explanations},
  author={Kouki, Pigi and Schaffer, James and Pujara, Jay and O'Donovan, John and Getoor, Lise},
  booktitle={Proceedings of the Eleventh ACM Conference on Recommender Systems},
  pages={84--88},
  year={2017}
}

@article{willemsen2016understanding,
  title={Understanding the role of latent feature diversification on choice difficulty and satisfaction},
  author={Willemsen, Martijn C and Graus, Mark P and Knijnenburg, Bart P},
  journal={User Modeling and User-Adapted Interaction},
  volume={26},
  number={4},
  pages={347--389},
  year={2016},
  publisher={Springer}
}

@article{okoso2025impact,
  title={Impact of tone-aware explanations in recommender systems},
  author={Okoso, Ayano and Otaki, Keisuke and Koide, Satoshi and Baba, Yukino},
  journal={ACM Transactions on Recommender Systems},
  volume={3},
  number={4},
  pages={1--34},
  year={2025},
  publisher={ACM New York, NY}
}

@inproceedings{erdeniz2023employing,
  title={Employing nudge theory and persuasive principles with explainable ai in clinical decision support},
  author={Erdeniz, Seda Polat and Tran, Thi Ngoc Trang and Felfernig, Alexander and Lubos, Sebastian and Schrempf, Michael and Kramer, Diether and Rainer, Peter P},
  booktitle={2023 IEEE International Conference on Bioinformatics and Biomedicine (BIBM)},
  pages={2983--2989},
  year={2023},
  organization={IEEE}
}

@book{yoo2012persuasive,
  title={Persuasive recommender systems: conceptual background and implications},
  author={Yoo, Kyung-Hyan and Gretzel, Ulrike and Zanker, Markus},
  year={2012},
  publisher={Springer Science \& Business Media}
}

@article{zaichkowsky1985measuring,
  title={Measuring the involvement construct},
  author={Zaichkowsky, Judith Lynne},
  journal={ournal of Consumer Research},
  pages={341--352},
  year={1985},
  publisher={JSTOR}
}

@inproceedings{tran2021users,
  title={Do users appreciate explanations of recommendations? An analysis in the movie domain},
  author={Tran, Thi Ngoc Trang and Le, Viet Man and Atas, Muesluem and Felfernig, Alexander and Stettinger, Martin and Popescu, Andrei},
  booktitle={Proceedings of the 15th ACM Conference on Recommender Systems},
  pages={645--650},
  year={2021}
}

@article{de2007value,
  title={Value orientations and environmental beliefs in five countries: Validity of an instrument to measure egoistic, altruistic and biospheric value orientations},
  author={De Groot, Judith IM and Steg, Linda},
  journal={Journal of Cross-Cultural Psychology},
  volume={38},
  number={3},
  pages={318--332},
  year={2007},
  publisher={Sage Publications Sage CA: Los Angeles, CA}
}

@article{o2016impact,
  title={The impact of sustainability information on consumer decision making},
  author={O'Rourke, Dara and Ringer, Abraham},
  journal={Journal of Industrial Ecology},
  volume={20},
  number={4},
  pages={882--892},
  year={2016},
  publisher={Wiley Online Library}
}

@inproceedings{el2022nudging,
  title={Nudging towards health? examining the merits of nutrition labels and personalization in a recipe recommender system},
  author={El Majjodi, Ayoub and Starke, Alain D and Trattner, Christoph},
  booktitle={Proceedings of the 30th ACM Conference on User Modeling, Adaptation and Personalization},
  pages={48--56},
  year={2022}
}

@article{lisha2025service,
  title={From service to intention: A Technology Acceptance Model Perspective on E-Service Quality in Online Hotel Booking},
  author={Lisha, Chen and Goh, Chin Fei and Low, Yun Min and Tan, Owee Kowang and Lim, Kim Yew},
  journal={Sage Open},
  volume={15},
  number={4},
  pages={21582440251397834},
  year={2025},
  publisher={SAGE Publications Sage CA: Los Angeles, CA}
}

@article{weber2013doing,
  title={Doing the right thing willingly},
  author={Weber, Elke},
  journal={The Behavioral Foundations of Public Policy},
  volume={380},
  year={2013},
  publisher={Princeton University Press}
}

@article{tversky1981framing,
  title={The framing of decisions and the psychology of choice},
  author={Tversky, Amos and Kahneman, Daniel},
  journal={Science},
  volume={211},
  number={4481},
  pages={453--458},
  year={1981},
  publisher={American Association for the Advancement of Science}
}

@article{levin1998all,
  title={All frames are not created equal: A typology and critical analysis of framing effects},
  author={Levin, Irwin P and Schneider, Sandra L and Gaeth, Gary J},
  journal={Organizational Behavior and Human Decision Processes},
  volume={76},
  number={2},
  pages={149--188},
  year={1998},
  publisher={Elsevier}
}

@article{cialdini2003crafting,
  title={Crafting normative messages to protect the environment},
  author={Cialdini, Robert B},
  journal={Current Directions in Psychological Science},
  volume={12},
  number={4},
  pages={105--109},
  year={2003},
  publisher={SAGE Publications Sage CA: Los Angeles, CA}
}

@article{cialdini1990focus,
  title={A focus theory of normative conduct: Recycling the concept of norms to reduce littering in public places.},
  author={Cialdini, Robert B and Reno, Raymond R and Kallgren, Carl A},
  journal={Journal of Personality and Social Psychology},
  volume={58},
  number={6},
  pages={1015},
  year={1990},
  publisher={American Psychological Association}
}

@article{goldstein2008room,
  title={A room with a viewpoint: Using social norms to motivate environmental conservation in hotels},
  author={Goldstein, Noah J and Cialdini, Robert B and Griskevicius, Vladas},
  journal={ournal of Consumer Research},
  volume={35},
  number={3},
  pages={472--482},
  year={2008},
  publisher={The University of Chicago Press}
}

@inproceedings{zhu2012switch,
  title={To switch or not to switch: understanding social influence in online choices},
  author={Zhu, Haiyi and Huberman, Bernardo and Luon, Yarun},
  booktitle={Proceedings of the SIGCHI Conference on Human Factors in Computing Systems},
  pages={2257--2266},
  year={2012}
}

@inproceedings{cosley2003seeing,
  title={Is Seeing Believing? How Recommender Interfaces Affect Users' Opinion},
  author={Cosley, D and Lam, SK and Albert, I and Konstan, JA and Riedl, J},
  booktitle={Conf. on Human Factors in Computing Systems-CHI 2003},
  pages={585--592},
  year={2003}
}

@inproceedings{halimeh2025towards,
  title={Towards Greener Choices: Decision Information Nudging for Sustainability-Aware Recommender Explanations},
  author={Halimeh, Haya and M{\"u}ller, Oliver},
  booktitle={International Workshop on Recommender Systems for Sustainability and Social Good},
  pages={27--42},
  year={2025},
  organization={Springer}
}

@article{cialdini2004social,
  title={Social influence: Compliance and conformity},
  author={Cialdini, Robert B and Goldstein, Noah J},
  journal={Annu. Rev. Psychol.},
  volume={55},
  number={1},
  pages={591--621},
  year={2004},
  publisher={Annual Reviews}
}

@article{petty1983central,
  title={Central and peripheral routes to advertising effectiveness: The moderating role of involvement},
  author={Petty, Richard E and Cacioppo, John T and Schumann, David},
  journal={ournal of Consumer Research},
  volume={10},
  number={2},
  pages={135--146},
  year={1983},
  publisher={The University of Chicago Press}
}

@article{lu2023user,
  title={User perception of recommendation explanation: Are your explanations what users need?},
  author={Lu, Hongyu and Ma, Weizhi and Wang, Yifan and Zhang, Min and Wang, Xiang and Liu, Yiqun and Chua, Tat-Seng and Ma, Shaoping},
  journal={ACM Transactions on Information Systems},
  volume={41},
  number={2},
  pages={1--31},
  year={2023},
  publisher={ACM New York, NY}
}

@article{zhang2020explainable,
  title={Explainable recommendation: A survey and new perspectives},
  author={Zhang, Yongfeng and Chen, Xu and others},
  journal={Foundations and Trends{\textregistered} in Information Retrieval},
  volume={14},
  number={1},
  pages={1--101},
  year={2020},
  publisher={Now Publishers, Inc.}
}

@inproceedings{zanker2010knowledgeable,
  title={Knowledgeable explanations for recommender systems},
  author={Zanker, Markus and Ninaus, Daniel},
  booktitle={2010 IEEE/WIC/ACM International Conference on Web Intelligence and Intelligent Agent Technology},
  volume={1},
  pages={657--660},
  year={2010},
  organization={IEEE}
}

@inproceedings{gkika2014persuasive,
  title={The Persuasive Role of Explanations in Recommender Systems.},
  author={Gkika, Sofia and Lekakos, George},
  booktitle={BCSS@ PERSUASIVE},
  pages={59--68},
  year={2014}
}

@inproceedings{hendrawan2024explanations,
  title={Explanations in open user models for personalized information exploration},
  author={Hendrawan, Rully Agus and Brusilovsky, Peter and Lekshmi Narayanan, Arun Balajiee and Barria-Pineda, Jordan},
  booktitle={Adjunct Proceedings of the 32nd ACM Conference on User Modeling, Adaptation and Personalization},
  pages={256--263},
  year={2024}
}

@inproceedings{guo2023towards,
  title={Towards explainable conversational recommender systems},
  author={Guo, Shuyu and Zhang, Shuo and Sun, Weiwei and Ren, Pengjie and Chen, Zhumin and Ren, Zhaochun},
  booktitle={Proceedings of the 46th International ACM SIGIR Conference on Research and Development in Information Retrieval},
  pages={2786--2795},
  year={2023}
}

@book{thaler2021nudge,
  title={Nudge: The final edition},
  author={Thaler, Richard H and Sunstein, Cass R},
  year={2021},
  publisher={Penguin}
}

@inproceedings{starke2020little,
  title={With a little help from my peers: Depicting social norms in a recommender interface to promote energy conservation},
  author={Starke, Alain D and Willemsen, Martijn C and Snijders, Chris},
  booktitle={Proceedings of the 25th International Conference on Intelligent User Interfaces},
  pages={568--578},
  year={2020}
}

@inproceedings{musto2019justifying,
  title={Justifying recommendations through aspect-based sentiment analysis of users reviews},
  author={Musto, Cataldo and Lops, Pasquale and de Gemmis, Marco and Semeraro, Giovanni},
  booktitle={Proceedings of the 27th ACM Conference on User Modeling, Adaptation and Personalization},
  pages={4--12},
  year={2019}
}

@inproceedings{hirschmeier2020approach,
  title={An approach to explanations for public radio recommendations},
  author={Hirschmeier, Stefan and Schoder, Detlef},
  booktitle={Adjunct Publication of the 28th ACM Conference on User Modeling, Adaptation and Personalization},
  pages={237--240},
  year={2020}
}

@inproceedings{herlocker2000explaining,
  title={Explaining collaborative filtering recommendations},
  author={Herlocker, Jonathan L and Konstan, Joseph A and Riedl, John},
  booktitle={Proceedings of the 2000 ACM Conference on Computer Supported Sooperative Work},
  pages={241--250},
  year={2000}
}

@article{hou2019explainable,
  title={Explainable recommendation with fusion of aspect information},
  author={Hou, Yunfeng and Yang, Ning and Wu, Yi and Yu, Philip S},
  journal={World Wide Web},
  volume={22},
  number={1},
  pages={221--240},
  year={2019},
  publisher={Springer}
}

@article{musto2021generating,
  title={Generating post hoc review-based natural language justifications for recommender systems: C. Musto et al.},
  author={Musto, Cataldo and de Gemmis, Marco and Lops, Pasquale and Semeraro, Giovanni},
  journal={User Modeling and User-Adapted Interaction},
  volume={31},
  number={3},
  pages={629--673},
  year={2021},
  publisher={Springer}
}

@inproceedings{biran2017explanation,
  title={Explanation and justification in machine learning: A survey},
  author={Biran, Or and Cotton, Courtenay},
  booktitle={IJCAI-17 Workshop on Explainable AI (XAI)},
  volume={8},
  number={1},
  pages={8--13},
  year={2017}
}

@article{petty1997elaboration,
  title={The elaboration likelihood model: Implications for the practice of school psychology},
  author={Petty, Richard E and Heesacker, Martin and Hughes, Jan N},
  journal={Journal of School Psychology},
  volume={35},
  number={2},
  pages={107--136},
  year={1997},
  publisher={Elsevier}
}

@inproceedings{reimers2019sentence,
  title={Sentence-bert: Sentence embeddings using siamese bert-networks},
  author={Reimers, Nils and Gurevych, Iryna},
  booktitle={Proceedings of the 2019 Conference on Empirical Methods in Natural Language Processing and the 9th International Joint Conference on Natural Language Processing (EMNLP-IJCNLP)},
  pages={3982--3992},
  year={2019}
}

@inproceedings{chang2016crowd,
  title={Crowd-based personalized natural language explanations for recommendations},
  author={Chang, Shuo and Harper, F Maxwell and Terveen, Loren Gilbert},
  booktitle={Proceedings of the 10th ACM Conference on Recommender Systems},
  pages={175--182},
  year={2016}
}

@article{jannach2021survey,
  title={A survey on conversational recommender systems},
  author={Jannach, Dietmar and Manzoor, Ahtsham and Cai, Wanling and Chen, Li},
  journal={ACM Computing Surveys (CSUR)},
  volume={54},
  number={5},
  pages={1--36},
  year={2021},
  publisher={ACM New York, NY, USA}
}

@article{gao2021advances,
  title={Advances and challenges in conversational recommender systems: A survey},
  author={Gao, Chongming and Lei, Wenqiang and He, Xiangnan and De Rijke, Maarten and Chua, Tat-Seng},
  journal={AI Open},
  volume={2},
  pages={100--126},
  year={2021},
  publisher={Elsevier}
}

@inproceedings{kouki2019personalized,
  title={Personalized explanations for hybrid recommender systems},
  author={Kouki, Pigi and Schaffer, James and Pujara, Jay and O'Donovan, John and Getoor, Lise},
  booktitle={Proceedings of the 24th International Conference on Intelligent User Interfaces},
  pages={379--390},
  year={2019}
}

@article{alslaity2021users,
  title={Users’ responsiveness to persuasive techniques in recommender systems},
  author={Alslaity, Alaa and Tran, Thomas},
  journal={Frontiers in Artificial Intelligence},
  volume={4},
  pages={679459},
  year={2021},
  publisher={Frontiers Media SA}
}

@inproceedings{gkika2014investigating,
  title={Investigating the effectiveness of persuasion strategies on recommender systems},
  author={Gkika, Sofia and Lekakos, George},
  booktitle={2014 9th International Workshop on Semantic and Social Media Adaptation and Personalization},
  pages={94--97},
  year={2014},
  organization={IEEE}
}

@inproceedings{oshchepkov2026prompting,
  title={How Prompting Shapes LLM-Generated Explanations for Recommender Systems: A Multi-Prompt Comparison Across Domains},
  author={Oshchepkov, Hlib and Tommasel, Antonela},
  booktitle={Joint Proceedings of the ACM UMAP Workshops 2026, UMAP 2026},
  year={2026}
}

@article{demarque2015nudging,
  title={Nudging sustainable consumption: The use of descriptive norms to promote a minority behavior in a realistic online shopping environment},
  author={Demarque, Christophe and Charalambides, Laetitia and Hilton, Denis J and Waroquier, Laurent},
  journal={Journal of Environmental Psychology},
  volume={43},
  pages={166--174},
  year={2015},
  publisher={Elsevier}
}

@article{allcott2011social,
  title={Social norms and energy conservation},
  author={Allcott, Hunt},
  journal={Journal of Public Economics},
  volume={95},
  number={9-10},
  pages={1082--1095},
  year={2011},
  publisher={Elsevier}
}

@article{zhang2022green,
  title={Green product types modulate green consumption in the gain and loss framings: An event-related potential study},
  author={Zhang, Guanfei and Li, Mei and Li, Jin and Tan, Min and Li, Huie and Zhong, Yiping},
  journal={International Journal of Environmental Research and Public Health},
  volume={19},
  number={17},
  pages={10746},
  year={2022},
  publisher={MDPI}
}

@article{farshbafiyan2025framing,
  title={The framing effect and sustainable hotel booking behaviour: A nudge marketing study},
  author={Farshbafiyan Hosseininezhad, Mirfarshad and Heidari, Majid and Letizia Guerra, Maria and others},
  journal={European Journal of Tourism Research},
  volume={39},
  pages={1--14},
  year={2025}
}

@article{symeonidis2008providing,
  title={Providing justifications in recommender systems},
  author={Symeonidis, Panagiotis and Nanopoulos, Alexandros and Manolopoulos, Yannis},
  journal={IEEE Transactions on Systems, Man, and Cybernetics-Part A: Systems and Humans},
  volume={38},
  number={6},
  pages={1262--1272},
  year={2008},
  publisher={IEEE}
}

@article{kaptein2015personalizing,
  title={Personalizing persuasive technologies: Explicit and implicit personalization using persuasion profiles},
  author={Kaptein, Maurits and Markopoulos, Panos and De Ruyter, Boris and Aarts, Emile},
  journal={International Journal of Human-Computer Studies},
  volume={77},
  pages={38--51},
  year={2015},
  publisher={Elsevier}
}

@inproceedings{rostami2025recommender,
  title={Recommender Systems for Sustainable Development through Responsible Nudging},
  author={Rostami, Mehrdad and Felfernig, Alexander and W{\"o}rndl, Wolfgang and Oussalah, Mourad and Anand, Avishek and Jalili, Mahdi and Banerjee, Ashmi},
  booktitle={Proceedings of the 34th ACM International Conference on Information and Knowledge Management},
  pages={6923--6926},
  year={2025}
}

@article{kahnemanMapsBoundedRationality2003,
	title = {Maps of {Bounded} {Rationality}: {Psychology} for {Behavioral} {Economics}},
	volume = {93},
	issn = {00028282},
	number = {5},
	journal = {The American Economic Review},
	publisher = {American Economic Association},
	author = {Kahneman, Daniel},
	year = {2003},
	pages = {1449--1475},
}

@online{statista2025ecommerceshare,
  author  = {{Statista}},
  title   = {E-commerce as share of total retail sales worldwide 2017--2030},
  year    = {2025},
  url     = {https://www.statista.com/statistics/534123/e-commerce-share-of-retail-sales-worldwide/},
  urldate = {2026-06-23}
}

@inproceedings{chen2014sentiment,
  title={Sentiment-enhanced explanation of product recommendations},
  author={Chen, Li and Wang, Feng},
  booktitle={Proceedings of the 23rd International Conference on World Wide Web},
  pages={239--240},
  year={2014}
}

@inproceedings{roitman2010increasing,
  title={Increasing patient safety using explanation-driven personalized content recommendation},
  author={Roitman, Haggai and Messika, Yossi and Tsimerman, Yevgenia and Maman, Yonatan},
  booktitle={Proceedings of the 1st ACM International Health Informatics Symposium},
  pages={430--434},
  year={2010}
}

@inproceedings{terano1989cses,
  title={CSES: an approach to integrating graphic, music and voice information into a user-friendly interface},
  author={Terano, Takao and Suzuki, Michio and Onoda, Takashi and Uenishi, K and Matsuura, T},
  booktitle={International Workshop on Industrial Applications of Machine Intelligence and Vision,},
  pages={349--354},
  year={1989},
  organization={IEEE}
}

@article{fornell1981evaluating,
  title={Evaluating structural equation models with unobservable variables and measurement error},
  author={Fornell, Claes and Larcker, David F},
  journal={Journal of Marketing Research},
  volume={18},
  number={1},
  pages={39--50},
  year={1981},
  publisher={Sage Publications Sage CA: Los Angeles, CA}
}

@article{gefen2005practical,
  title={A practical guide to factorial validity using PLS-Graph: Tutorial and annotated example},
  author={Gefen, David and Straub, Detmar},
  journal={Communications of the Association for Information Systems},
  volume={16},
  number={1},
  pages={5},
  year={2005}
}

@article{eisinga2013reliability,
  title={The reliability of a two-item scale: Pearson, Cronbach, or Spearman-Brown?},
  author={Eisinga, Rob and Grotenhuis, Manfred te and Pelzer, Ben},
  journal={International Journal of Public Health},
  volume={58},
  number={4},
  pages={637--642},
  year={2013},
  publisher={Springer}
}

@article{faul2007g,
  title={G* Power 3: A flexible statistical power analysis program for the social, behavioral, and biomedical sciences},
  author={Faul, Franz and Erdfelder, Edgar and Lang, Albert-Georg and Buchner, Axel},
  journal={Behavior Research Methods},
  volume={39},
  number={2},
  pages={175--191},
  year={2007},
  publisher={Springer}
}

@article{lee2025ecolabels,
  title   = {Developing ecolabels to encourage sustainable eating in restaurants: A randomized experiment},
  author  = {Lee, Cristina J. Y. and Petimar, Joshua and Zeitlin, Amanda B. and Collis, Caroline and Cleveland, Lauren and Musicus, Aviva A. and Grummon, Anna H.},
  journal = {PLOS ONE},
  year    = {2025},
  doi     = {10.1371/journal.pone.0335724}
}

@inproceedings{hou2026bridging,
  title={Bridging Language and Items for Retrieval and Recommendation: Benchmarking LLMs as Semantic Encoders},
  author={Hou, Yupeng and Li, Jiacheng and Fu, Xiangjun and He, Zhankui and Yan, An and Chen, Xiusi and McAuley, Julian},
  booktitle={Proceedings of the 64th Annual Meeting of the Association for Computational Linguistics (Volume 1: Long Papers)},
  pages={3251--3265},
  year={2026}
}

@article{norton2022exploring,
  title={Exploring consumers’ understanding and perception of sustainable food packaging in the UK},
  author={Norton, Victoria and Waters, Carys and Oloyede, Omobolanle O and Lignou, Stella},
  journal={Foods},
  volume={11},
  number={21},
  pages={3424},
  year={2022},
  publisher={MDPI}
}

@article{bocoli2025perspectives,
  title={Perspectives on Eco-Friendly food packaging: challenges, solutions, and trends},
  author={B{\'o}coli, Paula Fernanda Janetti and Gomes, Vitor Emanuel de Souza and Maia, Amanda Alves Domingos and Marangoni J{\'u}nior, Lu{\'\i}s},
  journal={Foods},
  volume={14},
  number={17},
  pages={3062},
  year={2025},
  publisher={MDPI}
}

@article{gilardi2023chatgpt,
  title={ChatGPT outperforms crowd workers for text-annotation tasks},
  author={Gilardi, Fabrizio and Alizadeh, Meysam and Kubli, Ma{\"e}l},
  journal={Proceedings of the National Academy of Sciences},
  volume={120},
  number={30},
  pages={e2305016120},
  year={2023},
  publisher={National Academy of Sciences}
}

@article{jannach2015item,
  title={Item familiarity as a possible confounding factor in user-centric recommender systems evaluation},
  author={Jannach, Dietmar and Lerche, Lukas and Jugovac, Michael},
  journal={I-Com},
  volume={14},
  number={1},
  pages={29--39},
  year={2015},
  publisher={De Gruyter Oldenbourg}
}

@article{martinez2018customer,
  title={Customer responses to environmentally certified hotels: The moderating effect of environmental consciousness on the formation of behavioral intentions},
  author={Mart{\'\i}nez Garc{\'\i}a de Leaniz, Patricia and Herrero Crespo, {\'A}ngel and G{\'o}mez L{\'o}pez, Raquel},
  journal={Journal of Sustainable Tourism},
  volume={26},
  number={7},
  pages={1160--1177},
  year={2018},
  publisher={Taylor \& Francis}
}

@article{homar2021effects,
  title={The effects of framing on environmental decisions: A systematic literature review},
  author={Homar, Aja Ropret and Cvelbar, Ljubica Kne{\v{z}}evi{\'c}},
  journal={Ecological Economics},
  volume={183},
  pages={106950},
  year={2021},
  publisher={Elsevier}
}

@article{segev2015effects,
  title={The effects of gain versus loss message framing and point of reference on consumer responses to green advertising},
  author={Segev, Sigal and Fernandes, Juliana and Wang, Weirui},
  journal={Journal of Current Issues \& Research in Advertising},
  volume={36},
  number={1},
  pages={35--51},
  year={2015},
  publisher={Taylor \& Francis}
}

@article{rudin2019stop,
  title={Stop explaining black box machine learning models for high stakes decisions and use interpretable models instead},
  author={Rudin, Cynthia},
  journal={Nature Machine Intelligence},
  volume={1},
  number={5},
  pages={206--215},
  year={2019},
  publisher={Nature Publishing Group UK London}
}

@article{iyengar2000choice,
  title={When choice is demotivating: Can one desire too much of a good thing?},
  author={Iyengar, Sheena S and Lepper, Mark R},
  journal={Journal of Personality and Social Psychology},
  volume={79},
  number={6},
  pages={995},
  year={2000},
  publisher={American Psychological Association}
}

@article{cowan2004constant,
  title={Constant capacity in an immediate serial-recall task: A logical sequel to Miller (1956)},
  author={Cowan, Nelson and Chen, Zhijian and Rouder, Jeffrey N},
  journal={Psychological Science},
  volume={15},
  number={9},
  pages={634--640},
  year={2004},
  publisher={SAGE Publications Sage CA: Los Angeles, CA}
}

@inproceedings{bollen2010understanding,
  title={Understanding choice overload in recommender systems},
  author={Bollen, Dirk and Knijnenburg, Bart P and Willemsen, Martijn C and Graus, Mark},
  booktitle={Proceedings of the fourth ACM Conference on Recommender Systems},
  pages={63--70},
  year={2010}
}

@article{palan2018prolific,
  title={Prolific. ac—A subject pool for online experiments},
  author={Palan, Stefan and Schitter, Christian},
  journal={Journal of Behavioral and Experimental Finance},
  volume={17},
  pages={22--27},
  year={2018},
  publisher={Elsevier}
}

@article{oppenheimer2009instructional,
  title={Instructional manipulation checks: Detecting satisficing to increase statistical power},
  author={Oppenheimer, Daniel M and Meyvis, Tom and Davidenko, Nicolas},
  journal={Journal of Experimental Social Psychology},
  volume={45},
  number={4},
  pages={867--872},
  year={2009},
  publisher={Elsevier}
}

@article{aguinis2013best,
  title={Best-practice recommendations for defining, identifying, and handling outliers},
  author={Aguinis, Herman and Gottfredson, Ryan K and Joo, Harry},
  journal={Organizational Research Methods},
  volume={16},
  number={2},
  pages={270--301},
  year={2013},
  publisher={Sage Publications Sage CA: Los Angeles, CA}
}

@book{hosmer2013applied,
  author    = {Hosmer, David W. and Lemeshow, Stanley and Sturdivant, Rodney X.},
  title     = {Applied Logistic Regression},
  edition   = {3rd},
  publisher = {Wiley},
  year      = {2013}
}

@article{chen2012critiquing,
  author  = {Chen, Li and Pu, Pearl},
  title   = {Critiquing-based recommenders: survey and emerging trends},
  journal = {User Modeling and User-Adapted Interaction},
  volume  = {22},
  number  = {1--2},
  pages   = {125--150},
  year    = {2012}
}

@article{stillwell1981comparison,
  title={A comparison of weight approximation techniques in multiattribute utility decision making},
  author={Stillwell, William G and Seaver, David A and Edwards, Ward},
  journal={Organizational Behavior and Human Performance},
  volume={28},
  number={1},
  pages={62--77},
  year={1981},
  publisher={Elsevier}
}

@article{barron1996decision,
  title={Decision quality using ranked attribute weights},
  author={Barron, F Hutton and Barrett, Bruce E},
  journal={Management Science},
  volume={42},
  number={11},
  pages={1515--1523},
  year={1996},
  publisher={INFORMS}
}

@article{panickssery2024llm,
  title={Llm evaluators recognize and favor their own generations},
  author={Panickssery, Arjun and Bowman, Samuel R and Feng, Shi},
  journal={Advances in Neural Information Processing Systems},
  volume={37},
  pages={68772--68802},
  year={2024}
}

@article{kollmuss2002mind,
  title={Mind the gap: why do people act environmentally and what are the barriers to pro-environmental behavior?},
  author={Kollmuss, Anja and Agyeman, Julian},
  journal={Environmental Education Research},
  volume={8},
  number={3},
  pages={239--260},
  year={2002},
  publisher={Taylor \& Francis}
}

@article{white2019shift,
  title={How to SHIFT consumer behaviors to be more sustainable: A literature review and guiding framework},
  author={White, Katherine and Habib, Rishad and Hardisty, David J},
  journal={Journal of Marketing},
  volume={83},
  number={3},
  pages={22--49},
  year={2019},
  publisher={Sage Publications Sage CA: Los Angeles, CA}
}

@article{bruns2018can,
  title={Can nudges be transparent and yet effective?},
  author={Bruns, Hendrik and Kantorowicz-Reznichenko, Elena and Klement, Katharina and Jonsson, Marijane Luistro and Rahali, Bilel},
  journal={Journal of Economic Psychology},
  volume={65},
  pages={41--59},
  year={2018},
  publisher={Elsevier}
}

@inproceedings{caraban201923,
  title={23 ways to nudge: A review of technology-mediated nudging in human-computer interaction},
  author={Caraban, Ana and Karapanos, Evangelos and Gon{\c{c}}alves, Daniel and Campos, Pedro},
  booktitle={Proceedings of the 2019 CHI Conference on Human Factors in Computing Systems},
  pages={1--15},
  year={2019}
}

@article{hummel2019effective,
  title={How effective is nudging? A quantitative review on the effect sizes and limits of empirical nudging studies},
  author={Hummel, Dennis and Maedche, Alexander},
  journal={Journal of Behavioral and Experimental Economics},
  volume={80},
  pages={47--58},
  year={2019},
  publisher={Elsevier}
}

@article{montgomery2018conditioning,
  title={How conditioning on posttreatment variables can ruin your experiment and what to do about it},
  author={Montgomery, Jacob M and Nyhan, Brendan and Torres, Michelle},
  journal={American Journal of Political Science},
  volume={62},
  number={3},
  pages={760--775},
  year={2018},
  publisher={Wiley Online Library}
}

@article{rosenbaum1984consequences,
  title={The consequences of adjustment for a concomitant variable that has been affected by the treatment},
  author={Rosenbaum, Paul R},
  journal={Journal of the Royal Statistical Society Series A: Statistics in Society},
  volume={147},
  number={5},
  pages={656--666},
  year={1984},
  publisher={Oxford University Press}
}

@article{cicchetti1990high,
  title={High agreement but low kappa: II. Resolving the paradoxes},
  author={Cicchetti, Domenic V and Feinstein, Alvan R},
  journal={Journal of Clinical Epidemiology},
  volume={43},
  number={6},
  pages={551--558},
  year={1990},
  publisher={Elsevier}
}

@article{mills2022personalized,
  title={Personalized nudging},
  author={Mills, Stuart},
  journal={Behavioural Public Policy},
  volume={6},
  number={1},
  pages={150--159},
  year={2022},
  publisher={Cambridge University Press}
}

@article{soler2010point,
  title={Point-of-decision prompts to increase stair use: a systematic review update},
  author={Soler, Robin E and Leeks, Kimberly D and Buchanan, Leigh Ramsey and Brownson, Ross C and Heath, Gregory W and Hopkins, David H and Task Force on Community Preventive Services and others},
  journal={American Journal of Preventive Medicine},
  volume={38},
  number={2},
  pages={S292--S300},
  year={2010},
  publisher={Elsevier}
}

@article{ji2023survey,
  title={Survey of hallucination in natural language generation},
  author={Ji, Ziwei and Lee, Nayeon and Frieske, Rita and Yu, Tiezheng and Su, Dan and Xu, Yan and Ishii, Etsuko and Bang, Ye Jin and Madotto, Andrea and Fung, Pascale},
  journal={ACM Computing Surveys},
  volume={55},
  number={12},
  pages={1--38},
  year={2023},
  publisher={ACM New York, NY}
}

@article{matz2024potential,
  title={The potential of generative AI for personalized persuasion at scale},
  author={Matz, Sandra C and Teeny, Jacob D and Vaid, Sumer S and Peters, Heinrich and Harari, Gabriella M and Cerf, Moran},
  journal={Scientific Reports},
  volume={14},
  number={1},
  pages={4692},
  year={2024},
  publisher={Nature Publishing Group UK London}
}

@article{schultz2007constructive,
  title={The constructive, destructive, and reconstructive power of social norms},
  author={Schultz, P Wesley and Nolan, Jessica M and Cialdini, Robert B and Goldstein, Noah J and Griskevicius, Vladas},
  journal={Psychological Science},
  volume={18},
  number={5},
  pages={429--434},
  year={2007},
  publisher={SAGE Publications Sage CA: Los Angeles, CA}
}

@article{loewenstein2015warning,
  title={Warning: You are about to be nudged},
  author={Loewenstein, George and Bryce, Cindy and Hagmann, David and Rajpal, Sachin},
  journal={Behavioral Science \& Policy},
  volume={1},
  number={1},
  pages={35--42},
  year={2015},
  publisher={SAGE Publications Sage CA: Los Angeles, CA}
}

@article{ziems2024can,
  title={Can large language models transform computational social science?},
  author={Ziems, Caleb and Held, William and Shaikh, Omar and Chen, Jiaao and Zhang, Zhehao and Yang, Diyi},
  journal={Computational Linguistics},
  volume={50},
  number={1},
  pages={237--291},
  year={2024}
}

@article{bail2024can,
  title={Can generative AI improve social science?},
  author={Bail, Christopher A},
  journal={Proceedings of the National Academy of Sciences},
  volume={121},
  number={21},
  pages={e2314021121},
  year={2024},
  publisher={National Academy of Sciences}
}

@article{de2020concepts,
  title={Concepts and forms of greenwashing: A systematic review},
  author={de Freitas Netto, Sebasti{\~a}o Vieira and Sobral, Marcos Felipe Falc{\~a}o and Ribeiro, Ana Regina Bezerra and Soares, Gleibson Robert da Luz},
  journal={Environmental Sciences Europe},
  volume={32},
  number={1},
  pages={19},
  year={2020},
  publisher={Springer}
}

@book{sunstein2016ethics,
  title={The ethics of influence: Government in the age of behavioral science},
  author={Sunstein, Cass R},
  year={2016},
  publisher={Cambridge University Press}
}

@article{sunstein2017nudges,
  title={Nudges that fail},
  author={Sunstein, Cass R},
  journal={Behavioural Public Policy},
  volume={1},
  number={1},
  pages={4--25},
  year={2017},
  publisher={Cambridge University Press}
}

@article{allcott2014short,
  title={The short-run and long-run effects of behavioral interventions: Experimental evidence from energy conservation},
  author={Allcott, Hunt and Rogers, Todd},
  journal={American Economic Review},
  volume={104},
  number={10},
  pages={3003--3037},
  year={2014},
  publisher={American Economic Association 2014 Broadway, Suite 305, Nashville, TN 37203}
}

@article{cochran1954,
  author  = {Cochran, William G.},
  title   = {Some Methods for Strengthening the Common $\chi^2$ Tests},
  journal = {Biometrics},
  volume  = {10},
  number  = {4},
  pages   = {417--451},
  year    = {1954},
}

@article{fisher1922interpretation,
  title={On the interpretation of $\chi$ 2 from contingency tables, and the calculation of P},
  author={Fisher, Ronald A},
  journal={Journal of the Royal Statistical Society},
  volume={85},
  number={1},
  pages={87--94},
  year={1922},
  publisher={JSTOR}
}

@book{cohen2013statistical,
  title={Statistical power analysis for the behavioral sciences},
  author={Cohen, Jacob},
  year={2013},
  publisher={routledge}
}

@article{potter2021effects,
  title={The effects of environmental sustainability labels on selection, purchase, and consumption of food and drink products: a systematic review},
  author={Potter, Christina and Bastounis, Anastasios and Hartmann-Boyce, Jamie and Stewart, Cristina and Frie, Kerstin and Tudor, Kate and Bianchi, Filippo and Cartwright, Emma and Cook, Brian and Rayner, Mike and others},
  journal={Environment and Behavior},
  volume={53},
  number={8},
  pages={891--925},
  year={2021},
  publisher={Sage Publications Sage CA: Los Angeles, CA}
}

@book{schultz2002knowledge,
  title={Knowledge, information and household recycling: Examining the knowledge-deficit model},
  author={Schultz, W and Dietz, T and Stern, P},
  year={2002},
  publisher={National Academy of Sciences Washington, DC}
}

@article{dellavigna2022rcts,
  title={RCTs to scale: Comprehensive evidence from two nudge units},
  author={DellaVigna, Stefano and Linos, Elizabeth},
  journal={Econometrica},
  volume={90},
  number={1},
  pages={81--116},
  year={2022},
  publisher={Wiley Online Library}
}

@article{hirsh2012personalized,
  title={Personalized persuasion: Tailoring persuasive appeals to recipients’ personality traits},
  author={Hirsh, Jacob B and Kang, Sonia K and Bodenhausen, Galen V},
  journal={Psychological Science},
  volume={23},
  number={6},
  pages={578--581},
  year={2012},
  publisher={Sage Publications Sage CA: Los Angeles, CA}
}

@article{alves2026exploring,
  title={Exploring Personality-Aware Explanations for Recommender Systems},
  author={Alves, Gabrielle and Jannach, Dietmar and Soares de Souza, Luan and Manzato, Marcelo},
  journal={ACM Transactions on Recommender Systems},
  year={2026},
  publisher={ACM New York, NY}
}

@article{alves2024digitally,
  title={Digitally nudging users to explore off-profile recommendations: here be dragons: G. Alves et al.},
  author={Alves, Gabrielle and Jannach, Dietmar and de Souza, Rodrigo Ferrari and Damian, Daniela and Manzato, Marcelo Garcia},
  journal={User Modeling and User-Adapted Interaction},
  volume={34},
  number={2},
  pages={441--481},
  year={2024},
  publisher={Springer}
}

@article{bettman1998constructive,
  author  = {Bettman, James R. and Luce, Mary Frances and Payne, John W.},
  title   = {Constructive Consumer Choice Processes},
  journal = {Journal of Consumer Research},
  volume  = {25}, number = {3}, pages = {187--217}, year = {1998}
}

\appendix
\section{Appendix}

The appendix collects the supporting analyses referenced in the main text. 
Appendices~\ref{app:balance_low} and~\ref{app:balance_high} report the 
attribute-balance tests confirming that the higher- and lower-sustainability item 
pools were matched on the recommendation features in the low- and high-involvement 
domains, respectively.  
Appendix~\ref{app:measurement-validation} documents the reliability and 
validity of the five subjective evaluation measures on the pooled sample.
Appendix~\ref{app:length} presents the robustness analysis showing that explanation length does not account for the 
choice effects.
Appendix~\ref{app:randomization-per-domain} gives the 
per-domain randomization checks that complement the pooled balance. 
Appendix~\ref{app:subjective-models} reports the full covariate-adjusted regression models underlying their condition contrasts. Finally,
Appendix~\ref{app:per-domain-figures} shows the per-domain results descriptively.

\FloatBarrier

\subsection{Attribute Balance in the Low-Involvement Domain}
\label{app:balance_low}
This appendix reports the full attribute-balance test for the instant-coffee study, summarized in Section~\ref{study1:dataset}. To confirm that the matching procedure produced comparable pools, we tested whether the higher- and lower-sustainability tiers differ on the four recommendation features, using a chi-square test of independence together with Cram\'er's $V$ as an effect-size measure (Table~\ref{tab:balance_coffee}). All associations are small ($V \leq .151$) and none is statistically significant ($p \geq .350$), confirming that the two pools are well matched on product features, so any tier difference in choice reflects the sustainability attribute rather than feature imbalance.

\begin{table}[ht]
  \centering
  \caption{Attribute balance between the higher- and lower-sustainability instant-coffee pools ($n=46$ per tier), across the four recommendation features. For each feature we report its value set together with the chi-square statistic, $p$-value, and Cram\'er's $V$ from a chi-square test of independence.}
  \label{tab:balance_coffee}
  \begin{tabular}{l l r r r}
    \toprule
    \textbf{Feature} & \textbf{Values} & \textbf{$\chi^2$} & \textbf{$p$} & \textbf{Cram\'er's $V$} \\
    \midrule
    \midrule
    Roast level & light, medium, dark & 1.311 & .519 & .119 \\
    Bean type   & Arabica, blends \& other beans & 0.024 & .877 & .017 \\
    Caffeine    & caffeinated, decaffeinated & 0.000 & 1.000 & .000 \\
    Intensity   & balanced, bold, smooth & 2.100 & .350 & .151 \\
    \bottomrule
  \end{tabular}
\end{table}

\FloatBarrier

\subsection{Attribute Balance in the High-Involvement Domain}
\label{app:balance_high}
This appendix reports the full attribute-balance test for the hotel study, 
summarized in Section~\ref{study2:dataset}. We tested whether the higher- and 
lower-sustainability tiers differ on each facility attribute, using a chi-square test of independence with Cram\'er's $V$ (Table~\ref{tab:balance_hotel}); for the rarely offered elevator, whose minimum expected cell count fell below five, we used Fisher's exact test instead. All associations are small ($V \leq .085$) and none is statistically significant ($p \geq .238$), confirming that the two pools are well matched on facility attributes.

\begin{table}[ht]
  \centering
  \caption{Attribute balance between the higher- and lower-sustainability hotel pools ($n=97$ per tier), across all facility attributes. All attributes are binary (offered / not offered) except distance to city center, which has two coarse levels. Reported are the chi-square statistic, $p$-value, and Cram\'er's $V$ from a chi-square test of independence per attribute. \textsuperscript{\dag}Tested with Fisher's exact test (two-sided) because its minimum expected cell count fell below five. Wi-Fi is offered by every hotel and is therefore constant across tiers.}
  \label{tab:balance_hotel}
  \begin{tabular}{l r r r}
    \toprule
    \textbf{Attribute} & \textbf{$\chi^2$} & \textbf{$p$} & \textbf{Cram\'er's $V$} \\
    \midrule
    \midrule
    Distance to city centre & 0.000 & 1.000 & .000 \\
    Air conditioning        & 0.000 & 1.000 & .000 \\
    Bar                     & 0.031 & .859 & .013 \\
    Fitness / trainer offers & 0.463 & .496 & .049 \\
    Elevator                & -- & 1.000\textsuperscript{\dag} & -- \\
    Fully wheelchair accessible & 0.376 & .540 & .044 \\
    Garage / parking        & 0.520 & .471 & .052 \\
    Garden                  & 0.566 & .452 & .054 \\
    Jacuzzi                 & 0.385 & .535 & .045 \\
    Massage                 & 0.000 & 1.000 & .000 \\
    Minibar / refrigerator  & 0.000 & 1.000 & .000 \\
    Pool                    & 0.000 & 1.000 & .000 \\
    Pets allowed            & 0.000 & 1.000 & .000 \\
    Restaurant              & 1.029 & .310 & .073 \\
    Co-working space        & 1.394 & .238 & .085 \\
    Sauna                   & 0.383 & .536 & .044 \\
    Smart TV (streaming)    & 0.231 & .631 & .035 \\
    Wi-Fi                   & \multicolumn{3}{c}{constant (100\% in both tiers)} \\
    \bottomrule
  \end{tabular}
\end{table}

\FloatBarrier

\subsection{Validation of Subjective Evaluation Constructs }
\label{app:measurement-validation}

Before turning to the substantive results, we validate the self-reported subjective evaluation constructs (explanation clarity, explanation usability, sustainability impact, choice satisfaction, and choice difficulty) as introduced in Section~\ref{sec:measures}. Reliability was assessed with Cronbach's $\alpha$ for constructs with three or more items and with the Spearman-Brown coefficient for the two-item explanation clarity construct, the recommended statistic for two-item scales \cite{eisinga2013reliability}. All constructs met conventional reliability thresholds ($\alpha \geq .76$). We evaluated convergent and discriminant validity following established criteria. For convergent validity, all standardized factor loadings exceeded .60, composite reliability (CR) exceeded the recommended threshold of .70, and the average variance extracted (AVE) exceeded .50 for every construct \cite{fornell1981evaluating}. For discriminant validity, a joint exploratory factor analysis confirmed that every item loaded more strongly on its assigned construct than on any other factor \cite{gefen2005practical}, and the square root of each construct's AVE exceeded its correlations with all other constructs, satisfying the Fornell-Larcker criterion \cite{fornell1981evaluating}. Table~\ref{tab:measures} reports items, loadings, and reliability statistics. Explanation usability was measured with a single item and is therefore reported without construct-level statistics.

\begin{table*}[th]
  \centering
  \caption{Measurement instruments, factor loadings, and reliability statistics. All items were assessed on a 7-point Likert scale (1 = strongly disagree, 7 = strongly agree). Reliability is reported as Cronbach's $\alpha$ for constructs with three or more items and as the Spearman-Brown coefficient ($\rho$) for the two-item construct. CR = composite reliability; AVE = average variance extracted; (R) = reverse-coded.}
  \label{tab:measures}
  \begin{tabular}{p{3.8cm} p{9.3cm} p{1cm}}
    \toprule
    \textbf{Construct} & \textbf{Item} & \textbf{Loadings} \\
    \midrule
    \midrule
    \parbox[t]{3.8cm}{\textbf{Choice satisfaction} \\ \cite{knijnenburg2015evaluating,starke2021promoting} \\ $\alpha$=.82, CR=.91, AVE=.76} &
    \parbox[t]{9.3cm}{I like the product/hotel I have chosen. \\
      I think I would enjoy the product/stay at the hotel I have chosen. \\
      I would recommend the chosen product/hotel to others.} &
    \parbox[t]{1cm}{.92 \\ .88 \\ .82} \\[1ex]
    \midrule
    \parbox[t]{3.8cm}{\textbf{Choice difficulty} \\ \cite{willemsen2016understanding,el2022nudging} \\ $\alpha$=.76, CR=.86, AVE=.67} &
    \parbox[t]{9.3cm}{I changed my mind several times before making a decision. \\
      Making a choice was overwhelming. \\
      (R) It was easy to make this choice.} &
    \parbox[t]{1cm}{.82 \\ .87 \\ .77} \\[1ex]
    \midrule
    \parbox[t]{3.8cm}{\textbf{Explanation clarity} \\ \cite{lubos2024llm,peake2018explanation,tintarev2022beyond,tintarev2007explanations} \\ $\rho$=.83, CR=.92, AVE=.86} &
    \parbox[t]{9.3cm}{The explanations were easy to understand. \\
      The explanations made it clear why the products/hotels were recommended.} &
    \parbox[t]{1cm}{.93 \\ .93} \\[1ex]
    \midrule
 \parbox[t]{3.8cm}{\textbf{Sustainability impact} \\ \cite{tran2024less,chen2014sentiment} \\ $\alpha$=.93, CR=.96, AVE=.88} &
    \parbox[t]{9.3cm}{The explanations were convincing in communicating sustainability aspects of the recommended products/hotels. \\
      The explanations increased my awareness of sustainability aspects. \\
      The explanations helped me understand the sustainability aspects of the recommended products/hotels.} &
    \parbox[t]{1cm}{.94 \\ .93 \\ .94} \\[1ex]
    \midrule
    \parbox[t]{3.8cm}{\textbf{Explanation usability} \emph{(single item)} \\ \cite{tintarev2012evaluating,el2025nudging}} &
    \parbox[t]{9.3cm}{The explanations helped me choose among the recommended products/hotels.} &
    \parbox[t]{1cm}{--} \\[1ex]
    \midrule
    \parbox[t]{3.8cm}{\textbf{Biospheric value orientation} \\ \cite{de2007value} \\ $\alpha$=.94, CR=.96, AVE=.85} &
    \parbox[t]{9.3cm}{Preventing pollution is a guiding principle in my life. \\
      Respecting the earth is a guiding principle in my life. \\
      Unity with nature is a guiding principle in my life. \\
      Protecting the environment is a guiding principle in my life.} &
    \parbox[t]{1cm}{.92 \\ .93 \\ .90 \\ .93} \\[1ex]
    \bottomrule
  \end{tabular}
\end{table*}

\FloatBarrier

\subsection{Robustness to Explanation Length}
\label{app:length}

As a robustness check for the word-count differences noted in 
Section~\ref{sec:stimulus-validation}, we tested whether explanation length predicted choice. Within the three sustainability-aware conditions, the participant-level word-count asymmetry between the higher- and lower-sustainability explanations did not predict sustainable choice (OR $= 1.24$, 95\% CI $[0.84, 1.84]$, $p = .28$); moreover, the upper bound of this interval lies below the estimated framing and descriptive-social-norm effects, so even the largest length effect compatible with our data is too small to account for the presentation effects. The pattern across conditions points the same way: the mechanism-free condition produced the \emph{shortest} explanations ($M = 29.0$ words) yet left choices at chance, as did the longer preference-only baseline ($M = 34.1$)---a non-monotonic relationship that length alone cannot generate.
We therefore report length as a robustness check rather than adjusting for it in the main models. As noted in Section~\ref{sec:stimulus-validation}, length is not a nuisance variable but part of the treatment itself: a mechanism-based explanation is longer precisely because it frames a benefit or invokes a peer norm. Controlling for it would mean conditioning on a post-treatment quantity and stripping out part of the manipulation 
\cite{montgomery2018conditioning,rosenbaum1984consequences}.

\FloatBarrier

\subsection{Randomization Checks per Involvement Domain}
\label{app:randomization-per-domain}
Because random assignment was performed within each study, 
Table~\ref{tab:rand-per-domain} reports the condition-balance tests separately for each involvement domain, alongside the pooled test from
Section~\ref{sec:checks}. No characteristic differed across conditions in either domain. Only a handful of participants classified 
themselves as novices or experts in domain expertise---particularly among the hotel bookers in the high-involvement study---leaving several cells too sparse for a reliable chi-square test. We therefore assessed its balance with a Monte Carlo permutation test, which likewise found no differences across conditions in any subset.
\begin{table}[h]
\centering
\caption{Condition-balance tests for participant characteristics across the four 
conditions, pooled and per involvement domain. Non-significant results indicate 
successful randomization. Entries are $\chi^2$ (df), $p$. 
\textsuperscript{\dag}Domain expertise was tested with a Monte Carlo permutation 
test (5{,}000 resamples), as its sparse cells make the chi-square approximation 
unreliable; no degrees of freedom are reported.}
\label{tab:rand-per-domain}
\begin{tabular}{lccc}
\toprule
\textbf{Variable} & \textbf{Pooled} & \textbf{Low (coffee)} & \textbf{High (hotels)} \\
\midrule
Age                   & 7.21 (12), .84  & 13.32 (12), .35 & 6.71 (12), .88 \\
Gender                & 0.04 (3), $>$.99 & 0.09 (3), $>$.99 & 0.10 (3), $>$.99 \\
Ethnicity             & 3.83 (3), .28   & 2.00 (3), .57   & 2.75 (3), .43 \\
Consumption frequency & 4.38 (3), .22   & 5.40 (3), .14   & 3.95 (3), .27 \\
Domain expertise\textsuperscript{\dag} & 1.62, $>$.99 & 4.85, .86 & 5.51, .80 \\
\bottomrule
\end{tabular}
\end{table}

\FloatBarrier

\subsection{Regression Models for the Subjective Evaluation Measures (RQ2)}
\label{app:subjective-models}

We provide the complete covariate-adjusted linear regression models 
behind the subjective-evaluation results in Section~\ref{sec:results-rq2}. 
For each of the five constructs, Table~\ref{tab:subjective-models-full} lists the 
coefficient (with standard error) of every term in the model: the three condition 
contrasts against the preference-only baseline, all demographic and background 
covariates, item domain, and biospheric value orientation. The main text reports 
only the condition contrasts; the full models are given here for completeness and 
to show that the covariates leave the substantive pattern unchanged.

Two covariates are worth highlighting. Biospheric value orientation---participants' 
baseline concern for the environment---predicted more favorable perceptions across 
the board: higher sustainability impact ($b=+0.22$, $p<.001$), explanation clarity 
($b=+0.10$, $p=.001$), and choice satisfaction ($b=+0.14$, $p<.001$), and lower 
choice difficulty ($b=-0.15$, $p=.005$). Self-reported domain expertise showed a 
similar gradient, with more experienced participants rating usability and 
satisfaction higher (e.g., satisfaction: advanced $b=+0.75$, $p=.002$; expert 
$b=+0.74$, $p=.007$), and frequent consumers of the target item perceived greater 
sustainability impact ($b=+0.32$, $p=.030$). Item domain and the demographic 
covariates (age, gender, ethnicity) produced only scattered effects and did not 
moderate the condition effects (Section~\ref{sec:results-generalization}).

\begin{table*}[t]
  \centering
  \caption{Full covariate-adjusted linear model coefficients for the five 
  subjective evaluation constructs (pooled sample, $N=529$). Each cell reports the 
  unstandardized coefficient $b$ with its standard error in parentheses. Reference 
  categories: condition = preference-only baseline; age = 18--24; gender = male; 
  ethnicity = Caucasian; consumption frequency = infrequent; expertise = novice; 
  item domain = low involvement. MF = mechanism-free, FR = framing, DSN = 
  descriptive social norms; S.\ Impact = sustainability impact. $^{*}p<.05$, 
  $^{**}p<.01$, $^{***}p<.001$.}
  \label{tab:subjective-models-full}
  \begin{tabular}{l lllll}
    \toprule
    \textbf{Predictor} & \textbf{S.\ Impact} & \textbf{Exp. Usability} & \textbf{Exp. Clarity} & \textbf{Decision Difficulty} & \textbf{Decision Satisfaction} \\
    \midrule
    MF vs.\ baseline  & \makecell[r]{$+1.20^{***}$\\ \scriptsize(0.15)} & \makecell[r]{$+0.47^{**}$\\ \scriptsize(0.15)} & \makecell[r]{$+0.06$\\ \scriptsize(0.10)} & \makecell[r]{$-0.21$\\ \scriptsize(0.17)} & \makecell[r]{$+0.07$\\ \scriptsize(0.09)} \\
    FR vs.\ baseline  & \makecell[r]{$+1.67^{***}$\\ \scriptsize(0.15)} & \makecell[r]{$+0.50^{***}$\\ \scriptsize(0.15)} & \makecell[r]{$+0.11$\\ \scriptsize(0.10)} & \makecell[r]{$-0.84^{***}$\\ \scriptsize(0.17)} & \makecell[r]{$+0.07$\\ \scriptsize(0.09)} \\
    DSN vs.\ baseline & \makecell[r]{$+1.33^{***}$\\ \scriptsize(0.15)} & \makecell[r]{$+0.40^{**}$\\ \scriptsize(0.15)} & \makecell[r]{$+0.04$\\ \scriptsize(0.10)} & \makecell[r]{$-0.40^{*}$\\ \scriptsize(0.17)} & \makecell[r]{$+0.04$\\ \scriptsize(0.09)} \\
    \addlinespace
    Age 25--34 & \makecell[r]{$-0.07$\\ \scriptsize(0.15)} & \makecell[r]{$-0.13$\\ \scriptsize(0.14)} & \makecell[r]{$-0.00$\\ \scriptsize(0.09)} & \makecell[r]{$+0.28$\\ \scriptsize(0.16)} & \makecell[r]{$-0.02$\\ \scriptsize(0.09)} \\
    Age 35--44 & \makecell[r]{$-0.33$\\ \scriptsize(0.17)} & \makecell[r]{$-0.18$\\ \scriptsize(0.17)} & \makecell[r]{$-0.16$\\ \scriptsize(0.11)} & \makecell[r]{$+0.21$\\ \scriptsize(0.19)} & \makecell[r]{$-0.26^{*}$\\ \scriptsize(0.11)} \\
    Age 45--54 & \makecell[r]{$-0.36$\\ \scriptsize(0.21)} & \makecell[r]{$-0.10$\\ \scriptsize(0.20)} & \makecell[r]{$-0.22$\\ \scriptsize(0.13)} & \makecell[r]{$+0.13$\\ \scriptsize(0.22)} & \makecell[r]{$-0.37^{**}$\\ \scriptsize(0.13)} \\
    Age 55+ & \makecell[r]{$+0.05$\\ \scriptsize(0.27)} & \makecell[r]{$-0.25$\\ \scriptsize(0.26)} & \makecell[r]{$-0.17$\\ \scriptsize(0.17)} & \makecell[r]{$-0.13$\\ \scriptsize(0.29)} & \makecell[r]{$-0.33^{*}$\\ \scriptsize(0.16)} \\
    Female & \makecell[r]{$+0.15$\\ \scriptsize(0.11)} & \makecell[r]{$+0.02$\\ \scriptsize(0.11)} & \makecell[r]{$+0.14$\\ \scriptsize(0.07)} & \makecell[r]{$+0.10$\\ \scriptsize(0.12)} & \makecell[r]{$+0.08$\\ \scriptsize(0.07)} \\
    Ethnicity: Other & \makecell[r]{$+0.29^{*}$\\ \scriptsize(0.12)} & \makecell[r]{$+0.36^{**}$\\ \scriptsize(0.12)} & \makecell[r]{$+0.05$\\ \scriptsize(0.08)} & \makecell[r]{$+0.19$\\ \scriptsize(0.13)} & \makecell[r]{$+0.15^{*}$\\ \scriptsize(0.07)} \\
    Frequent consumer & \makecell[r]{$+0.32^{*}$\\ \scriptsize(0.15)} & \makecell[r]{$+0.12$\\ \scriptsize(0.14)} & \makecell[r]{$+0.15$\\ \scriptsize(0.10)} & \makecell[r]{$-0.25$\\ \scriptsize(0.16)} & \makecell[r]{$+0.17$\\ \scriptsize(0.09)} \\
    Expertise: advanced & \makecell[r]{$+0.87^{*}$\\ \scriptsize(0.39)} & \makecell[r]{$+1.10^{**}$\\ \scriptsize(0.38)} & \makecell[r]{$+0.41$\\ \scriptsize(0.25)} & \makecell[r]{$-0.61$\\ \scriptsize(0.42)} & \makecell[r]{$+0.75^{**}$\\ \scriptsize(0.24)} \\
    Expertise: expert & \makecell[r]{$+0.47$\\ \scriptsize(0.45)} & \makecell[r]{$+0.79$\\ \scriptsize(0.44)} & \makecell[r]{$+0.64^{*}$\\ \scriptsize(0.29)} & \makecell[r]{$-0.28$\\ \scriptsize(0.49)} & \makecell[r]{$+0.74^{**}$\\ \scriptsize(0.27)} \\
    Expertise: knowledgeable & \makecell[r]{$+0.36$\\ \scriptsize(0.37)} & \makecell[r]{$+0.71$\\ \scriptsize(0.36)} & \makecell[r]{$+0.30$\\ \scriptsize(0.24)} & \makecell[r]{$-0.45$\\ \scriptsize(0.41)} & \makecell[r]{$+0.42$\\ \scriptsize(0.23)} \\
    Item domain: high inv. & \makecell[r]{$-0.12$\\ \scriptsize(0.15)} & \makecell[r]{$-0.03$\\ \scriptsize(0.14)} & \makecell[r]{$+0.20^{*}$\\ \scriptsize(0.09)} & \makecell[r]{$-0.24$\\ \scriptsize(0.16)} & \makecell[r]{$+0.23^{**}$\\ \scriptsize(0.09)} \\
    Biospheric value & \makecell[r]{$+0.22^{***}$\\ \scriptsize(0.05)} & \makecell[r]{$+0.08$\\ \scriptsize(0.05)} & \makecell[r]{$+0.10^{**}$\\ \scriptsize(0.03)} & \makecell[r]{$-0.15^{**}$\\ \scriptsize(0.05)} & \makecell[r]{$+0.14^{***}$\\ \scriptsize(0.03)} \\
    \addlinespace
    Intercept & \makecell[r]{$+2.16^{***}$\\ \scriptsize(0.47)} & \makecell[r]{$+4.00^{***}$\\ \scriptsize(0.46)} & \makecell[r]{$+5.01^{***}$\\ \scriptsize(0.31)} & \makecell[r]{$+4.81^{***}$\\ \scriptsize(0.51)} & \makecell[r]{$+4.35^{***}$\\ \scriptsize(0.29)} \\
    \bottomrule
  \end{tabular}
\end{table*}

\subsection{Per-Domain Result Figures}
\label{app:per-domain-figures}
\begin{figure}[h]
\includegraphics[width=\textwidth]{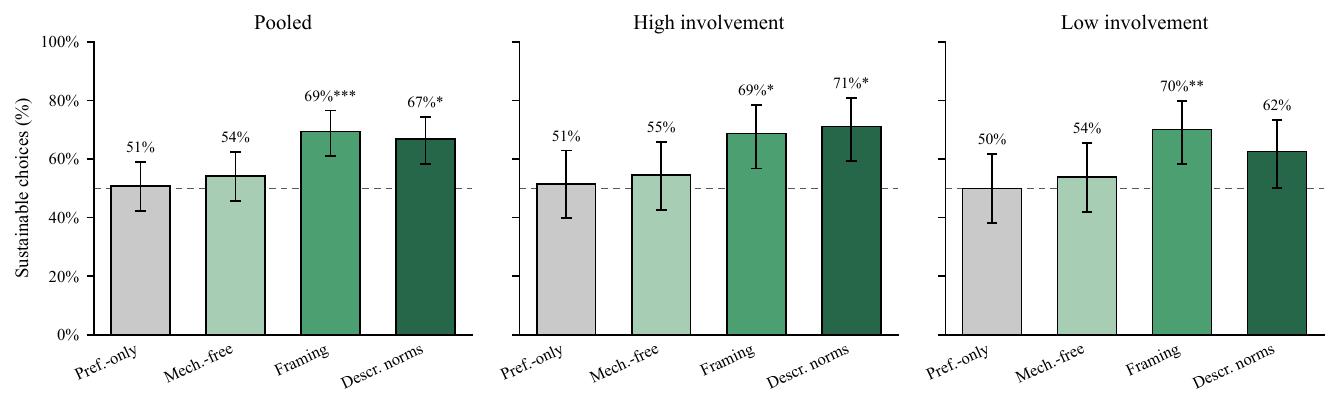}
\caption{Share of sustainable choices by condition, pooled and per involvement 
domain. }
\Description{Share of sustainable choices by condition, pooled and per involvement 
domain.}
\label{fig:choice-pd}
\end{figure}

\begin{figure}[h]
\includegraphics[width=\textwidth]{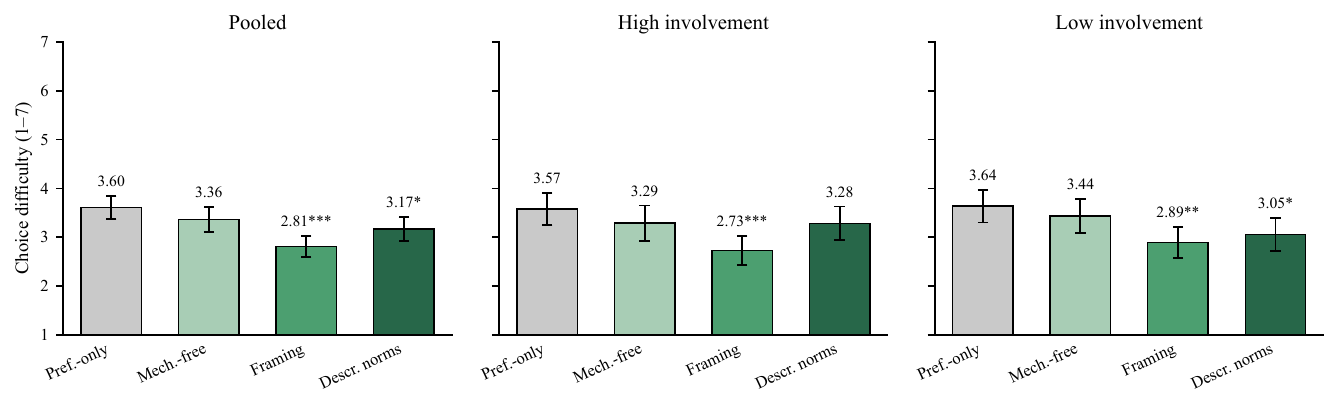}\\[2mm]
\includegraphics[width=\textwidth]{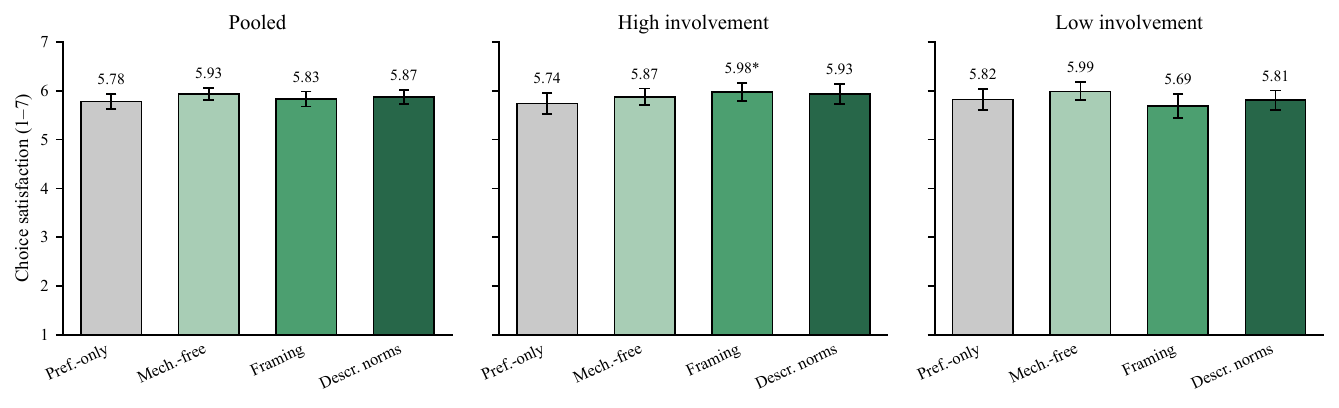}
\caption{Decision-related subjective evaluations by condition, pooled and per 
involvement domain.}
\Description{Decision-related subjective evaluations by condition, pooled and per 
involvement domain. }
\label{fig:subjective-decision-pd}
\end{figure}

The main text reports pooled results, since the condition effects did not differ by 
domain (Section~\ref{sec:results-generalization}). For completeness, 
Figures~\ref{fig:choice-pd}--\ref{fig:subjective-decision-pd} show the same 
outcomes broken down by involvement domain.

% choice per domain
% subjective decision-related per domain

% subjective explanation-related per domain (your existing 3-panel figure)
\begin{figure}[t]
\includegraphics[width=\textwidth]{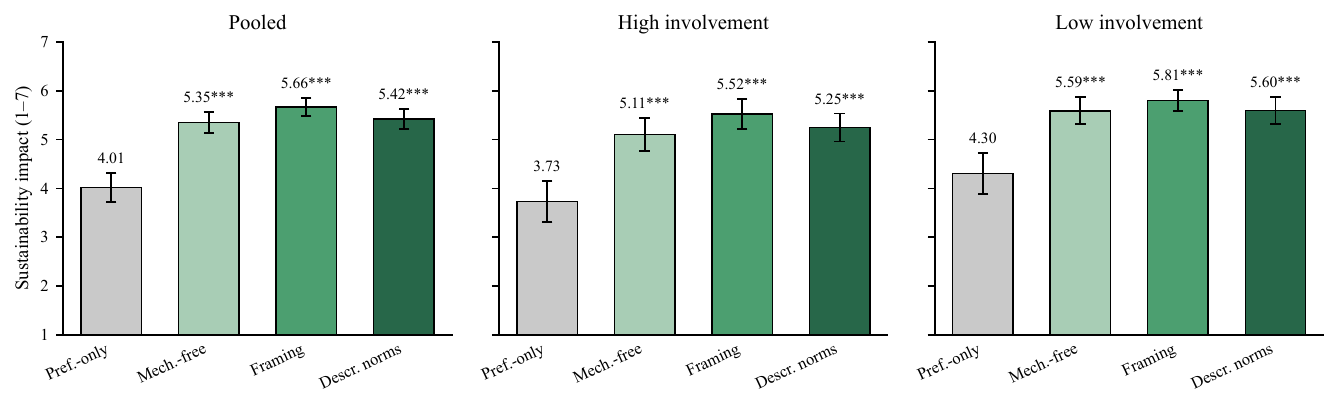}\\[2mm]
\includegraphics[width=\textwidth]{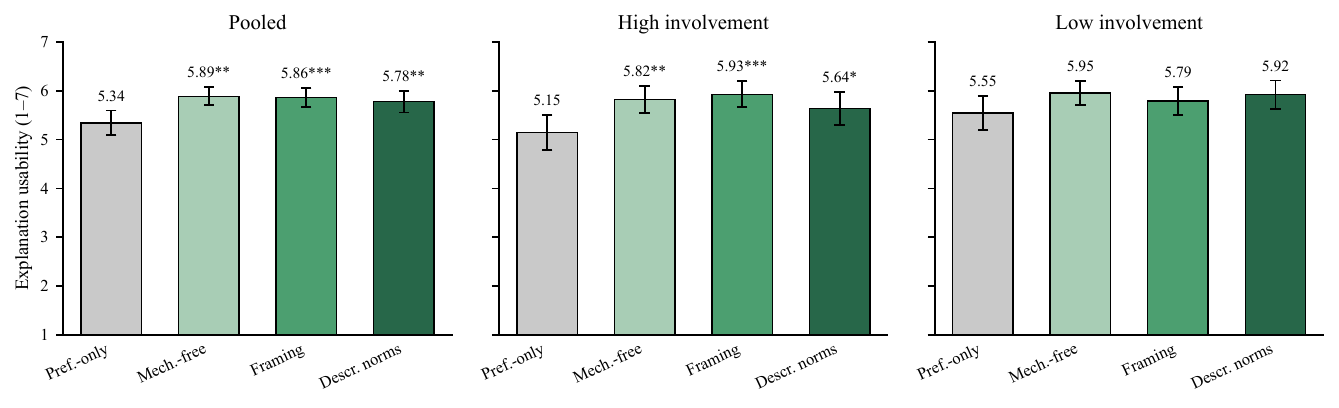}\\[2mm]
\includegraphics[width=\textwidth]{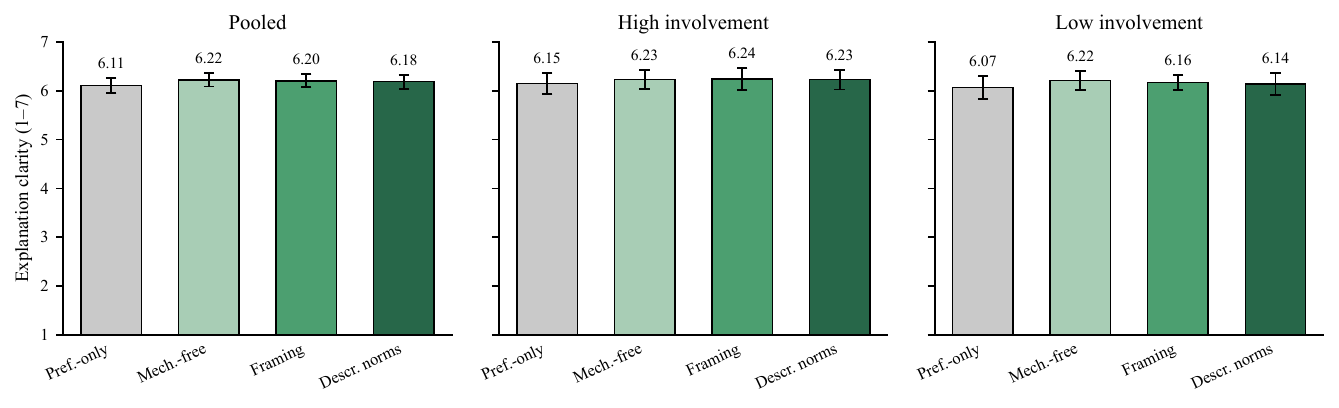}
\caption{Explanation-related subjective evaluations by condition, pooled and per 
involvement domain.}
\Description{Explanation-related subjective evaluations by condition, pooled and per 
involvement domain.}
\label{fig:subjective-explanation-pd}
\end{figure}

\end{document}